\documentclass[pdflatex,sn-nature]{sn-jnl}
\pdfoutput=1                 %

\usepackage{graphicx}%
\usepackage{multirow}%
\usepackage{amsmath,amssymb,amsfonts}%
\usepackage{amsthm}%
\usepackage{mathrsfs}%
\usepackage[title]{appendix}%
\usepackage{xcolor}%
\usepackage{textcomp}%
\usepackage{manyfoot}%
\usepackage{booktabs}%
\usepackage{listings}%
\usepackage{bm}
\usepackage{makecell}
\usepackage{diagbox}
\usepackage{algorithm}
\usepackage{algpseudocode}
\usepackage{todonotes}
\usepackage{placeins}
\usepackage[version=4]{mhchem}

\usepackage{tikz}
\usetikzlibrary{positioning,arrows.meta,calc}
\usepackage{longtable}            %
\usepackage{bibunits}             %
\usepackage{etoolbox}             %
\defaultbibliographystyle{sn-nature}
\defaultbibliography{bib}

\usepackage{siunitx}        %
\DeclareSIUnit\angstrom{\text{\AA}}%

\theoremstyle{thmstyleone}%

\theoremstyle{thmstyletwo}%

\theoremstyle{thmstylethree}%

\title{\bf ReactionAtlas: Ab origine exploration of chemical reaction networks with machine learning }

\author[1,2]{\fnm{Stefan} \sur{Gugler}}\equalcont{These authors contributed equally to this work.}
\author[1,2]{\fnm{Max} \sur{Eissler}}\equalcont{These authors contributed equally to this work.}
\author[1,2]{\fnm{Khaled} \sur{Kahouli}}
\author*[1,2,3,4]{\fnm{Klaus-Robert} \sur{Müller}}\email{klaus-robert.mueller@tu-berlin.de}

\affil[1]{\orgname{BIFOLD -- Berlin Institute for the Foundations of Learning and Data}, \orgaddress{\city{Berlin}, \country{Germany}}}
\affil[2]{\orgdiv{Machine Learning Group}, \orgname{TU Berlin}, \orgaddress{\city{Berlin}, \country{Germany}}}

\affil[3]{\orgname{Department of Artificial Intelligence, Korea University}, \orgaddress{\city{Seoul}, \country{South Korea}}}
\affil[4]{\orgname{MPI for Informatics}, \orgaddress{\city{Saarbr\"ucken}, \country{Germany}}}

\newcommand*{\rvx}{\mathbf{x}}

\begin{document}
\maketitle

\begin{abstract}
    MMapping a chemical reaction network, the graph of minima and transition states (TS) and the elementary reactions connecting them, is the natural language of chemistry, from catalysis to combustion to the origin of life. Constructing such a reaction network for a given chemistry has been impractical: it requires finding and characterizing tens of thousands of TS, a task for which traditional methods such as density functional theory (DFT) are typically prohibitively slow and require reactant and product as input. We introduce ReactionAtlas, which builds a reaction network \textit{ab origine} from a handful of seed molecules and without hand-crafted rules. Specifically, our machine-learned generative model proposes reactions from kinetically sampled candidate compounds and a DFT-trained machine learned force field (MLFF) filters them to valid TS, the resulting products of which enter the search as new seeds. Starting from eight pre-biotic seeds (CH$_2$O, H$_2$O, OH$^-$, H$_3$O$^+$, CO$_2$, H$_2$CO$_3$, HCO$_3^-$, H), ReactionAtlas discovers $\sim$47{,}000 reactions among $\sim$12{,}000 compounds. The MLFF TSs match the PBE0 references within 0.5~\AA{} RMSD in 85\% of the cases and can be easily brought to the PBE0 level. Thus, ReactionAtlas maps small carbohydrate chemistry up to C$_4$H$_8$O$_4$ at unprecedented scale and accuracy, including charge and stereo information. It enables novel insights into many well-studied reaction paths, including the formose cycle, which we highlight for its centrality to the chemical origins of life. Notably, our framework also allows establishing alternative reaction pathways for formose chemistry.

\end{abstract}

\begin{bibunit}

\addtolength{\textheight}{-4cm}

\section{Introduction}

\begin{figure}[!htbp]
    \centering
        \includegraphics[width=1\textwidth]{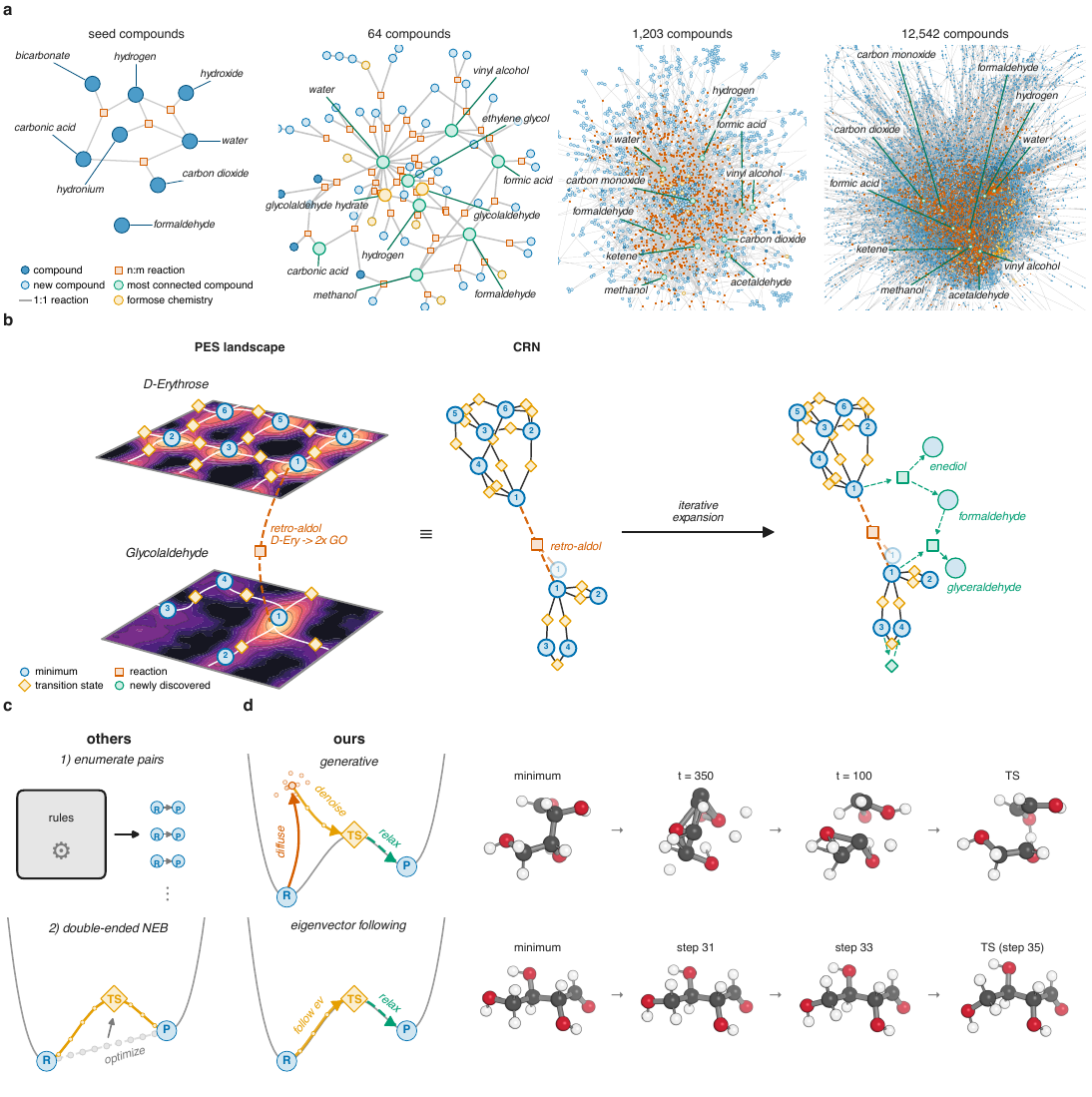}
\caption{
    \textbf{a}, Four snapshots at different stages of the chemical reaction network (CRN) exploration are shown, first the \textit{ab origine} state with only the seed compounds (dark blue) and subsequent states at 64, 1203, and 12,542 compounds (light blue) with ten most connected molecules (green) labeled. The formose relevant chemistry is highlighted in yellow. The direct edges denote isomerizations (1:1 reactions with implicit transition state (TS)) while orange squares denote n:m reactions with an associated TS.
    \textbf{b,} New minima in the CRN are found via intramolecular transformations, i.e. finding a TS on the same PES (yellow diamonds) or via intermolecular reactions (orange square), as illustrated with the retro-aldol reaction of D-erythrose to two glycolaldehyde molecules (left).
    In a more abstract manner, this multi-PES object can be represented as a CRN (middle), where the numbering corresponds to the minima on the PES.
    Iterative expansion via ReactionAtlas leads to new TS and minima, as illustrated in green on the right.
    \textbf{c,} Most common TS search methods enumerate and combine reactants (R) and products (P) via heuristics and find a TS through a form of interpolation between the two, e.g. the nudged elastic band (NEB) approach.
    \textbf{d,} Our method proposes a TS via a diffusion model: From a given minimum (the reactant), noise is injected (orange arrow, as illustrated as going from the minimum to $t=350$ noise) and then denoised (yellow arrow, i.e. going from $t=350$ to the TS. Finding the product and verifying can be done with relaxation (green arrow). Alternatively, following the eigenvectors (ev, yellow, bottom) can find or refine a TS candidate.
    }
    \label{fig:concept}
\end{figure}

The exploration and identification of chemical reaction networks (CRNs) is central to understanding how complex behavior can emerge from simple molecular building blocks.\cite{unsleber2020,sutherland2016,xavier2020,muchowska2020}
One of the prime examples of CRNs is the \textit{formose} reaction, an intricate web of aldol condensations, which is a prototypical case of an auto-catalytic cycle.\cite{kua2024,mizuno1974}
First observed by Butlerov in 1861 \cite{butlerow1861,orgel2000} and mechanistically elucidated by Breslow in 1959,\cite{breslow1959,appayee2014} it remains a striking example of complexity emerging from simple precursors.\cite{cleaves2008}
Through iterative condensations, the formose network generates sugars including glycolaldehyde and glyceraldehyde from \textit{form}aldehyde,\cite{venturini2024} eventually producing ald\textit{ose}s and regenerating formaldehyde (Fig.~\ref{fig:formose_results}a). 
It thus offers a plausible prebiotic route to riboses, underwriting the RNA world hypothesis\cite{gilbert1986,joyce2002,powner2009,szostak2001}.
However, the formose cycle's combinatorial explosion of intermediates and side reactions makes it difficult to analyze experimentally\cite{bris2024,sutton2025,yi2022}.
Hence, despite its importance it still eludes a complete quantitative, network-level characterization.

Likewise, manual reaction-by-reaction analysis and computational chemistry methods have struggled to keep pace with the network's complexity.\cite{kua2024,venturini2024}
These methods rely on rule-based enumeration of reactant-product pairs. However, such enumeration rules not only explode combinatorially but presuppose reaction mechanisms of the target chemistry, which figuratively speaking ``puts the cart before the horse''.
Recent autonomous frameworks address part of this gap\cite{perezvilla2020,wolos2020,granda2018}:
SCINE\cite{weymuth2024} recovers the cycle from a network built around formaldehyde plus glycolaldehyde\cite{simm2017b}, and \textit{ab initio} molecular dynamics methods now map portions of the formose network\cite{kan2025a,kan2026}.
While accurate, these approaches are limited in scale\cite{wang2014,stan-bernhardt2024,unsleber2022,zhao2021} and generality\cite{wen2024yaks,wolos2020,margraf2023}.

The cost of CRN construction not only lies in sheer network size but also in locating TSs. 
A reaction path connects two minima on the PES through a first-order saddle point, the TS. Once the TS is found the rest of the path and the connected minima can be recovered cheaply by following forces from the TS\cite{schlegel2011}. The TS itself, however, is difficult to find as it is energetically unstable and thus not recoverable using forces alone. TS-search methods divide into single-ended methods\cite{schlegel1982,maeda2011,zimmerman2015}, which start from one minimum and search for connected saddle points, and double-ended methods\cite{mills1995,zimmerman2013}, which condition the search on both reactant and product (Supplementary Information section~\ref{app:ts_methods_sec}). 
Existing autonomous CRN frameworks often build on double-ended TS search\cite{zhao2021,zhao2022,steiner2022} or on reactive-complex assembly followed by hybrid double-ended/single-ended TS refinement\cite{simm2017b,bergeler2015}: 
each candidate reactant-product pair needs to be supplied, usually through pairing rules based on heuristics (see Fig.~\ref{fig:concept}c). The TS search then either validates or rejects it which becomes increasingly costly as the networks and reactant-product molecules grow, which hampers generalization.

Machine learning is a viable route to address the complexity of this process, but it has so far not been used to drive open-ended exploration where reactant-product pairs are discovered rather than supplied\cite{kim2024,duan2023,schlama2026,vangerwen2022}.
Notably, a recently proposed generative CRN framework still relies on rule-based reaction enumeration and conditions TS generation on supplied reactant-product pairs\cite{bytecrn2026}.
Indeed, the systematic application of ML to constructing and exploring prebiotic CRNs has been identified as a still-open challenge\cite{anton2026}, with reactive-MLFF\cite{behler2007,chmiela2017,unke2021b} demonstrations so far mostly limited to single targets\cite{huet2024,zhang2024ani1xnr}.

Our ReactionAtlas framework circumvents these computational challenges with two innovations: (a) {\em single-ended} generative TS proposers \cite{kahouli2025,kahouli2024, eissler2026simple} and an MLFF critic\cite{eissler2026simple} and (b) a technique for prioritizing the exploration of kinetically validated compounds.
The combination of these advances allows ReactionAtlas to efficiently and accurately explore large CRNs. Notably, the {\em absence} of chemistry-specific rules inherent to our framework makes it  readily applicable to new chemistry. To evaluate ReactionAtlas' capabilities we conduct a large-scale exploration of carbohydrates at PBE0 accuracy starting from eight prebiotically plausible seeds (CH$_2$O, H$_2$O, OH$^-$, H$_3$O$^+$, CO$_2$, H$_2$CO$_3$, HCO$_3^-$, and H). We call this \textit{ab origine} CRN exploration: we start from only a small set of seed compounds and propagate the reaction network without imposed mechanisms.

In this manner, we could explore $\sim$12{,}000 compounds, $\sim$31{,}000 conformers, and $\sim$47{,}000 reactions up to 10 heavy atoms, yielding one of the first charge and stereo-resolved data sets of this kind. 
As a quantitative measure demonstrating relevance of our exploration we found more than 80\% of up to C$_4$ carbohydrates that have dedicated Wikipedia entries.
Interestingly, our CRN network contains the complete formose cycle and more: the canonical Breslow route through \textsc{d}-glyceraldehyde and \textsc{l}-erythrose, and a previously unreported enediol shortcut through a C$_3$ enediol and \textsc{d}-erythrulose, both running from the same C$_1$ - C$_2$ feedstock to the same tetrose products through different intermediates.
Resolving full stereochemistry, we recover the formose cycle across all six tetrose stereoisomers, the chiral landscape from which any account of biological homochirality must depart\cite{blackmond2004,blackmond2019}.

While our focus in this work was placed on formose chemistry, there are many other potential CRN exploration targets, e.g.~in catalysis,\cite{ulissi2017,steiner2022} combustion,\cite{broadbelt1996,sankaran2007} polymerization,\cite{vinu2012} atmospheric chemistry,\cite{vereecken2015} astrochemistry,\cite{meisner2016,herbst1973} environmental chemistry,\cite{petrus2025} or mass spectrometry.\cite{spotte-smith2023}
Our proposed ReactionAtlas framework would be readily applicable to any of the above as it provides a viable, accurate and efficient basis for exploring arbitrary CRNs across reaction chemistry.

\section{Results}
\label{sec:results}

\begin{figure}[!htbp]
    \centering
     \makebox[\linewidth][c]{%
        \includegraphics[width=1\textwidth]{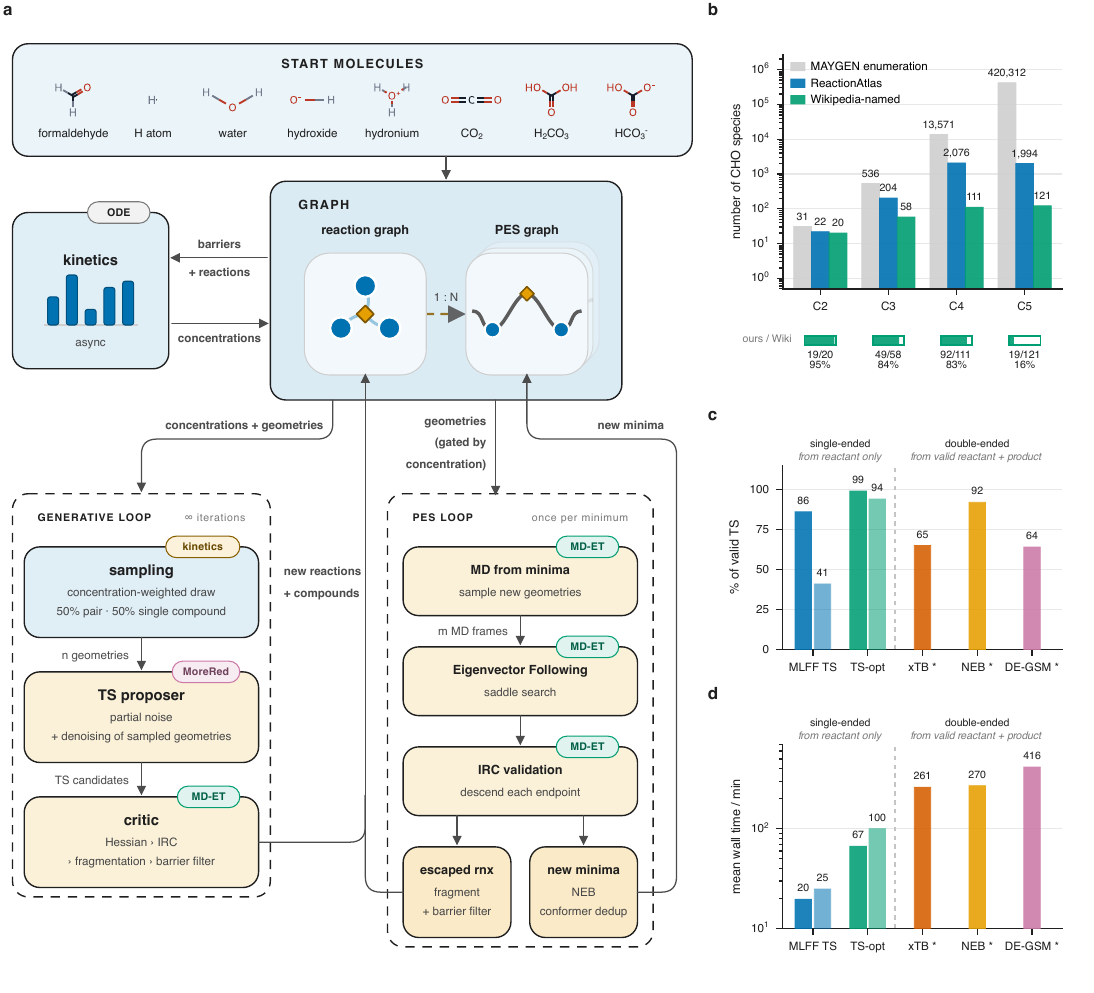}
    }
\caption{
    \textbf{a:} Overview of our proposer-critic framework.
    Start molecules initialize a CRN graph of compounds and reactions, whose barriers feed a kinetics module to get concentrations for sampling.
    The generative loop draws sampled geometries according to concentration weights (uni- or bimolecular), proposes TSs with MoreRed, and uses several methods to validate the candidates.
    The PES loop, for each new compound, runs a short MD trajectory with the MD-ET force field, walks uphill using P-RFO and validates each saddle point found this way similarly to the generative loop, also recovering inter-molecular reactions as `escaped' reaction paths.  Reactions, minima and conformers found in both loops are added to the CRN, which refreshes the kinetic model.
    \textbf{Right:} 
    \textbf{Relevance, quality, and speed of the MLFF TS workflow.}
  \textbf{(b)~Relevance}
  Number of species per carbon count:
      exhaustive enumeration (gray),
      ReactionAtlas (blue), 
      and Wikipedia-named compounds (green).
  The bars below each group give the fraction of Wikipedia-named species ReactionAtlas recovers.
  \textbf{(c)~Quality.}
  Fraction of valid TS, i.e., Hessian possesses one imaginary frequency. 
  The darker bar represents results for isomerizations (i.e. the 1$\to$1 reactions) and the lighter bar represents results for all other types of reactions (2$\to$1, 1$\to$2, 2$\to$2).
  Methods marked with $\star$ (xTB$\to$DFT, NEB-TS, DE-GSM) accept only single-fragment endpoints and so were evaluated on a set of 100 isomerization reactions only. A PBE0 TS-optimization starting from the MLFF TS prediction raises the validity at some additional cost (see panel d). 
  \textbf{(d)~Speed.}
  Mean wall time per successfully-found TS, i.e. we do not consider failed optimizations (see panel c).
   \textbf{Note}: For panels \textbf{c} and \textbf{d} single-ended and double-ended methods are not directly comparable as single-ended methods find TS from reactant geometries alone while we supply PBE0-validated pairs of reactants and products to the double-ended methods, i.e. low-energy TS can be found for all tested structures. However, proposing these matching reactant--product pairs is usually a costly part of constructing a CRN using double-ended methods not considered here.
  }
    \label{fig:crn_exploration}
\end{figure}

\subsection{\textit{Ab origine} Machine-Learned Chemical Reaction Networks}
\label{sec:theory_diff}

A crucial aspect of CRN construction is {\em efficiently} and {\em accurately} finding TSs.
Since we propose to conduct the CRN exploration \textit{ab origine}, the TS-search methods additionally need to be {\em single-ended}, i.e. only requiring reactants as inputs.
To satisfy these principles we propose two complementary ML methods for single-ended TS search (see Fig.~\ref{fig:crn_exploration}), the generative loop and the PES loop.
Both expand the CRN, which is hierarchically structured as a reaction graph of minima (blue circles) and TSs (yellow diamonds) coupled to $N$ PES graphs, one per compound (see Fig.~\ref{fig:crn_exploration}), each containing the corresponding conformers and TSs between them. 
Exploration is seeded from a small primordial feedstock of eight species (top panel: CH$_2$O, H, H$_2$O, OH$^-$, H$_3$O$^+$, CO$_2$, H$_2$CO$_3$, HCO$_3^-$). 
With the TSs we determine the reaction barriers and can hence solve the kinetic coupled ODEs\cite{eyring1935} to obtain a concentration for each compound (blue box, left).
We use these concentrations to allocate computational resources to regions of the graph deemed chemically plausible (see the kinetic exploration guidance in the Online Methods and Supplementary Information section~\ref{si:kinetics}).

Together, these components grow the reaction graph $\mathcal{G}$ \textit{ab origine},
\begin{equation*}
\mathcal{G}_{i+1} = 
    \mathcal{G}_i \cup 
    \bigl\{
        (\rvx_\text{TS},\rvx_\text{min}') :
            \rvx_\text{min} \sim \pi_T(\mathcal{G}_i),
            \rvx_\text{TS}  \sim  p_\phi(\cdot \mid \rvx_\text{min}),
            \mathcal{C}(\rvx_\text{TS}) = 1
    \bigr\} \, ,
\label{eq:fixpoint}
\end{equation*}
where
    $\pi_T$ is the kinetics-weighted steady-state distribution over the current network $\mathcal{G}_i$ at exploration step $i$,
    the generative proposer $p_\phi$ samples a TS $\rvx_\text{TS}$ from a \textit{single} drawn minimum $\rvx_\text{min}$,
    and the MLFF critic $\mathcal{C}$ accepts only validated saddle points together with the new minima $\rvx_\text{min}'$ they connect, which seed the next iteration (Supplementary Information section~\ref{si:generative}).

The generative loop (bottom left) of our framework consists of a generative TS proposer followed by a number of refinement steps.
First, current minima are weighted by their concentrations, where half of the draws are pairs of compounds, the other half single compounds (blue box). 
Specifically, we use MoreRed\cite{kahouli2025,kahouli2024}, a denoising diffusion model trained on the $\omega$B97X-D3 Grambow TS dataset\cite{grambow2020b}, to propose TSs for these samples after partial noising (Supplementary Information section~\ref{si:generative}). 
Then we refine these candidates by inspecting several properties a TS should possess, using a QCML-trained MD-ET model\cite{ganscha2025,eissler2026simple} (Supplementary Information section~\ref{si:shared}) as a zero-shot predictor of forces and Hessians. 
These properties include: 
    a Hessian with a single imaginary eigenvalue at the TS, 
    two distinct minima at the IRC endpoints, 
    and several conditions on the barrier heights (Supplementary Information section~\ref{si:generative}). 
    Reactions passing these conditions, along with their minima, enter the graph.

The PES loop (bottom right) within our framework is executed once per newly discovered minimum for compounds above a defined concentration threshold (Supplementary Information section~\ref{si:pes}) and it also uses the MD-ET model. 
First, a short 300~K Langevin MD trajectory (2.5~ps) samples the local configuration space. 
Points from this trajectory serve as starting points for partitioned rational-function optimization (P-RFO\cite{banerjee1985,baker1986}), i.e. following Hessian curvature directly to the nearest saddle point.
For the process to reliably work on unseen molecules using an MLFF, the model must be extraordinarily robust, which required several modifications to MD-ET in order to guarantee smoothness (Supplementary Information section~\ref{sec:shared-mdet}).

TS finding requires fast Hessian evaluation which costs on the order of tens of milliseconds with MD-ET; $\sim$1/3 of optimizations then converge to a saddle point.
We typically run 32 P-RFOs in parallel for each minimum for sufficiently exploring local geometry. 
TS candidates are validated by the same criteria as in the generative loop (Supplementary Information section~\ref{si:pes}). 
Valid TSs found by our method can be of two types: 
new conformational minima on the PES, which are deduplicated using nudged elastic band (NEB\cite{mills1995, jonsson1998}) optimization, or `escaped reactions', where the bond structures of the two minimum geometries are different. We add these to the reaction graph.

In our main experiment we do \textit{ab origine} exploration from the eight seed compounds as explained above. To resolve the formose cycle in more detail, a refinement run was re-initialized from a formose sub-network of 99 formose-relevant compounds (Supplementary Information section~\ref{app:formose_ic}) to find additional TS proposals and reactions.

\subsection{Evaluating relevance, quality, and speed}
\label{sec:validation}

An ideal CRN exploration method should satisfy three requirements:
it should be efficient, it should  yield accurate reaction barriers and the chemistry explored should be relevant.
We evaluate ReactionAtlas along each dimension.

\textbf{Relevance.} 
Any generator left to enumerate freely can produce vast numbers of formally valid but chemically meaningless molecules.
The question is, whether ReactionAtlas covers the chemically relevant region thoroughly while not enumerating all possible chemical permutations. At each carbon count, we compare the set of compounds recovered by ReactionAtlas with two other sets (Fig.~\ref{fig:crn_exploration}b).
The full combinatorial enumeration\cite{maygen} (gray)
and the small subset of compounds with a Wikipedia article (a proxy for established relevance, green).
At C$_2$, recovery of the named species is almost complete (19/20, 95\%), with only the four-membered ring 1,3-dioxetane missing. 
At C$_3$ and C$_4$ recovery remains high (84\% and 83\%), even as the combinatorics explode at C$_4$ and only $<$1\% of the possible chemistry is found on Wikipedia.
This proxy measure of relevance in the Wikipedia entry nicely indicates that ReactionAtlas is exploring relevant chemistry rather than the full combinatorial space.
At C$_5$ recovery drops to 16\%, reflecting the computational limit we imposed on the exploration. Supplementary Fig.~\ref{sfig:degree} shows additionally that Wikipedia-named compounds have systematically higher node degrees in the network. As an additional measure of completeness we checked for coverage of the set of  45 small C/H/O molecules detected in interstellar space, 41 of which are present (Supplementary Information section~\ref{si:relevance}). \cite{mcguire2022,pickett1998,muller2005}

\textbf{Quality.} A TS is the saddle point a reaction must pass through \cite{henkelman2000climbing}, and its geometry and energy determine the entire reaction path, so a reaction network is only as good as its TS.
We recomputed all $\sim$47{,}000 TS with DFT (PBE0/def2-TZVPP) and inspected whether each was a true first order saddle point, as required for a valid TS.
Most proposed structures are already valid:
86\% of the rearrangements (one molecule to one) and 41\% of the harder reactions that break or join molecules.
A DFT refinement, initialized at our proposed geometry, improves these fractions to 99\% and 94\%, respectively (Fig.~\ref{fig:crn_exploration}c).

To give additional context, we provide an \textit{imperfect} comparison to several double ended alternatives (see Fig.~\ref{fig:crn_exploration}c) - an NEB-TS\cite{mills1995, jonsson1998}, a double-ended GSM with pyGSM\cite{zimmerman2013} and a multi-level approach employing xTB-NEB\cite{grimme} with subsequent DFT TS-optimization. Since these alternatives are double-ended, each needs the product as well as the reactant, whereas ReactionAtlas proposes one or more TSs starting from a single reactant geometry. To make the comparison feasible we provide these methods with PBE0-valid endpoints for predicting a TS. 
They also handle only rearrangement, i.e., can not be directly used for bimolecular reactions, which is why they carry a single bar in Fig.~\ref{fig:crn_exploration}c. Even on rearrangements only and using ideal product geometries, they do not surpass ReactionAtlas (full benchmark in Supplementary
Information section~\ref{si:quality}).

Comparing \textbf{Speed} between ReactionAtlas and other methods for TS search faces similar comparison issues as discussed for assessing quality above: Single-ended and double-ended methods do not solve the same problem. Using only reactants as inputs and mainly using modest L4 GPU nodes during our main experiment ReactionAtlas finds a TS it considers valid every $\sim$23 min. Note that exploring PES graphs and finding multiple stable minima as well as transitions between them for each compound is a costless byproduct using this calculus (see Fig.~\ref{fig:crn_exploration}d, blue bars). We can further refine the found transition states to be uniformly of PBE0-quality, which requires only a small amount of extra computational overhead since the ML-discovered TS tend to either already match a PBE0-TS or are close to one (green bars).
Meanwhile, given PBE0-valid reactant and product endpoints NEB-TS requires 270 min, DE-GSM 416 min, and the xTB$\rightarrow$DFT pipeline 261 min per reaction using a 4-core CPU node. Note that we exclude the runtimes of failed searches (see Fig.~\ref{fig:crn_exploration}c), which is higher on average. The apparent wall time advantage of ReactionAtlas increases by a further 1-2 orders of magnitude in practice because double ended methods need to enumerate many proposed reactant-product pairs to find a pair with a low-energy TS, as we provide here. For further details, see Supplementary Information section~\ref{si:quality}.
  
\subsection{Discovering the formose auto-catalytic cycle}
\label{sec:formose_cycle}

\begin{figure}[!htbp]
    \centering
    \makebox[\linewidth][c]{%
        \includegraphics[width=1\textwidth]{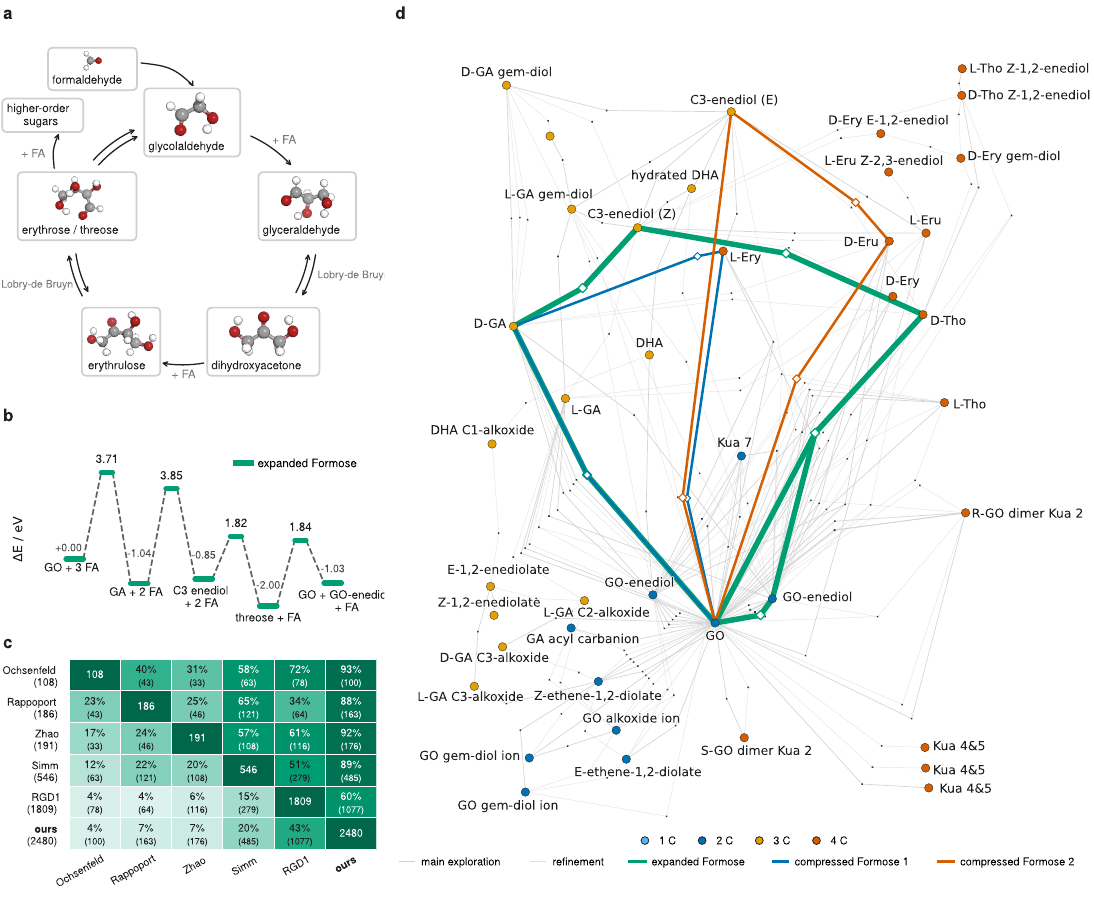}
    }
    \caption{
    \textbf{a)}
    The core formose auto-catalytic cycle with two formaldehydes (FA) condense to glycolaldehyde (GO), which undergoes aldol addition with another FA to glyceraldehyde (GA, in tautomeric equilibrium via Lobry de Bruyn--van~Ekenstein mechanism with dihydroxyacetone, DHA). Adding a third FA produces the tetrose isomers (erythrose, threose, erythrulose), and a retro-aldol step releases two GO, closing the auto-catalytic cycle with one GO in excess. Higher-order aldoses (including ribose) build up by further additions of aldols to the tetroses.
    \textbf{b)}
    Reaction-coordinate PBE0 energy profile of a 4-step formose cycle recovered as an elementary step sequence.
    \textbf{c)}
    Pairwise asymmetric coverage between published CHO reaction-network datasets and our network within the C$_4$H$_8$O$_4$ limit:
    cell $(i,j)$ is the fraction of dataset $i$ also contained in dataset $j$, the diagonal gives each set size. Note, we do not consider stereochemistry and charged species to make the comparison possible, as only ReactionAtlas models both.
    \textbf{d)}
    The formose reaction subgraph with three auto-catalytic cycles overlaid: the elementary cycle of panel (b) (green) and two compressed variants in which a retro-aldol and a tautomerization collapse into a single TS, the canonical Breslow route via \textsc{d}-GA and \textsc{l}-erythrose (blue) and the C$_3$-enediol route via \textsc{d}-erythrulose (orange).
    }
     \label{fig:formose_results}
\end{figure}

The formose reaction is the canonical example of prebiotically plausible auto-catalysis, run at 40-70$^\circ$C in alkaline water with catalysts such as Ca$^{2+}$\cite{butlerow1861,breslow1959,mizuno1974,appayee2014}.
We seed ReactionAtlas with eight pre-biotic compounds (CH$_2$O, H$_2$O, OH$^-$, H$_3$O$^+$, CO$_2$, H$_2$CO$_3$, HCO$_3^-$, H) and build the CRN \textit{ab origine}, notably without adding carbonyl-chemistry rules.
The discovered CRN contains 47,000 elementary steps between 12,000 compounds that resolve stereochemistry and charge (which no previous formose network has done) (Fig.~\ref{fig:formose_results}c).
Due to its comprehensiveness and internal consistency, we can systematically study three reaction mechanisms under catalyst-free gas-phase conditions.

The first and oldest question is how the formose cycle gets started.
To turn over, the cycle needs a first molecule of glycolaldehyde (GO), and the only way to make one from the feedstock is to dimerize two formaldehyde (FA) molecules (Fig.~\ref{fig:formose_results}a).
This first carbon-carbon bond forms very slowly, so experiments usually seed with GO directly\cite{socha1981,kieboom1984,orgel2004,simm2017b} and it has long been suspected that runs \textit{without} added GO are in fact initiated by trace sugar impurities already in the flask\cite{weiss1970}.
ReactionAtlas was \textit{not} seeded with GO and nevertheless recovers 2FA$\to$GO as an elementary step at a forward barrier of 33.4 kcal/mol (Supplementary Information section~\ref{si:wider_literature}).
Hence, we show that a dimerization is chemically viable.
Whether this dimerization is actually faster than the trace-impurity route is a consideration of solution-phase kinetics that our gas-phase barriers do not decide.

The second question  that has been subject to debate in chemistry for decades is how sugars interconvert. For the formose cycle to run, aldose and its ketose must interconvert, i.e. their carbonyl group needs to relocate (Fig.~\ref{fig:formose_results}a). 
The proposed mechanism are 
    direct internal hydrogen transfer (a 1,2-hydride shift)\cite{appayee2014,cheng2015,delidovich2014}
    and
    a two-step route through a shared enediol intermediate (the Lobry de Bruyn--Alberda van Ekenstein mechanism, or LdB--AvE mechanism)\cite{lobrydebruyn1895,yi2022,sutton2025,kim2011,kua2024}.
In our ReactionAtlas' CRN  the interconversion runs entirely 
through the enediol, while the direct transfer between the neutral sugars is absent, supporting LdB--AvE.
This absence is not a blind spot of the method as ReactionAtlas finds 223 hydrogen-transfer transition states elsewhere.
Note, however that classical barriers alone cannot exclude a quantum tunneling contribution\cite{appayee2014} to the direct route (Supplementary Information section~\ref{si:wider_literature}).

The third question, raised only recently\cite{bris2024}, is how the cycle closes.
A 4-carbon sugar (a tetrose) must break in two parts by a retro-aldol step (Fig.~\ref{fig:formose_results}a), which can happen in two ways:
A symmetric 2+2 split yielding two GO, doubling the GO that started the cycle (which makes it autocatalytic).
An asymmetric 3+1 split gives a C$_3$+FA, only shuffling a carbon without doubling anything. Which split dominates depends on the sugar.

Our results indicate that aldotetroses favor the productive 2+2 route (by 28 kcal/mol),
and
ketotetroses the unproductive 3+1 route (by 33 kcal/mol).
Bri\v{s} \textit{et al.}\cite{bris2024} inferred the ketose preference experimentally. ReactionAtlas reproduces these results and adds the complementary aldose preference that closes the loop (Supplementary Information section~\ref{si:wider_literature}).

Together, these mechanisms form auto-catalytic cycles, which we recover
\textit{ab origine} as a sequence of elementary steps without any reaction rules.
Fig.~\ref{fig:formose_results}b, shows an aldol reaction of GO and FA to GA, enolization to the C$_3$ enediol, a second aldol to a tetrose, and the closing
retro-aldol bringing GO+3FA to 2GO+FA.
This elementary description is the mechanistically most detailed including the LdB-AvE steps.
The same loop is further recovered in two ``compressed'' forms (Fig.~\ref{fig:formose_results}d), the canonical Breslow route via \textsc{d}-GA and \textsc{l}-erythrose (blue) and a C$_3$-enediol route
via \textsc{d}-erythrulose (orange, also found by \cite{simm2017b}), each collapsing a retro-aldol and a
tautomerization into a single TS and in this manner increasing its worst barrier to 4.25 and
4.90~eV, respectively (Supplementary Information section~\ref{app:compressed}).
On the C$_1$/C$_2$ side we recover most of the species
manually mapped in the FA-\cite{kua2013} and GO-oligomerization\cite{kua2013b}
studies of Kua \textit{et al.} (Supplementary Information section~\ref{si:wider_literature}).
The network also allows us to count auto-catalytic motifs systematically.
Blokhuis \textit{et al.}\cite{blokhuis2020} classified the possible subgraphs of an auto-catalytic cycle into topological types, ordered by how intricate the feedback is (type I, e.g. Breslow, is common, type V is rarest and most intricate). Types IV and V had been predicted on paper but \textit{never identified} in any real chemical system.
Our network contains around 90 distinct auto-catalytic carbohydrate subgraphs of Types I to III, all sharing GO as the regenerated catalyst, and a single Type IV core.
To our knowledge, this is the first time a Type IV motif has been discovered rather than constructed in theory (Supplementary Information section~\ref{si:autocat}).
  
Because ReactionAtlas' CRN is very large, it gives a global view, which shows that formose chemistry is only a transient stage in the network.
At every size from C1 to C4,
the most stable molecule is \textit{not} a sugar but a carboxylic acid or ester, 27--38 kcal/mol lower in energy\cite{larralde1995,kua2025}.
Sugars therefore accumulate only kinetically, i.e. due to the accessible barriers, but given enough time the system will end up with the more stable acids.
Several side reactions continuously remove material out of the sugar chemistry to that acidic fate.

Restricted to the same size limits, ReactionAtlas is a superset of the species discovered by five earlier formose studies built using very different methods.
Specifically, the rule-based explorations of Simm and Reiher\cite{simm2017b}
and Rappoport \textit{et al.}\cite{rappoport2014},
the MD nanoreactor\cite{wang2014,wang2016} of Stan \textit{et al.}\cite{stan2022},
the glucose-pyrolysis network of Zhao and Savoie\cite{zhao2023},
and the enumerated reaction database RGD1\cite{zhao2023rgd1} (Fig.~\ref{fig:formose_results}c).
In summary, ReactionAtlas has discovered about 90\% of the dedicated formose and pyrolysis networks, and 60\% of the much larger general database RGD1, while those prior CRNs in turn contain at most 43\% of ours (Supplementary Information section~\ref{si:pairwise_coverage}).

\section{Discussion and Conclusion}
\label{sec:discussion}
Chemistry \textit{is} the science of reactions. Here, CRNs are a powerful computational technique for exploring and understanding the complex web of reactions. However, so far, CRN methods were purpose-built for specific chemistries and were too slow for comprehensive exploration.
ReactionAtlas is a method for \textit{ab origine} CRN exploration which allows the exploration of reaction networks of unprecedented scale by using generative machine learning, specifically, we propose two methods for single-ended TS search. In principle, our ReactionAtlas framework can be applied to any chemistry. It is fast, accurate and finds relevant chemistry. This is demonstrated by exploring carbohydrate chemistry ab origine, starting from eight seed compounds only, and recovering a reaction network of unparalleled size, detail and accuracy. Analyzing this autonomously and bias-free explored CRN, we find, among others, the complete formose cycle as well as several previously undescribed formose pathways. An equivalent exploration would be viable for other starting compounds, e.g., nitrogen chemistry, which could in principle make the origin of amino acids accessible. 

ReactionAtlas' performance and its wide applicability will facilitate many future extensions. For example explicitly modeling solvation effects would be helpful as in its current form ReactionAtlas models compounds in a vacuum.
Further research directions are enabled by using our generated CRN dataset, such as iterative self-improvement, where the validated TS can be used to train an improved proposer model. Moreover, while we focus on broadly covering an entire reaction space by exploring all kinetically valid compounds, a narrower, dedicated exploration objective could be imposed to limit compute (e.g. driving towards a certain compound class of special interest, such as amino acids). Here too, the provided CRN may be helpful for training a guiding function for prioritizing directions of further exploration in chemical space. For this, the unique scale and graph structure of our CRN dataset is crucial.

In summary, ReactionAtlas demonstrates a viable, efficient and accurate route to large-scale systematic exploration and insightful analysis of reactions across many chemical application domains.

\section{Online Methods}
\label{sec:methods}

ReactionAtlas grows a chemical reaction network (CRN) from a small seed set by two coupled loops that read from and write to a shared graph: a \emph{generative loop} that proposes intermolecular reactions and a \emph{potential-energy-surface (PES) loop} that explores transitions around each known minimum.
Both loops use a machine-learned force field (MLFF)\cite{unke2021b} and both update the CRN. The current version of the CRN is used to periodically compute a kinetic model whose steady-state concentrations steer where the exploration is focusing resources.
Full algorithmic detail including all (hyper)parameter tables are given in Supplementary Information section~\ref{si:generative}--\ref{si:seed}.

\paragraph{Single-ended generative transition-state proposer.}
The proposer is a denoising diffusion model \cite{sohl-dickstein2015,ho2020,song2021} in the equivariant, joint-time (JT) MoreRed formulation with time step prediction \cite{kahouli2025,kahouli2024}, trained on the $\omega$B97X-D3 transition states (TSs) of the Grambow dataset \cite{grambow2020b}.
We use a variance-preserving Gaussian schedule of length $T=1000$ with a polynomial noise profile \cite{hoogeboom2022}, $\bar{\alpha}_t = (1-2s)\,(1-(t/T)^p)^2 + s$ with $p=2$, $s=10^{-5}$. Unlike for image generation, the proposer does not start from pure noise: a reactant geometry $\mathbf{x}_0$ (one minimum, or two minima merged into an encounter complex) is partially noised to a level $t\!\sim\mathcal{U}(100,650)$ as $\mathbf{x}_t = \sqrt{\bar{\alpha}_t}\,\mathbf{x}_0 + \sqrt{1-\bar{\alpha}_t}\,\boldsymbol{\epsilon}$ and then denoised by the learned reverse process
\begin{equation}
    \mathbf{x}_{t-1} = \tfrac{1}{\sqrt{\alpha_t}}\Bigl(\mathbf{x}_t - \tfrac{1-\alpha_t}{\sqrt{1-\bar{\alpha}_t}}\,\boldsymbol{\epsilon}_\theta(\mathbf{x}_t,t)\Bigr) + \sigma_t\,\mathbf{z}, \qquad \mathbf{z}\sim\mathcal{N}(0,\mathbf{I}),
\label{eq:ddpm}
\end{equation}
which samples a TS candidate.
The model is single-ended in the strict sense, i.e., no product, driving coordinate, or reaction template is supplied. Candidate reactants are drawn from the current network with probability proportional to a softmax over the base-ten logarithm of their steady-state concentrations (below), so the proposer concentrates on the kinetically relevant part of the network.

\paragraph{MLFF critic and saddle-point validation.}
Every candidate is scored by MD-ET\cite{eissler2026simple}, a transformer force field trained on the QCML dataset\cite{ganscha2025}.
MD-ET predicts energies and forces from independent heads and derives Hessians by automatic differentiation of the force head. Since MD-ET is not energy-conserving, Hessian predictions produced by it are not always perfectly symmetric. For convenience we symmetrize the predicted Hessians and discard the asymmetric part (see SI \ref{sec:shared-mdet}).
From the Hessian matrices $\mathbf{H}$ thus obtained we remove the six translation/rotation modes by projection, $\mathbf{H}_\text{proj} = \mathbf{P}\,\mathbf{H}\,\mathbf{P}$ with $\mathbf{P} = \mathbf{I} - \sum_{k=1}^{6}\mathbf{v}_k\mathbf{v}_k^{\top}$ and examine its significantly negative eigenvalues. A candidate further has to connect two distinct minima at the ends of its intrinsic reaction coordinate (IRC\cite{fukui1981}) and to have forward and backward barriers within physical bounds before it enters the graph (Supplementary Information section~\ref{si:generative}).

\paragraph{PES loop: molecular dynamics and saddle point optimization.}
For each new minimum above a concentration threshold, a short Langevin trajectory at \SI{300}{K} samples nearby configurations,
and frames sampled from it, seed a batched partitioned rational-function optimization (P-RFO\cite{cerjan1981,banerjee1985,baker1986,bofill1994}). 
Diagonalising the projected Hessian into modes $\{\lambda_i,\mathbf{u}_i\}$ with gradient projections $g_i = \mathbf{u}_i\!\cdot\!\nabla E$, each step follows
\begin{equation}
    p_i = \frac{\mu_i}{g_i}, \qquad \mu_i = \frac{\lambda_i \pm \sqrt{\lambda_i^2 + 4g_i^2}}{2},
\label{eq:prfo}
\end{equation}
taking the $+$ root on the followed mode (walk uphill) and the $-$ root on all others (relax downhill), under a partitioned trust region. 
Minimizations that do not need curvature use FIRE\cite{bitzek2006fire}. 
Each validated saddle point is followed downhill along its imaginary mode $\hat{\mathbf{u}}_\text{TS}$ from $\mathbf{R}^{\pm} = \mathbf{R}_\text{TS} \pm \Delta x\,\hat{\mathbf{u}}_\text{TS}$ to its two endpoints; transitions between conformers of one compound become edges of that compound's PES graph, while transitions that change the bonding pattern are registered as reactions. Minima are deduplicated by permutation-invariant Kabsch RMSD\cite{kabsch1976} with a nudged-elastic-band (NEB) same-basin test\cite{mills1995,jonsson1998,henkelman2000climbing}.

\paragraph{Kinetic exploration guidance.}
After each accepted reaction the CRN is integrated as a mass-action system,
\begin{equation}
    \frac{d\mathbf{y}}{dt} = \mathbf{S}_\text{net}\,\mathbf{r}(\mathbf{y}), \qquad r_j(\mathbf{y}) = k_j^{+}\!\prod_i y_i^{\nu_{ij}^{-}} - k_j^{-}\!\prod_i y_i^{\nu_{ij}^{+}},
\label{eq:massaction}
\end{equation}
with $\mathbf{S}_\text{net}$ the net stoichiometry matrix, integrated to a long-time steady state with a stiff implicit solver.
Rate constants follow Eyring transition-state theory built on the computed barriers\cite{eyring1935,evans1935},
\begin{equation}
    k = \min\!\Bigl(\tfrac{k_B T}{h}\exp\!\bigl(-E_a/k_B T\bigr),\ k_\text{diff}\Bigr),
\label{eq:eyring}
\end{equation}
capped at the diffusion limit $k_\text{diff}$.
The resulting steady-state composition both defines the sampling distribution for the proposer and provides the microkinetic read-out of the network\cite{gao2016a,hoops2006,goodwin2018}.

\paragraph{Dataset, Dataset Viewer and Code}
\label{sec:dataset}

 We provide the full CRN as a structured dataset. It comprises 12{,}542 unique compounds spanning 191 distinct molecular formulas (6{,}870 C$_4$, 4{,}309 C$_5$, 1{,}187 C$_3$, 148 C$_2$, 22 C$_1$, 6 C$_0$) and ranging in charge from $-3$ to $+2$ (9{,}122 neutral, 3{,}228 singly anionic, 178 doubly anionic, 9 triply anionic, 4 singly cationic, 1 doubly cationic). It further contains 33{,}065 PES minima, averaging 2.64 conformers per compound and reaching 42 for the most flexible (\texttt{O=CO[CH-]CO}); 24{,}134 intra-PES TSs connecting conformers of the same compound; and 44{,}657 elementary TSs collapsing to 21{,}008 unique chemical reactions (by reactant--product identity).
For every reaction we include the full ML-predicted reaction path:
the TS, the IRC trajectory, and the connected reactant and product minima. 
For every reactive TS, we recompute the barrier and the absolute energies of TS and connected minima at the PBE0/def2-TZVPP level\cite{adamo1999,weigend2005} with Orca\cite{orca} and pySCF\cite{pyscf}, and provide the corresponding Hessian.
Where DFT indicates the ML-predicted TS is invalid, we additionally provide the corrected PBE0 TS, the energy difference, and the RMSD to the ML prediction. 
The dataset is available at \url{reactionatlas.bifold.berlin/downloads}.

We also serve a web application at \url{reactionatlas.bifold.berlin} for browsing the full dataset.
The code for the exploration algorithm and database is available at \url{github.com/mx-e/reaction-atlas}.

\section*{Acknowledgments}
We thank Laure Ciernik for the help with some code and general discussion.
S.G. was supported by the Postdoc.Mobility fellowship by the Swiss National Science Foundation (project no. 225476).
S.G., K.K., M.E., and K.-R.M. acknowledge support by the German Ministry of Education and Research (BMBF) for BIFOLD (01IS18037A). Further, this work was in part supported by the BMBF under Grants 01IS14013A-E, 01GQ1115, 01GQ0850, 01IS18025A, 031L0207D, and 01IS18037A. KRM was partly supported by the Institute of Information \& Communications Technology Planning \& Evaluation (IITP) grants funded by the Korea government (MSIT) (No. 2019-0-00079, Artificial Intelligence Graduate School Program, Korea University and No. 2022-0-00984, Development of Artificial Intelligence Technology for Personalized Plug-and-Play Explanation and Verification of Explanation).

\section*{Contributions}
S.G. led the project and had a major role across nearly all parts of the work: conceptualization, data curation, formal analysis, investigation, methodology, project administration, software, supervision, validation, visualization, writing—original draft and writing—review and editing. M.E. helped lead the project and had a critical role in major parts of the work, in particular the software and visualization: conceptualization, data curation, formal analysis, investigation, methodology, software, validation, visualization, writing—original draft and writing—review and editing. K.-R.M. supervised the project, helped shape its ideas and provided resources and funding: conceptualization, funding acquisition, investigation, project administration, resources, supervision, validation, writing—original draft and  writing—review and editing. K.K. contributed to specific technical aspects of the project: data curation, formal analysis, methodology, software and writing—review and editing.

\newpage

\putbib[bib]            %
\end{bibunit}

\clearpage
\setcounter{section}{0}
\setcounter{subsection}{0}
\setcounter{figure}{0}
\setcounter{table}{0}
\setcounter{equation}{0}
\renewcommand{\thesection}{S\arabic{section}}
\renewcommand{\thesubsection}{S\arabic{section}.\arabic{subsection}}
\renewcommand{\thesubsubsection}{S\arabic{section}.\arabic{subsection}.\arabic{subsubsection}}
\renewcommand{\thefigure}{S\arabic{figure}}
\renewcommand{\thetable}{S\arabic{table}}
\renewcommand{\theequation}{S\arabic{equation}}
\renewcommand{\theHsection}{S.\arabic{section}}
\renewcommand{\theHsubsection}{S.\arabic{section}.\arabic{subsection}}
\renewcommand{\theHsubsubsection}{S.\arabic{section}.\arabic{subsection}.\arabic{subsubsection}}
\renewcommand{\theHfigure}{S.\arabic{figure}}
\renewcommand{\theHtable}{S.\arabic{table}}
\renewcommand{\theHequation}{S.\arabic{equation}}
\setlength{\parskip}{6pt}
\setlength{\parindent}{0pt}
\begin{center}
  {\LARGE\bfseries Supplementary Information}\\[6pt]
  {\large\itshape ReactionAtlas: Ab origine exploration of chemical reaction networks with machine learning}
\end{center}
\vspace{1em}

\newcounter{refoffset}\setcounter{refoffset}{113}
\makeatletter
\apptocmd{\thebibliography}{\global\c@NAT@ctr=\value{refoffset}\relax}{}{}
\makeatother

\begin{bibunit}

\section{Detailed related work}
\label{app:ts_methods_sec}

Mapping a chemical reaction network (CRN) means enumerating the relevant species together with the elementary reactions that connect them, each passing through a transition state (TS). The idea is general and has a long history.
Rule-based reaction-mechanism generation originated in combustion modeling, where Broadbelt \textit{et al.} developed the \textsc{NetGen} pipeline\cite{broadbelt1994,broadbelt1996,broadbelt2005}, building hydrocarbon-pyrolysis networks of $10^3$--$10^4$ reactions.
In synthetic chemistry, Grzybowski's group built the original network representation of organic-synthesis rules\cite{fialkowski2005,gothard2012,kowalik2012}, which grew into the \textsc{Allchemy} prebiotic-network work\cite{wolos2020}.
Feinberg's textbook\cite{feinberg2019} provides the mathematical foundation for reaction-network theory (steady states, deficiency theory, persistence), and Dewyer, Arg\"uelles \& Zimmerman\cite{dewyer2018} review the methodology landscape up to 2018; further developments include the mechanism-finding work of Dewyer \& Zimmerman\cite{dewyer2017}, CRN studies of electrocatalytic and bicarbonate systems\cite{garza2018,lees2022}, and a perspective on machine learning across reaction networks\cite{wen2023}.
The same network abstraction also appears in biochemistry, e.g. in the protein-folding networks of Rao \& Caflisch\cite{rao2004} and the conformational transition networks of No\'e \& Fischer\cite{noe2008}.

A common difficulty in all these research directions is deciding which reactions to consider and finding their TS, ideally without knowing the products in advance.
The rest of this section surveys the methods that have been developed for these tasks.

\subsection{Transition state search and exploration frameworks}
\label{si:ts-frameworks}

Table~\ref{tab:ts_methods} compares various TS finding methods to ReactionAtlas, grouped by the prior information each requires.
Local TS optimizers, e.g., eigenvector following,\cite{cerjan1981,banerjee1985}
partitioned RFO,\cite{baker1986}
and Bofill quasi-Newton,\cite{bofill1994}
refine a structure that is already close to the TS and in addition need a ``rearrangement direction'' (mode) to follow, indicating e.g. which bond breaks or where a hydrogen atom will move.
Double-ended path methods, like nudged elastic band (NEB)\cite{mills1995,jonsson1998,henkelman2000climbing} or double-ended growing string method (GSM)\cite{zimmerman2013}, interpolate between a given reactant and a product.
Other methods are single-ended but still need a prior that points the search in a chosen direction, such as a driving coordinate for single-ended GSM\cite{zimmerman2015} (the bond changes defining the target reaction) or an initial mode for the dimer method\cite{henkelman1999dimer}.

Any such method can in principle be made ``explorative'' by combinatorially enumerating all possible prior directions or even reactant-product pairs, which is often infeasible.
The GRRM family\cite{maeda2021} of methods automates this:
ADDF\cite{maeda2005addf} scans the anharmonic distortion directions around a minimum, and AFIR\cite{maeda2011,maeda2011ab} (single-component form SC-AFIR\cite{maeda2014scafir}) scans atom or fragment pairs to react together.
However the number of pairs to scan grows combinatorially with molecule size. GRRM is, moreover, available only as proprietary licensed software.
The generative proposer of this work is explorative \textit{without} just enumerating priors but by \textit{sampling} plausible TSs directly from the empirical distribution of TSs, which is intractably large.

\begin{table}[!ht]
\centering
\caption{TS search and optimization methods, organized by the prior information needed.
``Explorative'' marks methods that locate TSs from a single minimum without knowing the product or a reaction-specific driving coordinate, and can discover new reactions.}
\label{tab:ts_methods}
\footnotesize
\setlength{\tabcolsep}{4pt}
\begin{tabular}{p{2.2cm}p{2.4cm}p{3.1cm}cl}
\toprule
\textbf{Method} & \textbf{Class} & \textbf{Required prior} & \textbf{Explor.} & \textbf{Ref.} \\
\midrule
Eigenvector following & Local saddle-point opt. & Guess near the TS; unstable mode to follow & $\times$ & \cite{cerjan1981,banerjee1985} \\
\addlinespace
Partitioned RFO (P-RFO) & Local saddle-point opt. & Guess near the TS; (approximate) Hessian & $\times$ & \cite{baker1986} \\
\addlinespace
Bofill quasi-Newton update & Local saddle-point opt. & Guess near the TS; Hessian updated on the fly & $\times$ & \cite{bofill1994} \\
\addlinespace
Nudged elastic band (NEB) & Double-ended & Reactant \emph{and} product & $\times$ & \cite{mills1995,jonsson1998,henkelman2000climbing} \\
\addlinespace
Growing string (DE-GSM) & Double-ended & Reactant \emph{and} product & $\times$ & \cite{zimmerman2013} \\
\addlinespace
Growing string (SE-GSM) & Single-ended, driven & One minimum + a driving coordinate (the target bond changes) & $\times$ & \cite{zimmerman2015} \\
\addlinespace
Dimer method & Single-ended, mode-following & One structure + an initial mode/axis biasing the search toward a chosen saddle point & $\times$ & \cite{henkelman1999dimer} \\
\addlinespace
GRRM (AFIR\,/\,ADDF) & Single-ended, automated scan & One minimum; combinatorial scan over fragment pairs or distortion directions & $\checkmark$ & \cite{maeda2005addf,maeda2011,maeda2011ab,maeda2014scafir,sameera2016,maeda2021} \\
\addlinespace
\textbf{ReactionAtlas (this work)} & Generative single-ended & One minimum & $\checkmark$ & --- \\
\bottomrule
\end{tabular}
\end{table}

\begin{table}[!ht]
\centering
\caption{CRN exploration frameworks, grouped by how new reactions are proposed and which TS engine is used.
``Heuristics-free'' marks frameworks that propose reactions without functional-group, retrosynthetic, or template rules.}
\label{tab:crn_frameworks}
\footnotesize
\setlength{\tabcolsep}{4pt}
\begin{tabular}{p{2.6cm}p{4.5cm}p{2.2cm}cl}
\toprule
\textbf{Framework} & \textbf{How reactions are proposed} & \textbf{TS engine} & \textbf{Heur.-free} & \textbf{Ref.} \\
\midrule
Heuristics-guided / context-driven (Reiher) & Reactive-complex assembly from conceptual-DFT functional-group rules; context-driven graph traversal (demonstrated on formose) & SE-GSM + Bofill & $\times$ & \cite{bergeler2015,simm2017b} \\
\addlinespace
Chemoton 2.0 (SCINE) & Newton-trajectory elementary-step searches (NT1/NT2) replacing the templates, but still needing reactive-site identification & Newton trajectories & $\sim$ & \cite{unsleber2022} \\
\addlinespace
Heuristics-aided QC (Rappoport) & Polar, Baldwin, and pericyclic rule templates over a growing graph & DE refinement & $\times$ & \cite{rappoport2014} \\
\addlinespace
YARP (Zhao/Savoie) & Bond-electron-matrix permutations enumerate single-step graph changes & DE-GSM & $\checkmark$ & \cite{zhao2023} \\
\addlinespace
YAKS (Woulfe/Savoie) & YARP enumeration $+$ on-the-fly microkinetic ranking via an MLFF surrogate before DFT & DE-GSM & $\checkmark$ & \cite{wen2024yaks} \\
\addlinespace
Nanoreactor / hyperreactor (Mart\'inez; Ochsenfeld) & High-temperature AIMD in a periodically compressed sphere; products from bond-order trajectories (hyperreactor: softer wall, applied to formose) & none (no barriers) & $\checkmark$ & \cite{wang2014,wang2016,stan2022,stan-bernhardt2024} \\
\addlinespace
Reactive MLFF dynamics (Smith; Benayad; Huet) & General or active-learned reactive neural-network potentials driving condensed-phase MD & none (dynamics) & $\checkmark$ & \cite{zhang2024ani1xnr,benayad2024,huet2024,tiwary2024} \\
\addlinespace
\textbf{ReactionAtlas (this work)} & Diffusion-based single-ended TS proposer + MLFF critic; kinetics-weighted compound sampler & generative SE + P-RFO & $\checkmark$ & --- \\
\bottomrule
\end{tabular}
\end{table}

While Table~\ref{tab:ts_methods} summarizes the saddle-point search itself, we discuss now the existing CRN exploration frameworks (Table~\ref{tab:crn_frameworks}).

A mature first-principles CRN exploration framework is SCINE from the Reiher group.
\cite{baiardi2022}
The original heuristics-guided approach\cite{bergeler2015} and the context-driven extension\cite{simm2017b}\cite{simm2019} (against which we compare the formose cycle in the main paper) assemble reactive complexes from rule-based functional-group templates.
In \textsc{Chemoton 2.0}\cite{unsleber2022} those templates are replaced with two Newton-trajectory algorithms (NT1, NT2) but the process still requires reactive-site identification at every step.
SCINE further provides \textsc{Pathfinder}\cite{turtscher2022} for ranking compound availability and the \textsc{Steering Wheel}\cite{steiner2024} adds interactive steering.
Both are used post-hoc on an already explored CRN, so they are omitted from Table~\ref{tab:crn_frameworks}.

\textsc{YARP}\cite{zhao2023,zhao2023rgd1} (Yet Another Reaction Program, Savoie group) avoids the rule templates by enumerating all single-step bond-electron-matrix permutations, and then uses DE-GSM on each pair or reactant and product.
The follow-up \textsc{YAKS}\cite{wen2024yaks} adds micro-kinetics:
computationally inexpensive MLFF energies pre-screen candidates by their predicted  kinetic contribution before a DFT calculation is launched.
Further autonomous exploration programs include the multicomponent AFIR search for $A+B$ reactions\cite{maeda2011ab}, the AutoMeKin package\cite{martineznunez2021}, and deep reaction-network exploration at heterogeneous catalytic interfaces\cite{zhao2022}.

A second family of methods choses a different approach: 
The \emph{ab initio nanoreactor}\cite{wang2014,wang2016} puts reactive species in a periodically contracting spherical wall (the reactor) at high temperature, sampling reactions directly.
Later versions\cite{stan2022,stan-bernhardt2024} relax the wall geometry and apply this to formose specifically, and general-purpose reactive MLFFs\cite{zhang2024ani1xnr,benayad2024,huet2024} have emerged to model bond breaking.
These methods sample reactivity without proposing edges, so the resulting CRNs often lack barriers (and thus rate constants) and must be paired with a later TS search for kinetic modeling.

Recent diffusion- and flow-matching TS generators\cite{kim2024,duan2023,duan2024,schlama2026} predict TSs from molecular graphs or reactant-product pairs.
However, the double-ended formulation requires the product to be specified at inference and thus cannot drive open-ended exploration.

ReactionAtlas differs from all of these in a way critical for exploration:
it proposes TSs without enumerating or manually assuming any prior.
Rule-based frameworks need reaction templates (e.g. carbonyl chemistry, etc.) while double-ended classical and generative methods need the reaction product and the single-ended methods need a driving coordinate.
ReactionAtlas instead generates a TS directly from one reactant.

Table~\ref{tab:datasets} summarizes how each compared network was constructed and which chemical degrees of freedom it resolves. The datasets differ widely in how reactions are obtained -- from heuristic graph rules to brute-force enumeration to reactive MD -- and most resolve either charge or stereochemistry but not both, which is why we excluded both in the primary cross-dataset coverage analysis. 
However, ReactionAtlas can resolve both which is why including either degree of freedom widens the coverage asymmetry (Figure~\ref{sfig:charge_stereo_inclusive}).

\begin{table}[!ht]
\centering
\caption{CRN methods compared with ReactionAtlas across several axes of chemical resolution (cf.\ Figure~\ref{fig:formose_results}c). ``Charge'' marks methods that resolve charged species (anions, cations, zwitterions); ``Stereo'' marks those that retain stereochemistry (R/S, E/Z). ``\textit{Ab origine}'' marks networks grown by exploration from seed molecules, rather than presupposed or obtained by static enumeration. ``Heuristics-free'' marks those that propose reactions without functional-group or chemical-intuition rule templates. Our network is the only one that satisfies all four.}
\label{tab:datasets}
\footnotesize
\setlength{\tabcolsep}{4pt}
\begin{tabular}{p{2.3cm}p{4.0cm}ccccl}
\toprule
\textbf{Dataset / study} & \textbf{How reactions are obtained} & \rotatebox{90}{\textbf{Charge}} & \rotatebox{90}{\textbf{Stereo}} & \rotatebox{90}{\textbf{\textit{Ab origine}}} & \rotatebox{90}{\textbf{Heuristics-free}\,} & \textbf{Ref.} \\
\midrule
Simm \& Reiher & Heuristic graph rules: functional-group templates (aldol, retro-aldol, enediol tautomerisation) drive trial reactions; neutral by design & $\times$ & $\checkmark$ & $\checkmark$ & $\times$ & \cite{simm2017b} \\
\addlinespace
Rappoport \textit{et al.} & Heuristics-aided quantum chemistry: Baldwin, polar and pericyclic rule templates & $\times$ & $\times$ & $\checkmark$ & $\times$ & \cite{rappoport2014} \\
\addlinespace
Zhao \& Savoie (YARP) & Bond-electron-matrix permutations enumerate single-step graph changes; double-ended GSM locates each TS & $\times$ & $\times$ & $\checkmark$ & $\checkmark$ & \cite{zhao2023} \\
\addlinespace
RGD1 (Zhao \textit{et al.}) & Brute-force constitutional enumeration followed by DFT optimization & $\times$ & $\checkmark$ & $\times$ & $\checkmark$ & \cite{zhao2023rgd1} \\
\addlinespace
Grambow \textit{et al.} & Automated PES exploration with single-ended growing-string TS search & $\times$ & $\checkmark$ & $\times$ & $\checkmark$ & \cite{grambow2020b} \\
\addlinespace
Stan-Bernhardt \textit{et al.} & AIMD hyperreactor: walled sphere with radial compression and heating; reactive events post-processed to SMILES & $\checkmark$ & $\times$ & $\checkmark$ & $\checkmark$ & \cite{stan2022,stan-bernhardt2024} \\
\addlinespace
\textbf{ReactionAtlas (this work)} & ML proposer--critic single-ended exploration steered by on-the-fly kinetics & $\checkmark$ & $\checkmark$ & $\checkmark$ & $\checkmark$ & --- \\
\bottomrule
\end{tabular}
\end{table}

\subsection{Machine learning for transition state search}
\label{si:ml-ts}

A large body of work uses machine learning (ML) to speed up classical TS searches (as described in the previous section) rather than to drive exploration. This work is complementary to the open-ended proposal role ReactionAtlas plays.
One line of research accelerates the TS optimization itself with inexpensive models:
kernel- and Gaussian-process-driven saddle-point and NEB searches can reach nearly DFT accuracy with one to two orders of magnitude fewer functional evaluations\cite{pozun2012,garridotorres2019,denzel2018,peterson2016,koistinen2017,meyer2020,born2021}.
Deep learning potentials can predict energies, forces, and Hessians for GSM and NEB approaches\cite{yuan2024,yang2025deeppot}.

A second line of research predicts TS geometries directly, e.g. with deep learning, adversarial learning, or diffusion models\cite{pattanaik2020,makos2021,choi2023,jackson2021,lemm2021,heinen2022}
Some of the strongest, such as Duan \textit{et al.}'s \textsc{React-OT}\cite{duan2024}, reach ${\sim}0.05$~\AA{} median RMSD on Transition1x.\cite{t1x}
However, all such current methods are double-ended or product-conditioned, so they refine or predict the TS of an \emph{already-specified} reaction and cannot, by themselves, propose unknown products.
Beaglehole \textit{et al.}\cite{beaglehole2025} review the field in full.

The MLFF critic in ReactionAtlas, MD-ET\cite{eissler2026simple} and the diffusion model backbone in MoreRed PaiNN\cite{schutt2021} are relatively recent contributions to a long history of MLFFs.
Behler--Parrinello high-dimensional neural-network potentials\cite{behler2007} were among the first scalable form of this idea, followed by descriptor-based kernels\cite{rupp2012,de2016,christensen2020a}.
End-to-end equivariant message-passing networks are now the most common approach:
\textsc{SchNet}\cite{schutt2017b,schutt2018}, \textsc{SpookyNet}\cite{spookynet}, \textsc{PaiNN}\cite{schutt2021}, \textsc{ANI}\cite{smith2017}, \textsc{DimeNet}/\textsc{GemNet}\cite{gasteiger2022}, \textsc{NequIP}\cite{batzner2022}, \textsc{Allegro}\cite{musaelian2023}, \textsc{MACE}\cite{batatia2023}, tensor-field networks\cite{thomas2018}, E(n)-equivariant GNNs\cite{satorras2022}, the \textsc{So3krates} Euclidean transformer and its Euclidean fast-attention variant\cite{so3krates,efa}, transferable fragment-trained models\cite{unke2024}, the \textsc{So3LR} long-range foundation model\cite{so3lr}, and global \textsc{sGDML}\cite{chmiela2017,chmiela2019,chmiela2023} together with efficient interatomic descriptors for extended molecules\cite{kabylda2023}.
Reviews of the field are provided e.g. by No\'e \textit{et al.}\cite{noe2019b}, the force-field review of Unke \textit{et al.}\cite{unke2021b}, the perspective of Poltavsky and Tkatchenko\cite{poltavsky2021}, and the broader machine-learning-for-chemistry review of Keith \textit{et al.}\cite{keith2021}.

Any of these architectures could in principle be used as the MLFF or as the backbone network of a generative TS proposer for ReactionAtlas. We chose MD-ET\cite{eissler2026simple} because of its exceptionally fast and reliable Hessian predictions.

\subsection{Kinetic modeling}
\label{si:kinetics-lit}

ReactionAtlas has to decide where in the network the next exploration should be started. For this, the compound sampler weights candidate reactants by their steady-state concentration in a mass-action microkinetic model of the current network (the solver itself is detailed in SI-\ref{si:kinetics}).
In other words, a kinetic simulation is run using the reactions discovered thus far, where high barriers are crossed rarely and low barriers are crossed often. When this simulation converges, the resultant concentrations are used for the sampling weights.

The theory for this way of modeling reactions traces back to Eyring's absolute-rate-constant expression\cite{eyring1935} and the Bell-Evans-Polanyi relation between barrier and reaction enthalpy\cite{evans1935}, which allow us to translate the reactive barriers we compute into rate constants and, ultimately, ODE coefficients. Standard frameworks for solving such ODE systems include \textsc{COPASI}\cite{hoops2006}, \textsc{Cantera}\cite{goodwin2018}, \textsc{RMG}\cite{gao2016a}, and the rate-rule infrastructure of Susnow \textit{et al.}\cite{susnow1997} and Han \textit{et al.}\cite{Han2017}.
Sensitivity-analysis and uncertainty-propagation methodology is reviewed by Tur\'anyi\cite{turnyi2014}, with chemistry-specific extensions by Proppe \& Reiher\cite{proppe2017,proppe2019b} and quantum-mechanical rate-coefficient extraction by Suleimanov \textit{et al.}\cite{suleimanov2015}.
Rate-coefficient prediction can also be achieved using ML:
Komp \& Valleau\cite{komp2020,komp2022} predict quantum rate constants directly and Spiekermann \textit{et al.}\cite{spiekermann2022a,spiekermann2022b} reach near-coupled-cluster accuracy on activation energies and publish the \textsc{RDB7} benchmark dataset. Lei \textit{et al.}\cite{lei2024} apply the same idea inside production combustion mechanisms.

\subsection{Origins-of-life context}
\label{si:ool}

One broader scientific motivation for mapping prebiotic CRNs comes from the origins-of-life literature. Foundational treatments include Lazcano's historical account of the field\cite{lazcano2010}, the emergence perspective of Ruiz-Mirazo \textit{et al.}\cite{ruiz-mirazo2014} and the chemical evolution of Sutherland\cite{sutherland2016}.
Xavier \textit{et al.}\cite{xavier2020} formalize auto-catalytic CRNs as the origin of metabolism while experimentally, nonenzymatic analogues of core metabolic reactions have been found \textit{in vitro}.\cite{muchowska2020}. Such carbohydrate chemistry is the basis of the RNA-world hypothesis\cite{atkins2011,orgel2004,powner2009,szostak2001}, and experiments have shown that prebiotic CRNs can self-organize under changing environmental conditions\cite{robinson2022,vanduppen2023,baltussen2024}.
For a recent broader overview of self-organization in prebiotic chemistry see Moldogazieva \textit{et al.}\cite{moldogazieva2026}.
A general review of automated prebiotic CRN exploration was written by Sharma \textit{et al.}\cite{sharma2021}.

Beyond prebiotic chemistry CRN methods are widely applied in catalysis\cite{deutschmann1998,zhu2005,gossler2019}, combustion\cite{harper2011}, atmospheric chemistry\cite{upadhyay2023}, and astrochemistry\cite{wakelam2012}.

\newpage

\section{Detailed Results}

\subsection{Quality and Speed}
\label{si:quality}
This section gives additional details on the quality and speed evaluations referenced in the main text, section \ref{sec:results}. Quality measures how many of the recovered TS states are a valid TS (i.e. a first order saddle point) according to some reference potential, e.g. for our work PBE0/def2-TZVPP. 
We assess TS quality at two levels:
across the whole network, and structure by structure against standard TS search methods on a given sample of structures.

For all found TSs in the network, we obtain a single-point Hessian evaluation using PBE0/def2-TZVPP with PySCF\cite{pyscf} and classify each TS as valid TS when its Hessian had one imaginary mode (tolerance $|\nu|>100$~cm$^{-1}$). Table~\ref{tab:network_quality} breaks the valid subset down by reaction class: isomerization (1$\to$1, one molecule transforms to one other molecule), fragmentation (1$\to$2) and all others.

 Reactions found by the PES-escape loop (SI-\ref{si:pes}) are almost always already valid (96\% for 1$\to$1 and 97\% for everything else). The generative loop (SI-\ref{si:generative}), which provides most multi-component reactions, has a lower success rate (44\% for 1$\to$1 and 38\% for everything else) which is in part a consequence of a more lenient approach to ML-based validation we adopt here since it provides exploration speed. The reliable post-hoc optimization described below suggests the generative path still yields an acceptable speed-accuracy tradeoff. Averaging over both loops, validity is 86\% for 1$\to$1 and 41\% for everything else (59\% overall). A single DFT TS optimization seeded at the MLFF geometry increases these to 99\% and 94\% (96\% overall) and notably at a fraction of the cost of locating the TS without a prior ML proposal, as the ML proposal is often in close vicinity of a true TS.

\begin{table}[!ht]
  \centering \small
  \begin{tabular}{l c c c}
    \toprule
        & 1$\to$1 (40\%) & 1$\to$2 (50\%) & other (10\%) \\
    \midrule
    PES escape  & 95.9 (32.8\%) & 96.8 (3.1\%)  &  95.1 (0.3\%) \\
    Generative  & 43.5 (7.4\%)  & 39.4 (46.4\%) &  31.9 (10.0\%) \\
    Average     & 86.2 (40.2\%) & 43.0 (49.5\%) &  33.7 (10.3\%) \\
    \bottomrule
  \end{tabular}
  \caption{Fraction of each reaction group whose MLFF-predicted TS already coincides with a PBE0/def2-TZVPP TS. The percentages in parenthesis give the total share of that class.
  }
  \label{tab:network_quality}
\end{table}

To measure speed we examine the average worker node time ReactionAtlas requires during the main experiment for finding a single TS it deems valid (i.e. before PBE0 validation, using only ML methods, as above) called $\mathbb{E}[t_\text{pred}]$ below. 
Measured thus, the generative loop requires $\approx$25 min/reaction, while the PES loop takes around 18 min/reaction, which gives a weighted average of 22.6 min/reaction.
This is measured for the main experiment which mostly used cloud-based L4 GPU nodes as workers. 
Nodes with faster accelerators (A100, H100, B100, etc.) would consequently be even faster for this work.
The way we measure speed as total time divided by TS found for each loop implies that all other `objects' that come out of the CRN exploration (i.e. reaction paths, ML Hessians on reaction paths, conformational transition paths, deduplicated compound minima, ML energy predictions, etc.) are `free' by-products.

For DFT-corrected TS we calculate the average runtime as $\mathbb{E}[t_\text{DFT}]$, where:
\begin{equation}
    t_{DFT} = \mathbb{E}[t_\text{pred}] +
\begin{cases}
t_\text{val} & \text{if MLFF TS valid}  \\
t_\text{corr} & \text{if MLFF TS invalid}
\end{cases}
\end{equation}
i.e. the TS prediction cost reported above plus the expected cost of upgrading the MLFF saddle to a DFT-validated one for the cases where the MLFF does not predict a valid TS and the cost of checking the TS for all other cases.
We choose to not double-count the cost of a Hessian evaluation as optimization require the Hessian and can be initialized with it.
For isomerizations (1$\to$1 reactions) $\mathbb{E}[t_\text{DFT}]$ is 66.7 minutes, while for other types of reactions producing one PBE0 level TS requires 100.2 minutes on average. 
Note we do \textit{not} include failed PBE0 optimizations (which are rare, see quality results above) into this average. While $\mathbb{E}[t_\text{pred}]$ is expended on an GPU node (as described above), the remaining DFT optimization is computed using a 4-core CPU node.

\subsubsection{Comparison to double-ended methods}

Comparing the single-ended TS search of ReactionAtlas to other (double-ended) TS-search methods is not straightforward. 
On the one hand, ReactionAtlas quickly samples and discards many candidates before finding a TS connecting to an input reactant. 
Especially for 2$\to$1 and 2$\to$2 reactions ReactionAtlas takes many trials before finding a valid TS. 
On the other hand, double-ended methods need to be provided with a reactant--product pair. If there is no (or no sensible) TS between these reactant and product geometries, which is the case for most of these pairs, double-ended methods will fail. 

We thus construct the best possible but still imperfect comparison to give context. 
For ReactionAtlas a TS proposal includes the whole process of going through the pipeline from proposal to validation, exclusively based on ML predictions. 
For the double-ended methods we use TSs found by ReactionAtlas, correct them using PBE0 if necessary, relax the reactant and product endpoints, also using PBE0, and use this \textit{valid} reactant--product pairs as input for the TS search to assess quality and speed of the double ended methods. 
Note that only 1$\to$1 reactions can be evaluated in this way since double-ended methods need a valid encounter complex for any reactant or product consisting of multiple compounds which is usually computed using complex heuristics. 
ReactionAtlas solves this issue simply by sampling random encounters when trying to merge multiple reactants. Given these deliberately favorable conditions, the quality scores of the double-ended methods can be seen as best case scenarios favoring the traditional methods.

Table~\ref{tab:method_compare} compares five TS search methods on a set of 100 isomerisation (1$\to$1) reactions sampled randomly from the CRN discovered by ReactionAtlas and prepared as described above. We also provide results for a post-hoc PBE0-driven TS search initalized at the ML-proposed TS. The DFT TS-opt recovers the MLFF TS closely (median RMSD 0.13 \AA, mean $|\Delta E|$ 1.7 kcal/mol), confirming that the ML geometry already sits close to the true TS.

\begin{table}[!ht]
  \centering \small
  \begin{tabular}{l c c c c c}
    \toprule
        & MLFF TS & DFT TS-opt & xTB$\to$DFT & NEB-TS & DE-GSM \\
    \midrule
    $n_\mathrm{imag}=1$ ($n/N$) & 90/100 & 100/100 & 60/63 & 91/98 & 56/66 \\
    \bottomrule
  \end{tabular}
  \caption{TS finding methods on a sample of 100 1$\to$1 reactions. $n$ is the number of valid TSs and $N$ is the number of jobs that produced a candidate TS.
  The double-ended methods use the PBE0 IRC-derived minima of the converged TS-opt as endpoints.
  }
  \label{tab:method_compare}
\end{table}

\begin{figure}[!htbp]
    \centering
    \includegraphics[width=\linewidth, trim=0 0 0 450, clip]{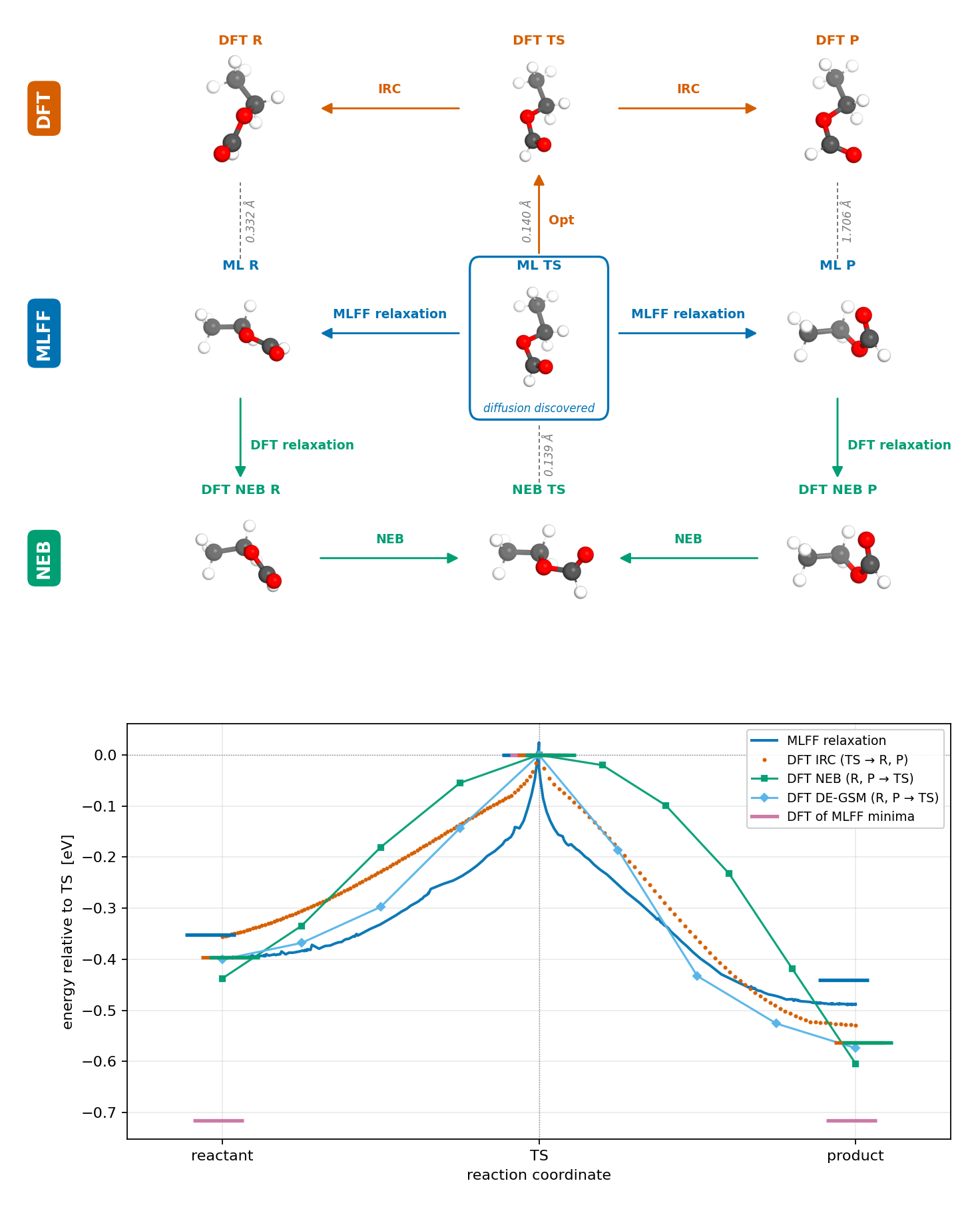}
    \caption{
    Energy profiles along the reaction coordinate relative to the TS for the MLFF relaxation, the DFT IRC, the DFT NEB and the DFT DE-GSM strings, with the DFT energies of the MLFF minima marked; the MLFF, DFT-IRC and DFT-NEB profiles agree closely around the saddle point, whereas DE-GSM tends to overshoot the barrier, consistent with its lower saddle-point yield (Table~\ref{tab:method_compare}).
    }
    \label{sfig:ts_profiles}
\end{figure}

Figure~\ref{sfig:ts_profiles} illustrates, for one representative reaction, how each search locates the TS and its corresponding minima with the energy profile along the reaction coordinate.

Table~\ref{tab:speed} conducts a similarly imperfect speed comparison between ReactionAtlas and the double-ended methods introduced above.
ReactionAtlas' speed is derived as described in the previous section, while we report the mean time elapsed of each successful TS prediction on a 4-core CPU node using the same 100 sample set we use above to determine quality. 
It is important to stress how much this comparison favors the double-ended methods: 
We do not take into account failed optimizations which may not fail outright but rather not converge until a user-defined maximum time has elapsed (in our case 48 hours). 
This usually happens in the majority of cases, since no chemically plausible TS exists between most reactant--product pairs, unlike for the pre-validated set this experiment is conducted on. Using rules-based enumerations of reactant product pairs could therefore increase the time per valid TS found by several orders of magnitude depending on molecule size and complexity of the rule set employed while also introducing chemical biases.

\begin{table}[!ht]
  \centering \small
  \begin{tabular}{l c c}
    \toprule
    Method & 1$\to$1 / min & multi-component / min \\
    \midrule
    MLFF TS (our ReactionAtlas)              & 20  & 25 \\
    DFT TS-opt (weighted) & 67  & 100 \\
    xTB$\to$DFT $\star$   & 261 & --- \\
    NEB-TS $\star$        & 270 & --- \\
    DE-GSM $\star$        & 416 & --- \\
    \bottomrule
  \end{tabular}
  \caption{\textbf{Median wall time per TS found} (four CPU cores).}
  \label{tab:speed}
\end{table}

\subsection{Relevance}
\label{si:relevance}

\subsubsection{Constitutional coverage and the Wikipedia relevance proxy}
\label{app:coverage}

To gauge relevance of the compounds in the discovered CRN we use the set of compounds scraped from English Wikipedia, i.e. those named compounds which have a dedicated Wikipedia article.
Table~\ref{tab:wiki_coverage} compares, at each carbon count, three populations of closed-shell neutral CHO species: 
the full constitutional-isomer enumeration generated with MAYGEN\cite{maygen}, the species present in ReactionAtlas, and those judged ``relevant'' by having a Wikipedia entry.
All three are restricted to the same carbohydrate-range grid ($1\le n_\mathrm{C}\le 5$, $n_\mathrm{O}\le n_\mathrm{C}$, $n_\mathrm{H}\le 2n_\mathrm{C}$
(Table~\ref{tab:wiki_coverage}).

ReactionAtlas recovers almost all named species at low carbon count (19/20 at C$_2$, the sole miss being the strained four-membered ring 1,3-dioxetane) and stays high through C$_3$ and C$_4$ (84.5\% and 82.9\%), even as the constitutional universe grows from 31 to 13,571 isomers and the named fraction falls below 1\%. 

\begin{table}[!ht]
  \centering
  \caption{Coverage of the carbohydrate-range CHO grid ($1\leq n_\mathrm{C}\leq 5$, $n_\mathrm{O}\leq n_\mathrm{C}$, $n_\mathrm{H}\leq 2n_\mathrm{C}$, i.e.\ degree of unsaturation $\geq 1$) by ReactionAtlas (\emph{ours}) versus the full MAYGEN constitutional-isomer enumeration (\emph{enum}) and named compounds with an English Wikipedia entry (\emph{Wiki}). All three populations are filtered to closed-shell neutral CHO species on stereo- and isotope-stripped canonical SMILES, so the columns are directly comparable.}
  \label{tab:wiki_coverage}
  \begin{tabular}{lrrrrrr}
    \toprule
    $n_\mathrm{C}$ & enum & ours & ours/enum & Wiki & Wiki/enum & ours/Wiki \\
    \midrule
    C2 & 31 & 22 & 71.0\% & 20 & 64.5\% & 95.0\% \\
    C3 & 536 & 204 & 38.1\% & 58 & 10.8\% & 84.5\% \\
    C4 & 13,571 & 2,076 & 15.3\% & 111 & 0.8\% & 82.9\% \\
    C5 & 420,312 & 1,994 & 0.5\% & 121 & 0.0\% & 15.7\% \\
    \midrule
    all & 434,450 & 4,296 & 1.0\% & 310 & 0.1\% & 57.7\% \\
    \bottomrule
  \end{tabular}
\end{table}

\begin{figure}[H]
    \centering
    \includegraphics[width=0.72\linewidth]{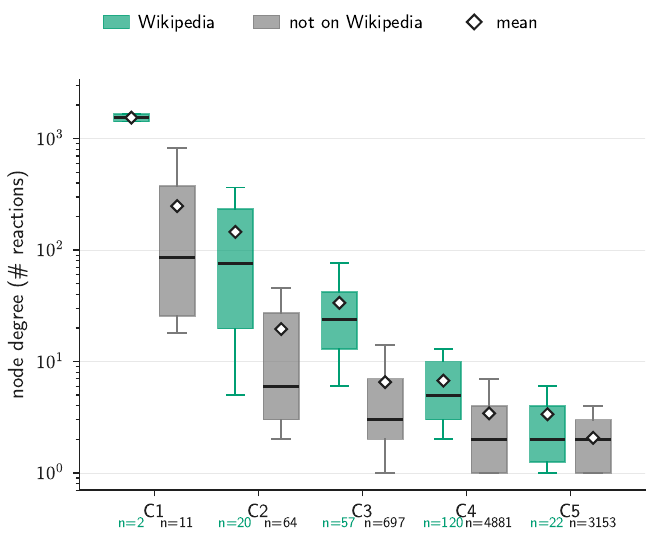}
    \caption{
    Node degree (number of incident reactions) per compound for each carbon count as box plot, separating species with an English Wikipedia entry (green) from those without (gray). Below the x-axis are the counts $n$ in each group.
    }
    \label{sfig:degree}
\end{figure}

Consistent with this picture, Wikipedia-named compounds occupy systematically higher node degrees in the network than unnamed isomers (Fig.~\ref{sfig:degree}). We compare, per carbon count, the node degree (number of incident reactions) of compounds with and without a dedicated English Wikipedia entry.
The Wikipedia-named compounds have systematically higher degrees, confirming that the kinetic sampler concentrates compute on the relevant chemistry.

\subsubsection{Astrochemical coverage}
\label{app:astrochem}
As an additional measure of completeness and relevance we looked for the recovery of compounds detected in interstellar space, the reasoning here being that kinetically stable compounds in interstellar vacuum should match those found by a bias-free CRN which also simulates molecules in vacuum. We compared the CRN against the catalog of closed-shell C/H/O-only neutral species with $\leq 4$ heavy atoms (C + O) detected in the interstellar medium, circumstellar envelopes, or comets, as collected by the McGuire 2022 census~\cite{mcguire2022} and its companion \texttt{astromol} (which integrates JPL~\cite{pickett1998} and CDMS~\cite{muller2005} entries). 
The catalog contains 45 closed-shell CHO neutrals and ReactionAtlas recovers 41 of them, or 91\% (Table~\ref{tab:astro}).
The four misses are all small carbon clusters or singlet cumulene carbenes (C$_2$, C$_3$, propadienylidene H$_2$CCC, butatrienylidene H$_2$C$_4$). Open-shell radicals (CH, OH, HCO, HO$_2$, O$_2$ in its triplet ground state, $\ldots$) and ions (HCO$^+$, H$_3$O$^+$, $\ldots$) detected in the same regime are excluded from the comparison because our diffusion proposer generates closed-shell neutrals, and the only ions present in the network are the seed set.

\begingroup
\small
\setlength{\LTcapwidth}{\textwidth}
\begin{longtable}{lllcl}
\caption{Closed-shell CHO neutral species with $\leq 4$ heavy atoms detected in the interstellar medium, circumstellar envelopes, or comets according to the McGuire 2022 census~\cite{mcguire2022} and the \texttt{astromol} database. ReactionAtlas recovers 41 of 45 (91\%). Open-shell radicals (e.g.\ OH, HCO, O$_2$) and ions (e.g.\ HCO$^+$, H$_3$O$^+$) detected in the same regime are excluded.}
\label{tab:astro}\\
\toprule
\textbf{Species} & \textbf{Formula} & \textbf{$N_\mathrm{heavy}$} & \textbf{Found} & \textbf{ISM source / year} \\
\midrule
\endfirsthead
\multicolumn{5}{l}{\itshape\small Table~\ref{tab:astro} continued} \\
\toprule
\textbf{Species} & \textbf{Formula} & \textbf{$N_\mathrm{heavy}$} & \textbf{Found} & \textbf{ISM source / year} \\
\midrule
\endhead
\midrule
\multicolumn{5}{r}{\itshape\small continued on next page} \\
\endfoot
\bottomrule
\endlastfoot
Molecular hydrogen      & H$_2$            & 0 & $\checkmark$ & $\xi$~Per LOS (1970) \\
Methane                 & CH$_4$           & 1 & $\checkmark$ & NGC 7538 LOS (1991) \\
Water                   & H$_2$O           & 1 & $\checkmark$ & Sgr~B2 (1969) \\
Dicarbon                & C$_2$            & 2 & $\times$     & Cyg OB2 \#12 LOS (1977) \\
Acetylene               & C$_2$H$_2$       & 2 & $\checkmark$ & IRC+10216 (1976) \\
Ethylene                & C$_2$H$_4$       & 2 & $\checkmark$ & IRC+10216 (1981) \\
Methanol                & CH$_3$OH         & 2 & $\checkmark$ & Sgr~A, Sgr~B2 (1970) \\
Carbon monoxide         & CO               & 2 & $\checkmark$ & Orion (1970) \\
Formaldehyde            & H$_2$CO          & 2 & $\checkmark$ & Multiple (1969) \\
Hydrogen peroxide       & HOOH             & 2 & $\checkmark$ & $\rho$~Oph A (2011) \\
Tricarbon               & C$_3$            & 3 & $\times$     & IRC+10216 (1988) \\
Cyclopropenylidene      & c-C$_3$H$_2$     & 3 & $\checkmark$ & Sgr~B2, Orion, TMC-1 (1985) \\
Propadienylidene        & H$_2$CCC         & 3 & $\times$     & TMC-1 (1991) \\
Vinyl alcohol           & CH$_2$=CHOH      & 3 & $\checkmark$ & Sgr~B2 (2001) \\
Ethylene oxide          & c-C$_2$H$_4$O    & 3 & $\checkmark$ & Sgr~B2 (1997) \\
Acetaldehyde            & CH$_3$CHO        & 3 & $\checkmark$ & Sgr~B2 (1973) \\
Propylene               & CH$_2$=CHCH$_3$  & 3 & $\checkmark$ & TMC-1 (2007) \\
Methylacetylene         & CH$_3$CCH        & 3 & $\checkmark$ & Sgr~B2 (1973) \\
Dimethyl ether          & CH$_3$OCH$_3$    & 3 & $\checkmark$ & Orion (1974) \\
Ethanol                 & CH$_3$CH$_2$OH   & 3 & $\checkmark$ & Sgr~B2 (1975) \\
Carbon dioxide          & CO$_2$           & 3 & $\checkmark$ & AFGL LOS (1989) \\
Ketene                  & H$_2$CCO         & 3 & $\checkmark$ & Sgr~B2 (1977) \\
Formic acid             & HCOOH            & 3 & $\checkmark$ & Sgr~B2 (1971) \\
Tricarbon monoxide      & C$_3$O           & 4 & $\checkmark$ & TMC-1 (1985) \\
Cyclopropenone          & c-H$_2$C$_3$O    & 4 & $\checkmark$ & Sgr~B2 (2006) \\
Vinylacetylene          & CH$_2$=CHC$\equiv$CH & 4 & $\checkmark$ & TMC-1 (2021) \\
Acrolein (propenal)     & CH$_2$=CHCHO     & 4 & $\checkmark$ & Sgr~B2 (2004) \\
Glycolaldehyde          & CH$_2$OHCHO      & 4 & $\checkmark$ & Sgr~B2 (2000)~\cite{hollis2000} \\
Propylene oxide         & CH$_3$CHCH$_2$O  & 4 & $\checkmark$ & Sgr~B2 (2016) \\
Methylketene            & CH$_3$CHCO       & 4 & $\checkmark$ & TMC-1 (2023) \\
1-Butyne                & CH$_3$CH$_2$CCH  & 4 & $\checkmark$ & TMC-1 (2024) \\
Propanal                & CH$_3$CH$_2$CHO  & 4 & $\checkmark$ & Sgr~B2 (2004) \\
Acetone                 & (CH$_3$)$_2$CO   & 4 & $\checkmark$ & Sgr~B2 (1987) \\
Acetic acid             & CH$_3$COOH       & 4 & $\checkmark$ & Sgr~B2 (1997) \\
Methoxymethanol         & CH$_3$OCH$_2$OH  & 4 & $\checkmark$ & NGC 6334 (2017) \\
Ethyl methyl ether      & C$_2$H$_5$OCH$_3$ & 4 & $\checkmark$ & Orion (2018) \\
Butatrienylidene        & H$_2$C$_4$       & 4 & $\times$     & IRC+10216 (1991) \\
Propynal                & HC$_2$CHO        & 4 & $\checkmark$ & TMC-1 (1988) \\
Diacetylene             & HC$_4$H          & 4 & $\checkmark$ & CRL 618 (2001) \\
Methyl formate          & HCOOCH$_3$       & 4 & $\checkmark$ & Sgr~B2 (1975) \\
(Z)-Ethenediol          & (Z)-HOCH=CHOH    & 4 & $\checkmark$ & G+0.693$-$0.027 (2022) \\
Ethylene glycol         & HOCH$_2$CH$_2$OH & 4 & $\checkmark$ & Sgr~B2 (2002) \\
Carbonic acid           & HOC(O)OH         & 4 & $\checkmark$ & G+0.693$-$0.027 (2023) \\
$n$-Propanol            & $n$-CH$_3$CH$_2$CH$_2$OH & 4 & $\checkmark$ & G+0.693$-$0.027 (2022) \\
$i$-Propanol            & $i$-(CH$_3$)$_2$CHOH     & 4 & $\checkmark$ & Sgr~B2 (2022) \\
\end{longtable}
\endgroup

\subsubsection{Cross-dataset coverage}
\label{si:pairwise_coverage}

\begin{figure}[H]
    \centering
    \includegraphics[width=0.85\linewidth]{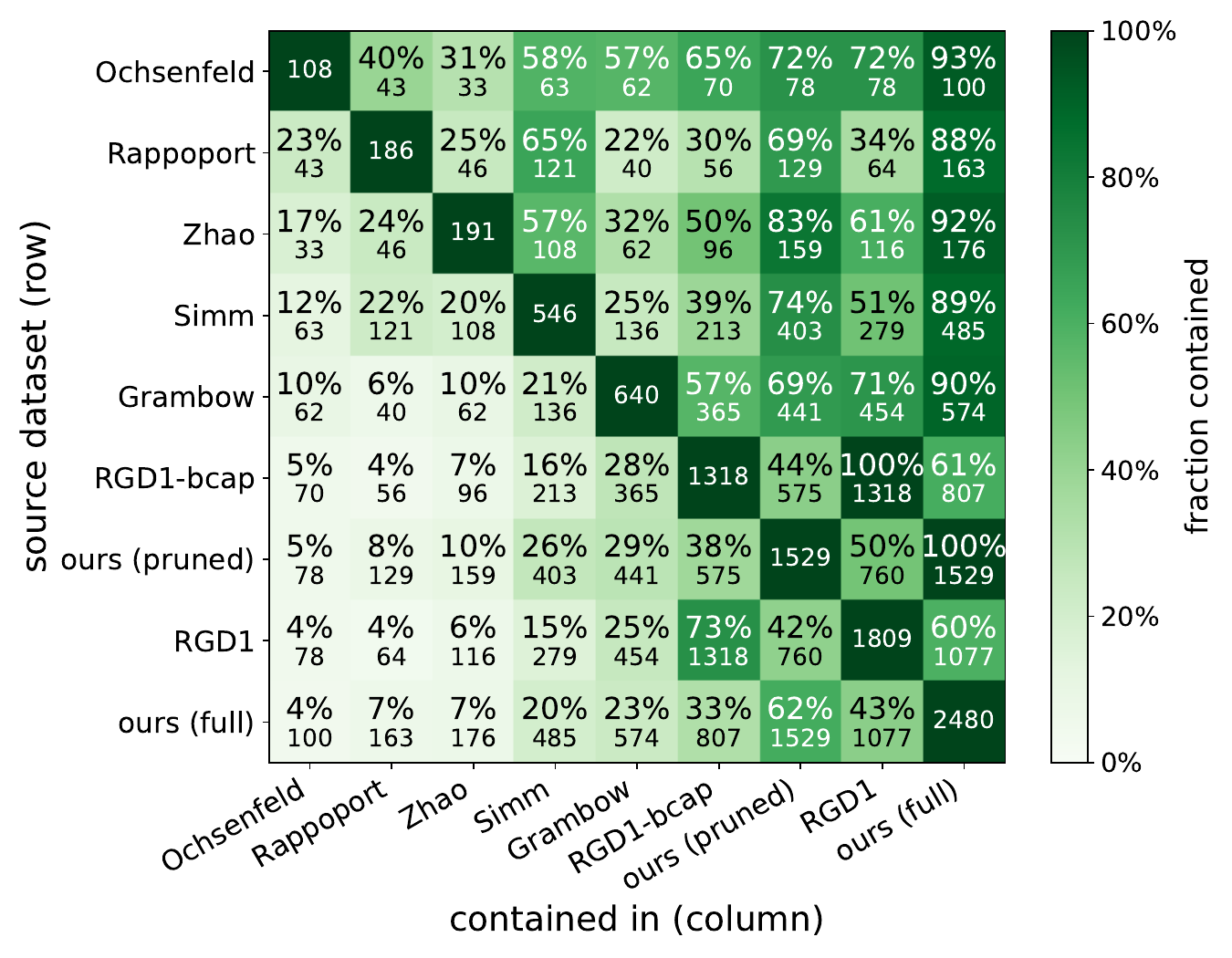}
    \caption{
    Extended coverage matrix with including the reaction database of Grambow \textit{et al.}\cite{grambow2020b}, RGD1-bcap (RGD1\cite{zhao2023rgd1} restricted to reactions with barriers $\leq 200$\,kJ/mol), and crn-pruned (our network under the same barrier constraint and Simm's ``$\leq 1$ non-starting-material reactant'' heuristic, seeds \{CH$_2$O, H$_2$O, glycolaldehyde\}).
    The barrier-pruned variants confirm that the asymmetric coverage is not an artefact of counting kinetically inaccessible species: 
    crn-pruned still covers 74\% of Simm, 69\% of Rappoport, 83\% of Zhao and 69\% of Grambow, and remains $4\times$ the size of the largest dedicated formose comparator.
    Removing the high-barrier subset of RGD1 drops its coverage of crn-cloud from 43\% to only 33\%.
    }
    \label{sfig:coverage_si}
\end{figure}

To place the discovered network in the context of prior work, we computed pairwise asymmetric coverage between ReactionAtlas and published CHO reaction-network datasets, i.e., for each method, we compare the set of unique compounds contained in the explored CRN.
To conduct an apples-to-apples comparison we first restrict each CRNs set of compounds to a common chemistry, i.e., compounds which \textit{could} have been found given the methods capabilities. Thus, unless stated otherwise, every set is restricted to the C$_4$H$_8$O$_4$ envelope ($\leq 4$\,C, $\leq 8$\,H, $\leq 4$\,O), to closed-shell CHO neutrals, and canonicalized to non-isomeric SMILES so that only constitutional isomers contribute. The resulting matrix can be seen below (Figure \ref{sfig:coverage_si}):
The cell at row $i$, column $j$ reports $|D_i \cap D_j| / |D_i|$, the fraction of dataset $D_i$ that is also present in $D_j$.

\begin{figure}[H]
    \centering
    \includegraphics[width=0.5\linewidth]{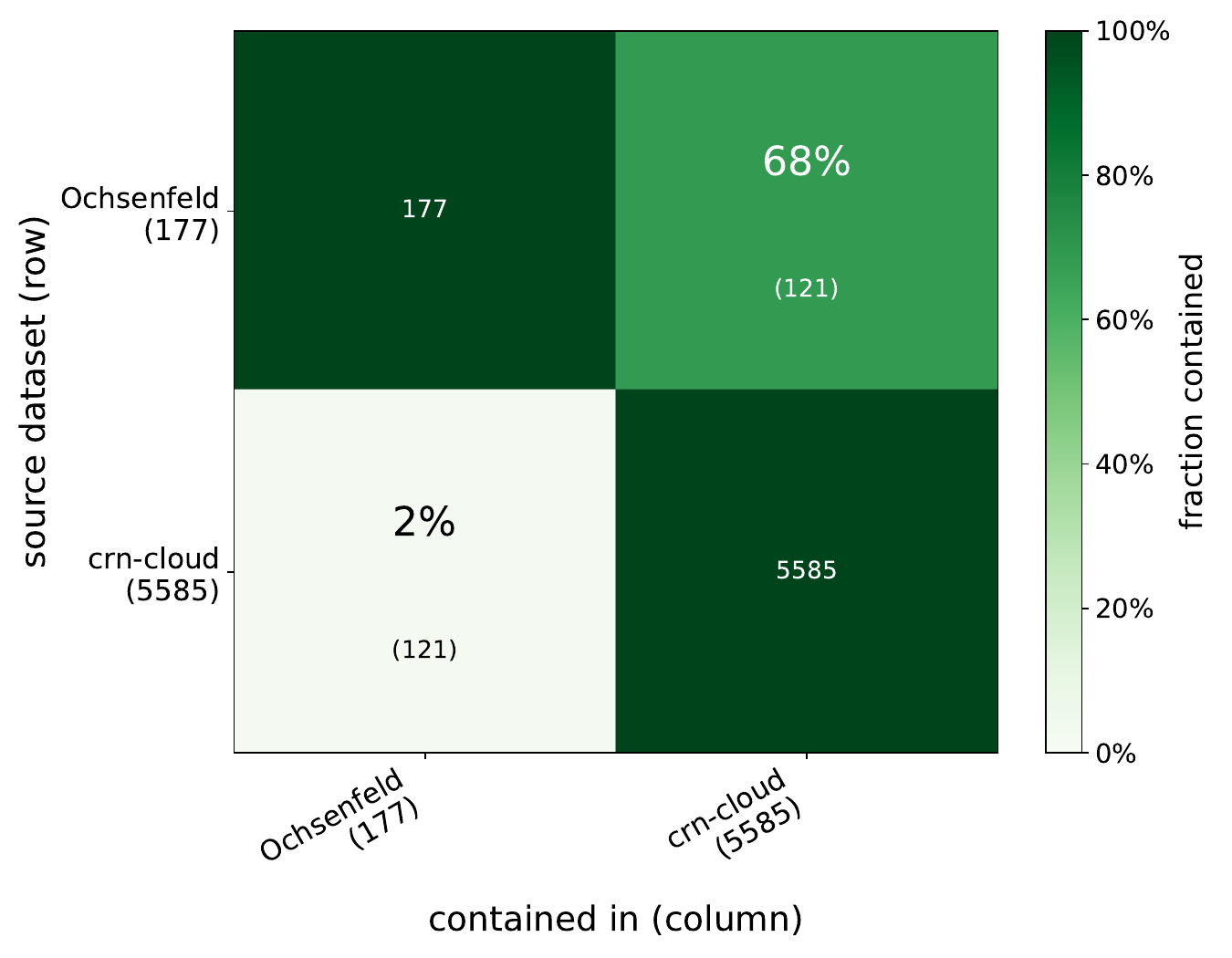}%
    \includegraphics[width=0.5\linewidth]{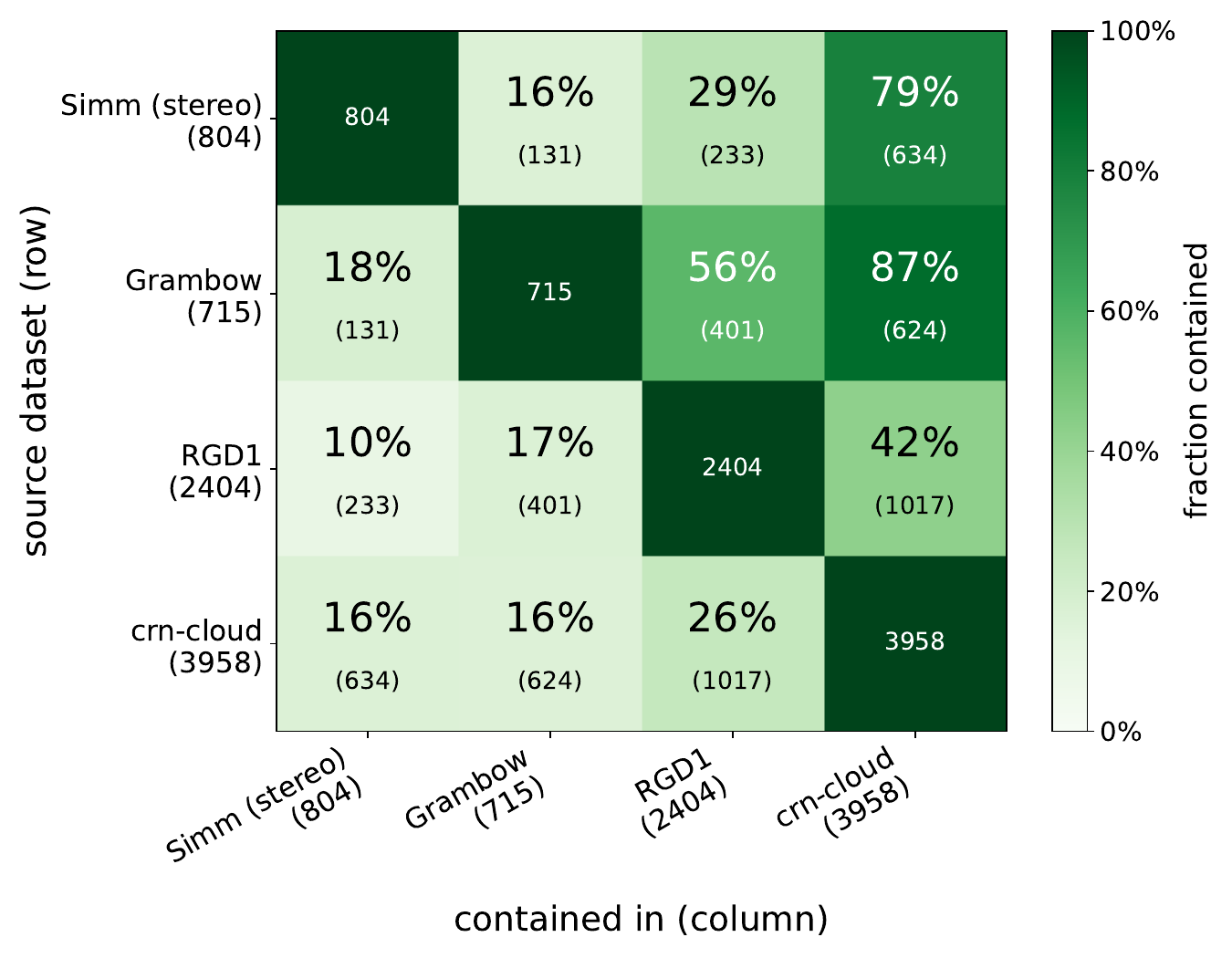}
    \caption{
    \emph{Top:} Charge-inclusive coverage, restricted to the two datasets that publish charged species: the nanoreactor study by Stan \& Ochsenfeld\cite{stan2022} and ReactionAtlas (crn-cloud). Including anions, cations and zwitterions grows ReactionAtlas' set from 2{,}480 to 5{,}585 species and the nanoreactor set from 108 to 177. Coverage of the nanoreactor dataset drops from 93\% (neutrals only) to 68\%, because their nanoreactor trajectories pass through 56 anionic intermediates we do not store; the reverse asymmetry deepens, with 5{,}464 charge-state species in crn-cloud absent from theirs against 56 the other way.
    \emph{Bottom:} Stereo-inclusive coverage, restricted to the datasets where stereochemistry is recoverable. 
   Adding stereochemistry grows the four sets by $+47\%$, $+12\%$, $+33\%$ and $+60\%$ respectively. 
   Off-diagonal literature-versus-literature comparisons should be read descriptively only, as they mix genuine scope differences with stereo-perception heterogeneity between datasets.
    }
    \label{sfig:charge_stereo_inclusive}
\end{figure}

As Table~\ref{tab:datasets} summarized, the datasets differ widely in how reactions are obtained. Most resolve either charge or stereochemistry but not both, which is why we excluded both in the primary cross-dataset coverage analysis. 
However, ReactionAtlas can resolve both which is why including either degree of freedom widens the coverage asymmetry (Figure~\ref{sfig:charge_stereo_inclusive}).

\subsection{Formose refinement}

\subsubsection{Initial conditions for the formose refinement pass}
\label{app:formose_ic}

To increase the resolution of the CRN dataset around the formose cycle, we conducted a refinement search around compounds relevant for formose chemistry (described in  Section~\ref{sec:formose_cycle}).
The refinement pass re-seeded the exploration from the 92 formose-relevant species listed in Table~\ref{tab:formose_ic}. 
For the refinement TS proposals were drawn from this subset of compounds only.
The set covers the canonical aldol intermediates (FA, GO, GA, DHA, the C$_3$ enediols, all six tetrose stereoisomers), their hydration partners (gem-diols), and a representative selection of the charge-separated tautomers.

\begin{footnotesize}
\begin{longtable}{@{}l l@{}}
\caption{The 92 formose-relevant species used to re-seed the refinement pass of Section~\ref{sec:formose_cycle}, given as SMILES and a short label.}\\
\label{tab:formose_ic}\\
\toprule
\textbf{SMILES} & \textbf{Species} \\
\midrule
\endfirsthead
\multicolumn{2}{@{}l}{\footnotesize\itshape Table~\ref{tab:formose_ic} continued from previous page}\\
\toprule
\textbf{SMILES} & \textbf{Species} \\
\midrule
\endhead
\midrule
\multicolumn{2}{r@{}}{\footnotesize\itshape continued on next page}\\
\endfoot
\bottomrule
\endlastfoot
\texttt{[O-]C1(O)OCO1} & 1,3-dioxetane-2,2-diolate \\
\texttt{C/[O+]=[C-]/OCO} & methyl + methylene-acetal \\
\texttt{C/[O+]=C\textbackslash{}O[CH-]O} & methyl + methylenedioxy ylide \\
\texttt{O=CC(O)(CO)CO} & apiose precursor \\
\texttt{[O-]C1OCO1} & cyclo 2-FA anion \\
\texttt{[CH2-]O} & methanide-OH \\
\texttt{C=[O+][CH-]O} & methylenedioxocarbenium ylide \\
\texttt{[O-]CC/[O+]=[C-]\textbackslash{}O} & C3 alkoxide-ylide \\
\texttt{O/C=[O+]/[CH-]CO} & C3 oxocarbenium-ylide (E) \\
\texttt{O/C=[O+]\textbackslash{}[CH-]CO} & C3 oxocarbenium-ylide (Z) \\
\texttt{OC/[C-]=[O+]/[C@H](O)CO} & C4 C-O+ ylide \\
\texttt{O/[C-]=[O+]/C[C@@H](O)CO} & C4 C-O+ ylide (alt) \\
\texttt{[O-]C/C(O)=[O+]\textbackslash{}CCO} & C4 alkoxide-ylide \\
\texttt{OC[CH-]/[O+]=C(\textbackslash{}O)CO} & C4 ylide form \\
\texttt{O=C[C@H]([O-])CO} & D-GA C2-alkoxide \\
\texttt{[H]/[O+]=[C-]\textbackslash{}[C@H](O)CO} & D-GA carbonyl ylide \\
\texttt{O=C[C@H](O)[C@H](O)CO} & D-erythrose \\
\texttt{OC[C@@H](O)[C@H](O)C(O)O} & D-erythrose gem-diol \\
\texttt{O=C(CO)[C@H](O)CO} & D-erythrulose \\
\texttt{O=C[C@H](O)CO} & D-glyceraldehyde \\
\texttt{[H]/[O+]=[C-]/[C@H](O)CO} & D-glyceraldehyde carbonyl ylide \\
\texttt{OC[C@@H](O)C(O)O} & D-glyceraldehyde gem-diol \\
\texttt{O/C=C(/O)[C@H](O)CO} & D-tetrose E-1,2-enediol \\
\texttt{O/C=C(\textbackslash{}O)[C@H](O)CO} & D-tetrose Z-1,2-enediol \\
\texttt{O=C[C@@H](O)[C@H](O)CO} & D-threose \\
\texttt{O=C(CO)CO} & DHA \\
\texttt{OC/[C-]=[O+]/CO} & DHA C-O+ ylide (E) \\
\texttt{OC/[C-]=[O+]\textbackslash{}CO} & DHA C-O+ ylide (Z) \\
\texttt{O/C=C/O} & E-ethene-1,2-diol \\
\texttt{O/[C-]=[O+]/CCO} & E-GA-skeleton ylide \\
\texttt{[H]/[O+]=[C-]/CO} & E-GO carbonyl ylide \\
\texttt{[H]/[C-]=[O+]/CO} & E-GO carbonyl ylide (alt) \\
\texttt{O/[C-]=C/O} & E-ethene-1,2-diol carbanion \\
\texttt{[O-]/C=C/O} & E-ethene-1,2-diolate \\
\texttt{C/[O+]=[C-]/O} & E-methyl-vinyl-ether ylide \\
\texttt{[H]/[O+]=[C-]/[C@@H](O)[C@H](O)CO} & E-threose carbonyl ylide \\
\texttt{C=O} & FA \\
\texttt{[H]/[C-]=[O+]/[H]} & FA E-carbonyl-ylide \\
\texttt{[H]/[C-]=[O+]\textbackslash{}[H]} & FA Z-carbonyl-ylide \\
\texttt{OCC=C(O)O} & GA 1,2-enediol with C1-hydrate \\
\texttt{O=[C-]CO} & GA acyl carbanion \\
\texttt{O[CH-]/[O+]=C/CO} & GA ylide \\
\texttt{O=CC[O-]} & GO alkoxide ion \\
\texttt{OCC(O)O} & GO gem-diol hydrate (Kua 7) \\
\texttt{[O-][C@@H](O)CO} & GO gem-diol monoanion (R) \\
\texttt{[O-][C@H](O)CO} & GO gem-diol monoanion (S) \\
\texttt{O=CCO} & glycolaldehyde \\
\texttt{O[C@@H]1CO[C@@H](O)CO1} & Kua 1 \\
\texttt{O[C@@H]1CO[C@H](O)CO1} & Kua 1 (epimer) \\
\texttt{OC[C@@H]1OC[C@@H](O)O1} & Kua 4/5 \\
\texttt{OC[C@@H]1OC[C@H](O)O1} & Kua 4/5 \\
\texttt{OC[C@H]1OC[C@@H](O)O1} & Kua 4/5 \\
\texttt{O=C[C@@H]([O-])CO} & L-GA C2-alkoxide \\
\texttt{[H]/[O+]=[C-]/[C@@H](O)CO} & L-GA carbonyl ylide \\
\texttt{[H]/[O+]=[C-]\textbackslash{}[C@@H](O)CO} & L-GA carbonyl ylide (alt) \\
\texttt{O=C[C@@H](O)[C@@H](O)CO} & L-erythrose \\
\texttt{O=C(CO)[C@@H](O)CO} & L-erythrulose \\
\texttt{O=C[C@@H](O)CO} & L-glyceraldehyde \\
\texttt{OC[C@H](O)C(O)O} & L-glyceraldehyde gem-diol \\
\texttt{O/C=C(\textbackslash{}O)[C@@H](O)CO} & L-tetrose Z-1,2-enediol \\
\texttt{O=C[C@H](O)[C@@H](O)CO} & L-threose \\
\texttt{O=C[C@H](O)C[O-]} & LdB: D-GA C3-alkoxide \\
\texttt{O=C(C[O-])CO} & LdB: DHA C1-alkoxide \\
\texttt{O/C=C(/O)CO} & LdB: E-1,2-enediol \\
\texttt{[O-]/C(=C/O)CO} & LdB: E-prop-1-enetriol (DHA side) \\
\texttt{O=C[C@@H](O)C[O-]} & LdB: L-GA C3-alkoxide \\
\texttt{[O-]/C=C(\textbackslash{}O)CO} & LdB: Z-1,2-enediolate \\
\texttt{[O-]/C(=C\textbackslash{}O)CO} & LdB: Z-prop-1-enetriol (DHA side) \\
\texttt{[O-]/C=C(/O)CO} & LdB: E-1,2-enediolate \\
\texttt{O/C=C(\textbackslash{}O)CO} & LdB: Z-1,2-enediol \\
\texttt{OCO} & methanediol \\
\texttt{O=CCO[C@@H](O)CO} & R-GO dimer (Kua 2) \\
\texttt{O=CCO[C@H](O)CO} & S-GO dimer (Kua 2) \\
\texttt{O/C=C\textbackslash{}O} & Z-ethene-1,2-diol \\
\texttt{O/[C-]=[O+]\textbackslash{}CCO} & Z-GA-skeleton ylide \\
\texttt{[H]/[O+]=[C-]\textbackslash{}CO} & Z-GO carbonyl ylide \\
\texttt{[O-]/C=C\textbackslash{}O} & Z-ethene-1,2-diolate \\
\texttt{C/[O+]=[C-]\textbackslash{}O} & Z-methyl-vinyl-ether ylide \\
\texttt{O=CC[O-].[C-]\#[O+]} & GO alkoxide + CO \\
\texttt{[CH2-]OCOCO} & anionic methanediol trimer \\
\texttt{O/C=[O+]\textbackslash{}[C-](CO)CO} & branched apiose-ald-like ylide \\
\texttt{OC1OCOCO1} & cyclo 3-FA \\
\texttt{[CH-]=O} & formyl anion \\
\texttt{[H]/[C-]=[O+]\textbackslash{}CO} & glycolaldehyde carbonyl ylide (Z, H on C) \\
\texttt{OCC(O)(O)CO} & hydrated DHA \\
\texttt{OCOCO} & methanediol dimer (2-FA) \\
\texttt{OCOCOCO} & methanediol trimer (3-FA) \\
\texttt{[O-]CO} & methanediolate \\
\texttt{O[CH-]O} & methanediolate C-deprotonated \\
\texttt{C/[O+]=[C-]/[C@H](O)C(O)O} & methyl-O+=C- with alkylated gem-diol \\
\texttt{OC/C(O)=C(\textbackslash{}O)CO} & tetrose Z-2,3-enediol \\
\texttt{C/[O+]=[C-]\textbackslash{}C(O)O} & methyl-O+=C- with gem-diol \\
\end{longtable}
\end{footnotesize}

\subsubsection{Compressed formose cycles}
\label{app:compressed}
\begin{figure}[H]
    \centering
    \includegraphics[width=0.85\linewidth]{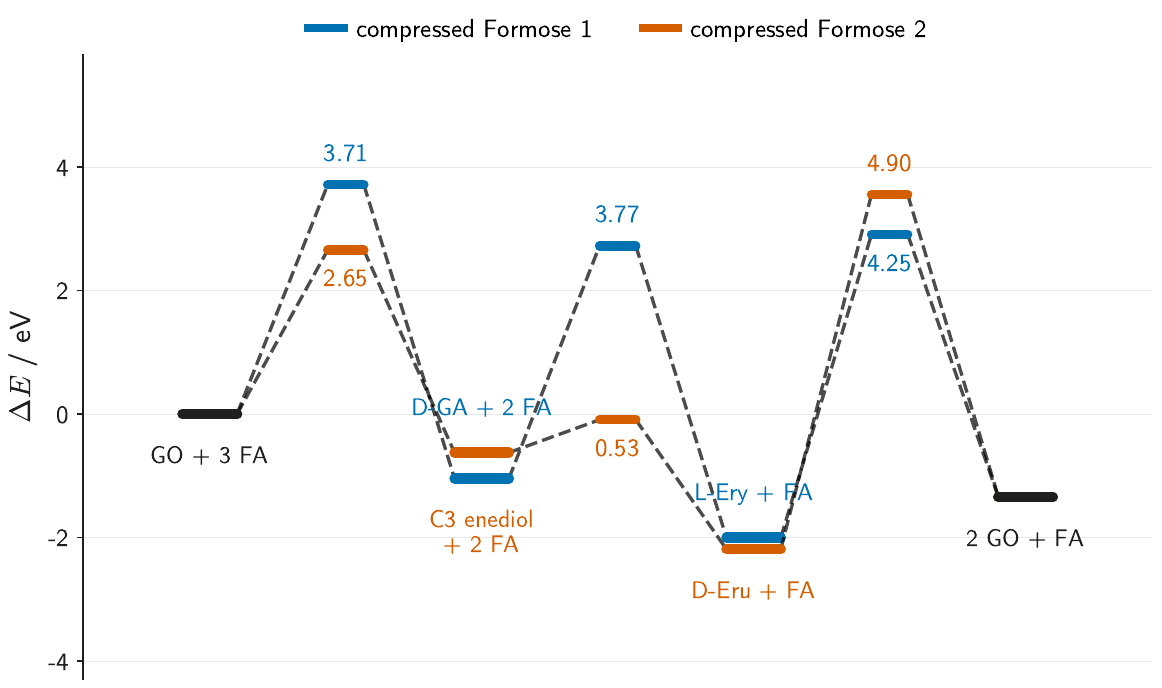}
    \caption{
    Energy profiles of the two compressed formose cycles of Figure~\ref{fig:formose_results}d, drawn between the shared GO + 3 FA and 2 GO + FA ensembles.
    Compressed Formose 1 (blue) is the compressed Breslow route through D-GA and L-erythrose; compressed Formose 2 (orange) is the C$_3$-enediol route through a C$_3$ enediol and D-erythrulose.
    Each closing step bundles a retro-aldol and a tautomerization into one TS, inflating the worst barrier to 4.25 and 4.90~eV respectively.
    PBE0/def2-TZVPP energies with DFT-separated barriers.
    }
    \label{sfig:compressed}
\end{figure}

As reported in the main text, in addition to the canonical Breslow route, two further auto-catalytic cycles between the same GO + 3 FA and 2 GO + FA ensembles are recovered, drawn as the blue and orange cycles of Figure~\ref{fig:formose_results}d.
We call them ``compressed'' because their closing step compresses a retro-aldol and a tautomerization into a single TS (tetrose $\to$ 2 GO).
This is in contrast to the fully elementary canonical Breslow route discussed in the main text.
Figure~\ref{sfig:compressed} shows their energy profiles:
the Breslow route (blue) through D-GA and L-erythrose, with a highest barrier of 4.25~eV, and the
C$_3$-enediol route (orange) through a C$_3$ enediol and D-erythrulose, with a highest barrier of 4.90~eV.
These barriers are higher than the 3.85~eV of the elementary cycle because these alternative routes combine the Lobry de Bruyn--van~Ekenstein\cite{lobrydebruyn1895} step into the final TS.

More broadly, recovering several distinct cycles for the same net transformation is an opportunity \textit{and} a desired redundancy: 
each route flows through different intermediates, handing experimentalists complementary ways to probe the same chemistry, e.g. different species to intercept, trap, or spectroscopically track, which a CRN mapped \textit{ab origine} delivers automatically.

\subsubsection{Autocatalytic cores in the explored network}
\label{si:autocat}

Blokhuis \textit{et al.}\cite{blokhuis2020} classify autocatalytic cycles by looking at stoichiometric motifs (a generalization of the basic chemical equation, e.g. A+B $\to$ 2C, etc.)
A minimal subgraph of a network (compounds and reactions), which they call an \textit{autocatalytic core}, can be used to characterize the type of cycle occurring in that network.

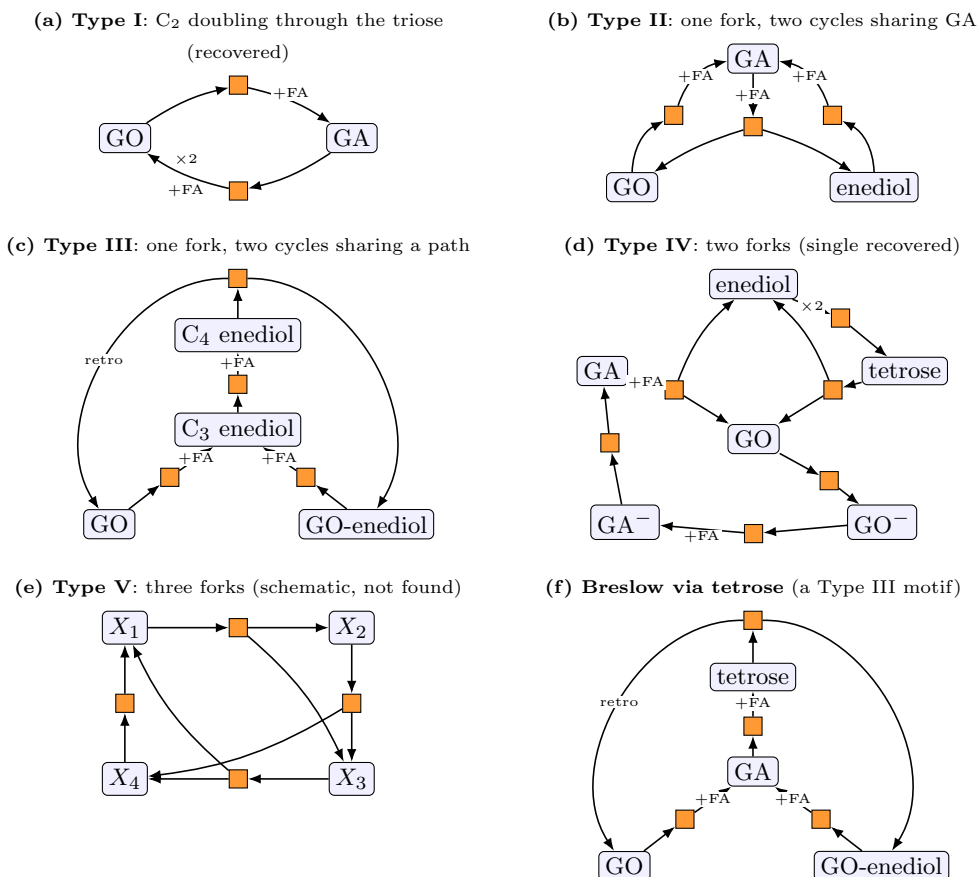
\begin{figure}[H]
\centering
\tikzset{
  cpd/.style={rounded corners=2pt, draw=black, fill=blue!6, align=center,
              font=\scriptsize, inner sep=2.4pt},
  rxn/.style={draw=black, fill=orange!80, minimum size=2.4mm, inner sep=0pt},
  ar/.style={-{Latex[length=1.7mm]}, semithick},
  lbl/.style={font=\tiny, fill=white, inner sep=0.6pt},
}
\begin{minipage}[t]{0.47\linewidth}\centering
{\footnotesize\textbf{(a) Type I}: C$_2$ doubling through the triose (recovered)}\\[4pt]
\begin{tikzpicture}[baseline]
  \node[cpd] (go) at (0,0) {GO};
  \node[cpd] (ga) at (3.0,0) {GA};
  \node[rxn] (r1) at (1.5,0.7) {};
  \node[rxn] (r2) at (1.5,-0.7) {};
  \draw[ar] (go) to[bend left=10] (r1);
  \draw[ar] (r1) to[bend left=10] node[lbl,above]{$+$FA} (ga);
  \draw[ar] (ga) to[bend left=10] (r2);
  \draw[ar] (r2) to[bend left=10] 
    node[lbl,below=0.1cm]{$+$FA} 
    node[lbl,above=0.1cm]{$\times2$} (go);
\end{tikzpicture}
\end{minipage}\hfill
\begin{minipage}[t]{0.47\linewidth}\centering
{\footnotesize\textbf{(b) Type II}: one fork, two cycles sharing GA}\\[4pt]
\begin{tikzpicture}[baseline]
  \node[cpd] (ga) at (1.6,1.7) {GA};
  \node[cpd] (go) at (0,0) {GO};
  \node[cpd] (ed) at (3.2,0) {enediol};
  \node[rxn] (rf) at (1.6,0.8) {};
  \node[rxn] (rg) at (0.55,0.95) {};
  \node[rxn] (re) at (2.65,0.95) {};
  \draw[ar] (ga) -- node[lbl]{$+$FA} (rf);
  \draw[ar] (rf) to[bend right=8] (go);
  \draw[ar] (rf) to[bend left=8] (ed);
  \draw[ar] (go) to[bend left=30] (rg);
  \draw[ar] (rg) to[bend left=30] node[lbl]{$+$FA} (ga);
  \draw[ar] (ed) to[bend right=30] (re);
  \draw[ar] (re) to[bend right=30] node[lbl]{$+$FA} (ga);
\end{tikzpicture}
\end{minipage}

\vspace{12pt}
\begin{minipage}[t]{0.47\linewidth}\centering
{\footnotesize\textbf{(c) Type III}: one fork, two cycles sharing a path}\\[4pt]
\begin{tikzpicture}[baseline]
  \node[cpd] (c4) at (1.5,2.5) {C$_4$ enediol};
  \node[cpd] (c3) at (1.5,1.25) {C$_3$ enediol};
  \node[cpd] (go) at (-0.2,0) {GO};
  \node[cpd] (ge) at (3.2,0) {GO-enediol};
  \node[rxn] (r1) at (0.6,0.62) {};
  \node[rxn] (r2) at (2.4,0.62) {};
  \node[rxn] (r3) at (1.5,1.85) {};
  \node[rxn] (rf) at (1.5,3.25) {};
  \draw[ar] (go) -- (r1); \draw[ar] (r1) -- node[lbl]{$+$FA} (c3);
  \draw[ar] (ge) -- (r2); \draw[ar] (r2) -- node[lbl]{$+$FA} (c3);
  \draw[ar] (c3) -- (r3); \draw[ar] (r3) -- node[lbl]{$+$FA} (c4);
  \draw[ar] (c4) -- (rf);
  \draw[ar] (rf) to[bend right=62] node[lbl]{retro} (go);
  \draw[ar] (rf) to[bend left=62] (ge);
\end{tikzpicture}
\end{minipage}\hfill
\begin{minipage}[t]{0.47\linewidth}\centering
{\footnotesize\textbf{(d) Type IV}: two forks (single recovered)}\\[4pt]
\begin{tikzpicture}[baseline]
  \node[cpd] (ed)  at (1.6,2.6)  {enediol};
  \node[cpd] (ga)  at (-0.4,1.45){GA};
  \node[cpd] (tet) at (3.6,1.45) {tetrose};
  \node[cpd] (go)  at (1.6,0.55) {GO};
  \node[cpd] (goa) at (3.3,-0.55){GO$^-$};
  \node[cpd] (gaa) at (-0.1,-0.55){GA$^-$};
  \node[rxn] (f1) at (0.55,1.2) {};
  \draw[ar] (ga) -- node[lbl]{$+$FA} (f1);
  \draw[ar] (f1) -- (go);
  \draw[ar] (f1) to[bend left=14] (ed);
  \node[rxn] (f2) at (2.65,1.2) {};
  \draw[ar] (tet) -- (f2);
  \draw[ar] (f2) -- (go);
  \draw[ar] (f2) to[bend right=14] (ed);
  \node[rxn] (m) at (2.75,2.15) {};
  \draw[ar] (ed) -- node[lbl]{$\times2$} (m);
  \draw[ar] (m) -- (tet);
  \node[rxn] (a1) at (2.6,0.0) {};
  \draw[ar] (go) -- (a1); \draw[ar] (a1) -- (goa);
  \node[rxn] (a2) at (1.6,-0.7) {};
  \draw[ar] (goa) -- (a2); \draw[ar] (a2) -- node[lbl,below]{$+$FA} (gaa);
  \node[rxn] (a3) at (-0.3,0.5) {};
  \draw[ar] (gaa) -- (a3); \draw[ar] (a3) -- (ga);
\end{tikzpicture}
\end{minipage}

\vspace{12pt}
\begin{minipage}[t]{0.47\linewidth}\centering
{\footnotesize\textbf{(e) Type V}: three forks (schematic, not found)}\\[4pt]
\begin{tikzpicture}[baseline]
  \node[cpd] (x1) at (0,2.0) {$X_1$};
  \node[cpd] (x2) at (3.0,2.0) {$X_2$};
  \node[cpd] (x3) at (3.0,0) {$X_3$};
  \node[cpd] (x4) at (0,0) {$X_4$};
  \node[rxn] (a) at (1.5,2.0) {};   %
  \node[rxn] (b) at (3.0,1.0) {};   %
  \node[rxn] (c) at (1.5,0) {};     %
  \node[rxn] (d) at (0,1.0) {};     %
  \draw[ar] (x1) -- (a); \draw[ar] (a) -- (x2);
  \draw[ar] (a) to[bend left=12] (x3);
  \draw[ar] (x2) -- (b); \draw[ar] (b) -- (x3);
  \draw[ar] (b) to[bend left=12] (x4);
  \draw[ar] (x3) -- (c); \draw[ar] (c) -- (x4);
  \draw[ar] (c) to[bend left=12] (x1);
  \draw[ar] (x4) -- (d); \draw[ar] (d) -- (x1);
\end{tikzpicture}
\end{minipage}\hfill
\begin{minipage}[t]{0.47\linewidth}\centering
{\footnotesize\textbf{(f) Breslow via tetrose} (a Type III motif)}\\[4pt]
\begin{tikzpicture}[baseline]
  \node[cpd] (tet) at (1.5,2.5) {tetrose};
  \node[cpd] (ga) at (1.5,1.25) {GA};
  \node[cpd] (go) at (-0.2,0) {GO};
  \node[cpd] (ge) at (3.2,0) {GO-enediol};
  \node[rxn] (r1) at (0.6,0.62) {};
  \node[rxn] (r2) at (2.4,0.62) {};
  \node[rxn] (r3) at (1.5,1.85) {};
  \node[rxn] (rf) at (1.5,3.25) {};
  \draw[ar] (go) -- (r1); \draw[ar] (r1) -- node[lbl]{$+$FA} (ga);
  \draw[ar] (ge) -- (r2); \draw[ar] (r2) -- node[lbl]{$+$FA} (ga);
  \draw[ar] (ga) -- (r3); \draw[ar] (r3) -- node[lbl]{$+$FA} (tet);
  \draw[ar] (tet) -- (rf);
  \draw[ar] (rf) to[bend right=62] node[lbl]{retro} (go);
  \draw[ar] (rf) to[bend left=62] (ge);
\end{tikzpicture}
\end{minipage}
\caption{
Example autocatalytic-core motifs in the ReactionAtlas explored formose network.
The light blue nodes are compounds, orange squares reactions (each with a TS), and a fork is a reaction with two product arrows.
Arrow labels give the feedstock (FA, formaldehyde).
For each type, in \textbf{(a)} through \textbf{(d)}, an example from our network is shown.
\textbf{(e)} Three-fork Type V which has never been found.
\textbf{(f)} As comparison, the textbook tetrose Breslow, which in the network is a Type III.}
\label{sfig:autocat_cores}
\end{figure}

All autocatalytic cycles have cores which fall into five motif types (see Figure~\ref{sfig:autocat_cores}, a--e) distinguished by the number of forks (one reactant with two product reactions):
Type I has one doubling fork (a species produced in two copies) in a single graph cycle,
Types II and III have one fork with two distinct products embedded in two overlapping cycles, Type IV has two forks, and Type V three.

We applied this taxonomy to the CRN explored by ReactionAtlas treating stereoisomers, protonation, and $E$/$Z$ geometric isomers as the same compound, since cores are defined over species rather than conformers.
Restricted to carbohydrate chemistry the network contains about 90 distinct minimal autocatalytic cores:
one Type~I, 90 of Types~II and III, and one Type~IV.
Across all these cores, glycolaldehyde (GO) is the autocatalyst
with glyceraldehyde (GA) and the C$_2$ compound enediol
coming in on second and third rank.
ReactionAtlas thus predicts that GO is the privileged seed of the formose reaction\cite{socha1981,kieboom1984} from its topology alone.

However, autocatalysis in ReactionAtlas is not restricted to sugars.
We find more autocatalytic cores when including single-carbon compounds,
including 25 additional Type~IV motifs.
ReactionAtlas's ability to find Type~IV autocatalytic motifs is noteworthy:
Blokhuis \textit{et al.}\cite{blokhuis2020} only find instances of Types~I--III in every previously described chemical or biological autocatalytic system, stating explicitly that they ``have not yet found examples of types IV and V''. To our knowledge the cores contained in our CRN dataset are the first described instances of Type~IV autocatalytic cores.
For further directions, we refer to \cite{golnik2026}.

\subsubsection{Wider technical comparison with the formose literature}
\label{si:wider_literature}

The classical formose cycle starts with the dimerization of two formaldehyde (FA) to a glycolaldehyde (GO),
  which is extremely slow as concerted dimerization ($\sim$75~kcal/mol) and the rate limiting step.
  How this actually proceeds has long been debated
  \cite{weiss1970,kim2011,bissette2013,schwartz1993}.
  
  Kan\cite{kan2026} propose a two-step \emph{anion-mediated} pathway: 
  a FA is first converted to the formyl anion HCO$^-$\cite{luisi} (FA$^-$), which then adds to a second FA to give the GO-alkoxide ${}^-$OCH$_2$CHO.
  This is a hypothesis to avoid the difficult dimerization step and lowers the barrier to $\sim26.9$ kcal/mol.
  This mechanism is \emph{not} found by ReactionAtlas as an elementary step, but not because ReactionAtlas is blind to this type of reaction:
  The formyl anion HCO$^-$ is present and occurs in 13 elementary reactions and the reaction family 
  (nucleophilic addition of a carbon-centred anion to a FA carbonyl, producing a new C-C bond and an alkoxide on the former FA oxygen),
  is also well represented in our CRN.
  
  ReactionAtlas proposes a dimerization via 
  a 4-center concerted $[2\sigma{+}2\sigma]$ ring-coupling with a simultaneous 1,4-hydrogen shift from one of the FA carbons to the other FA oxygen.  
  The PBE0/def2-TZVPP barrier (relative to two isolated PBE0-optimised FA monomers) is $\sim33.4$ kcal/mol.
  The magnitudes are comparable to Kan \emph{et al.}'s barrier, although the two refer to different mechanisms.
              
  Two further results from section ``Beyond the Basic Network'' of Kan et al.~\cite{kan2026} are confirmed:
  First, the \emph{branched tetrose}                            
  $\mathrm{HOCH_2{-}C(OH)(CHO){-}CH_2OH}$ (Eq. \textbf{30} in Kan)
  is found by ReactionAtlas in 66 reactions and as reported by Kan et al.:
  none of them undergoes an aldol forward reaction and is a dead-end.
  This is consistent with Benner et al.\cite{kim2011} and Kua and Tripoli\cite{kua2024}.
  Second, the \emph{furanose tetroses} occur as $33$ ring-closed isomers, with cyclic $\leftrightarrow$ open barriers of 24--50 kcal/mol consistent with Kan.
  These are a non-trivial finding of ReactionAtlas, as the generative model has no pre-determined rule that would prohibit such steps, yet it recovers the correct chemistry.

We compare our ReactionAtlas against Kua \textit{et al.}~\cite{kua2013,kua2013b}, which used B3LYP/6-311G** with solvent and entropic corrections to map the small CRNs around FA and GO manually.
From the GO study~\cite{kua2013b} we recover \textit{all} five unique dimer minima.
Our PBE0 barrier for the reaction of GO to gem-diol (42.5 kcal/mol) agrees with Kua's own \textit{in vacuo} 4-center variant (40.8 kcal/mol).
We also recover Kua's retro-aldol $C_4\to 2\mathrm{GO}$ pathway at 9.0 vs. their 14.3 kcal/mol.
From the FA study~\cite{kua2013} we recover 14 of the 19 manually-mapped minima within our imposed tetrose limit.

The studies of Yi \textit{et al.}\cite{yi2022} and Sutton \textit{et al.}\cite{sutton2025} investigated formose reactions via NMR and \textsuperscript{13}C labeling at pH 8.5, room temperature and no calcium.
Yi \textit{et al.} investigated the carbonyl migration in tetroses while Sutton \textit{et al.} extended the analysis to the aldol reactions of FA with GO, DHA, erythrulose, and erythrose.
We examine two of their observations:

i) Yi and Sutton both conclude that carbonyl migration of tetroses (and GA to DHA) proceeds via enediol(ate) intermediates (LdB-AvE), not the intramolecular 1,2-H shift proposed by Breslow under pH 12 and Ca\textsuperscript{2+} conditions~\cite{breslow1959,appayee2014,bris2024}. 
The main text already establishes that in our network this interconversion runs entirely through the enediol, with the direct neutral H-shift absent (Section~\ref{sec:formose_cycle}).

We find every enediol(ate) intermediate this route requires: four stereoisomers of the C$_4$ 1,2-enediol, one C$_4$ 2,3-enediol, and ca.\ 60 zwitterionic C$_4$ enediolates.

\begin{figure}[h]
\centering
\resizebox{\textwidth}{!}{%
\begin{tikzpicture}[
  >=Stealth, thick,
  neutral/.style={draw, rounded corners=2pt, fill=blue!8, inner sep=3pt, minimum width=18mm, align=center, font=\small},
  anion/.style={draw, rounded corners=2pt, fill=red!8, inner sep=3pt, minimum width=24mm, align=center, font=\small},
  ed/.style={draw, rounded corners=2pt, fill=orange!18, inner sep=3pt, minimum width=24mm, align=center, font=\small},
  link/.style={<->, line width=0.7pt},
  el/.style={font=\scriptsize, inner sep=1pt}
]
\node[neutral] (DGA) at (0, 1.3) {D-GA};
\node[neutral] (LGA) at (0,-1.3) {L-GA};
\node[anion]   (DCA) at (3.2, 1.3) {D-GA-C3-alkoxide};
\node[anion]   (LCA) at (3.2,-1.3) {L-GA-C3-alkoxide};
\node[ed]      (Eed) at (6.6, 0.55) {E-1,2-enediolate};
\node[ed]      (Zed) at (6.6,-0.55) {Z-1,2-enediolate};
\node[anion]   (DHAa) at (10.0, 0) {DHA-C1-alkoxide};
\node[neutral] (DHA)  at (13.0, 0) {DHA};
\draw[link] (DGA) -- (DCA);
\draw[link] (LGA) -- (LCA);
\draw[link] (DHA) -- (DHAa);
\draw[link] (DCA) -- (Eed);
\draw[link] (LCA) -- node[el, above, left=0.2cm] {39.1} (Eed);
\draw[link] (Eed) -- node[el, right=0.1cm] {43.2} (Zed);
\draw[link] (Zed) -- (DHAa);
\draw[link] (Eed) -- node[el, above=0.1cm] {81.5} (DHAa);
\draw[link] (DCA) -- (Zed);
\draw[link] (LCA) -- (Zed);
\end{tikzpicture}%
}
\caption{The sugars (blue; D-GA, L-GA, DHA) connect to the 1,2-enediolates (orange) through ionic intermediates (red).
Each labeled arrow is a single elementary PBE0 step with the forward barrier (kcal/mol).}
\label{fig:ldb-ave-triose}
\end{figure}
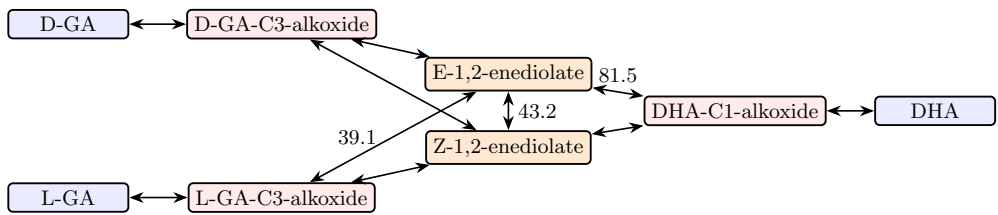

Resolving stereochemistry, ReactionAtlas moreover reveals that GA racemisation and the GA/DHA isomerisation are not two independent reactions but a single coupled cycle: \textsc{d}-GA, \textsc{l}-GA and DHA all connect to the same pair of 1,2-enediolates, so the \textsc{d}-GA $\leftrightarrow$ DHA $\leftrightarrow$ \textsc{l}-GA loop closes through one shared enediolate hub (Figure~\ref{fig:ldb-ave-triose}). This is precisely the LdB--AvE picture Yi and Sutton infer, here recovered \textit{ab origine} and with the explicit \textsc{d}/\textsc{l} stereochemistry that previous formose networks do not resolve.

ii) We test whether our network reproduces Sutton's observation that the formose reaction cannot selectively make \textit{linear} aldoses.
Sutton \textit{et al.}\cite{sutton2025} find that the FA aldol reaction with any sugar (GO, DHA, erythrulose, erythrose) proceeds to ketoses.
  Our network reproduces this: 
  the neutral FA aldol to a \textit{linear} aldotetrose is absent (chain growth runs through the enediolate, claim~i), and the branched tetrose is a retro-only dead-end (see above with Kan et al.\cite{kan2026}). 
  The one different channel that does build a linear aldotetrose is explained below.

Our network provides a computational rationale for the regiochemical preferences observed experimentally by Bri\v{s} et al.~\cite{bris2024} during the formose reaction.
Bri\v{s} et al. show that calcium-coordinated ketoses undergo fast retro-aldol cleavage by splitting off a FA.
For a canonical ketotetrose, our ReactionAtlas recovers this preference:
the lowest 4 $\to$ 3+1 barrier is 42~kcal/mol, compared to 75~kcal/mol for the 4 $\to$ 2+2 pathway.

Applying the same idea to the aldotetrose predicts a complementary 4 $\to$ 2+2 preference (so opposite).
Our network confirms this quantitatively, with a lowest 2+2 barrier of 40~kcal/mol versus 68~kcal/mol for the 3+1 path.
Because Bri\v{s}'s data are silent on this point (aldose retro-aldol may be too slow to characterize), this calculation fills a mechanistic gap.

This aldotetrose 2+2 split is the productive doubling step of the cycle, and ReactionAtlas obtains the needed linear aldotetrose via GO dimerization.
This is the one point where we diverge from Sutton \textit{et al.}\cite{sutton2025} (claim~ii), who do not observe linear aldotetroses in solution:
the discrepancy is one of \emph{gas phase versus solution regiochemistry} (not calcium, since both are calcium-free).

\section{Implementation Details}
Below the ReactionAtlas algorithm is documented in detail, including all hyperparameters used in the experiments. For further detail see the code at \hyperlink{github.com/mx-e/reaction-atlas}{github.com/mx-e/reaction-atlas}.

\subsection{Generative-Loop Reaction Discovery}
\label{si:generative}

At a high level, the generative loop repeatedly asks one question: given a single molecule, or a colliding pair of molecules, what reaction could it undergo and where is the transition state? A diffusion model trained on known transition states answers by proposing a candidate geometry directly from the reactant, and a number of physically motivated filters then retains only those candidates that correspond to a valid reaction. 

Each iteration processes a batch of $32$ \textbf{exploration contexts}. When an exploration context is created it contains the reactant geometry, either a single molecule or a merged pair. During the processing additional artifacts are added to the context such as: TS conformer, product conformer, IRC trajectories, fragment list, barriers. The exploration contexts are processed in eight stages:
\begin{enumerate}
    \item \textbf{sampling}; choosing which candidate reactant geometries to explore.
    \item \textbf{merge enhancement}; for merged geometries, sampling orientations in which these geometries are merged.
    \item \textbf{MoreRed diffusion}; noise is added and the MoreRed diffusion model attempts to denoise the sampled geometries to a nearby TS.
    \item \textbf{forward barrier gate}; TS candidates with a forward energy barrier that is too high are discarded
    \item \textbf{Hessian + IRC validation}; TS Hessians and connected minima are evaluated
    \item \textbf{endpoint identity}; connected minima are checked, they should be distinct and one should be the reactant geometry.
    \item \textbf{fragmentation detection}; if a connected minima has disconnected fragements they are separated. 
    \item \textbf{backward barrier gate}; TS candidates with a backward energy barrier that is too high are discarded.
\end{enumerate}
Critic stages (4--8) reuse the shared machinery in SI-\ref{si:shared} (ML force field, geometry primitives, optimizers), this section documents what is specific to the generative path.

\subsubsection{Stage 1: Concentration-weighted context sampling}
\label{sec:sampling}

To find diverse reaction types (we consider $1\rightarrow 1, 1\rightarrow2, 2\rightarrow1 \text{, and } 2\rightarrow2$-reactions where $n\rightarrow m$ counts the number of reactants and products), half of  exploration contexts start with a single-molecule geometry, while the other half contain bimolecular merge pairs.

For sampling single molecules, a species is drawn from the kinetic snapshot's steady-state distribution (SI-\ref{si:kinetics}); compounds with $n_\text{atoms} < 3$ are excluded. Within the chosen compound, the conformer is drawn from the room-temperature Boltzmann distribution,
\begin{equation}
    P(m) = \frac{\exp(-(E_m - E_\text{min}) / k_B T)}{\sum_{m'} \exp(-(E_{m'} - E_\text{min}) / k_B T)}, \qquad k_B T = \SI{0.0257}{eV}.
\end{equation}
For bimolecular pair sampling $50 \times n_\text{needed}$ candidate pairs are drawn from the steady-state distribution, then filtered against (i)~a composition cap of $\{C{:}5,\ H{:}10,\ O{:}5\}$ and (ii)~a per-pair coverage cap of $4$ reactions in the current CRN (the per-pair coverage cap is removed for the formose drilldown experiment). These caps serve to constrain the exploration to chemistry we care about for this experiment and avoid unneeded computation. For each pair matching both criteria, one minimum is drawn from each compound at random. If more than $n_\text{needed}$ pairs survive filtering only $n_\text{needed}$ are kept. Each component receives an independent random $SO(3)$ rotation; a random unit vector defines the separation direction; the smallest displacement that yields a non-negative minimum van der Waals gap $\min_{ij}\bigl(\lVert r_i - r_j \rVert - r^{\text{vdW}}_i - r^{\text{vdW}}_j\bigr)$ is found by binary search, then $\SI{0.2}{\angstrom}$ of repulsive buffer is added. The merged structure is COM-centred.

\noindent\begin{tabular}{lr}
\toprule
Parameter & Value \\
\midrule
Batch size & 32 contexts \\
Single/merge split & 50/50 \\
Oversample factor (merges) & 50 \\
Min compound size & 3 atoms \\
Composition cap (merge) & $\{C{:}5,H{:}10,O{:}5\}$ \\
Per-pair reaction cap & 4 (main experiment only) \\
$k_B T$ (conformer pick) & \SI{0.0257}{eV} \\
vdW separation buffer & \SI{0.2}{\angstrom} \\
\bottomrule
\end{tabular}

\subsubsection{Stage 2: Merge encounter enhancement}
\label{sec:merge-prep}

The random encounter geometry from Stage 1 is geometrically valid but typically far from any saddle point. Pre-conditioning reduces the burden on the diffusion stage. For each merge context we: (i)~generate $5$ additional random merge geometries and pick the lowest-ML-energy candidate; (ii)~run $5$ FIRE micro-relaxation steps with $f_\text{max} = \SI{0.05}{eV/\angstrom}$ to settle the fragments into a reasonable encounter geometry without converging to the encounter-complex minimum; (iii)~reject the context if the binding energy $E_\text{bind} = E(A) + E(B) - E(\text{complex}) > \SI{0.3}{eV}$; (iv)~apply per-atom Gaussian noise of $\sigma = \SI{0.15}{\angstrom}$ to break symmetry, since MoreRed is equivariant and produces no movement on a symmetric stationary input; (v)~re-centre for the translation-invariant diffusion.

\noindent\begin{tabular}{lr}
\toprule
Parameter & Value \\
\midrule
Extra orientations ($K$) & 5 \\
FIRE micro-relax steps & 5 \\
FIRE $f_\text{max}$ & \SI{0.05}{eV/\angstrom} \\
Symmetry-break $\sigma$ & \SI{0.15}{\angstrom} \\
Encounter-complex cap & \SI{0.3}{eV} \\
Min pairwise distance & \SI{0.8}{\angstrom} \\
\bottomrule
\end{tabular}
\subsubsection{Stage 3: Diffusion-based TS proposer}
\label{sec:diffusion}

We use a SchNetPack equivariant graph network (Sch{\"u}tt et al.\ 2023), trained as a MoreRed-JT joint-time denoiser (Kahouli et al.\ 2024) on relaxed transition states. Per atom it predicts a noise vector $\boldsymbol{\epsilon}_\theta(\mathbf{x}_t, t) \in \mathbb{R}^3$ and a time-step $\hat{t}_\theta(\mathbf{x}_t)$ used for adaptive convergence. The forward process is a variance-preserving Gaussian DDPM (Ho et al.\ 2020) of length $T = 1000$ with a polynomial noise schedule (Hoogeboom et al.\ 2022),
\begin{equation}
    \bar{\alpha}_t = (1 - 2s)\bigl(1 - (t/T)^p\bigr)^2 + s, \qquad p = 2,\ s = 10^{-5},
\end{equation}
clipped so $\alpha_t / \alpha_{t-1} \geq 10^{-3}$, with forward conditional $q(\mathbf{x}_t \mid \mathbf{x}_0) = \mathcal{N}\bigl(\sqrt{\bar{\alpha}_t}\,\mathbf{x}_0,\,(1 - \bar{\alpha}_t)\,\mathbf{I}\bigr)$ and the standard DDPM lower-bound posterior variance for the reverse step. Translation invariance is enforced by re-centring at every step. Unlike text-to-image diffusion, the TS proposer does not sample from pure Gaussian noise: it partially noises a chemically meaningful input and then denoises. Each batch samples a single noise level $t \sim \mathcal{U}[100, 650]$, forward-diffuses each starting geometry as $\mathbf{x}_t = \sqrt{\bar{\alpha}_t}\,\mathbf{x}_0 + \sqrt{1 - \bar{\alpha}_t}\,\boldsymbol{\epsilon}$ with $\boldsymbol{\epsilon} \sim \mathcal{N}(0, I)$, and reverse-diffuses for $1000$ steps:
\begin{equation}
    \mathbf{x}_{t-1} = \tfrac{1}{\sqrt{\alpha_t}}\bigl(\mathbf{x}_t - \tfrac{1-\alpha_t}{\sqrt{1 - \bar{\alpha}_t}}\,\boldsymbol{\epsilon}_\theta\bigr) + \sigma_t\,\mathbf{z}, \quad \mathbf{z} \sim \mathcal{N}(0,I).
\end{equation}
Each context receives a denoised \texttt{ts\_conformer} (COM-centred), an ML \texttt{ts\_energy}, and discovery metadata (\texttt{discovery\_method = "generative"}, the injected noise level $t$, timestamp).

\noindent\begin{tabular}{lr}
\toprule
Parameter & Value \\
\midrule
$T$ (schedule length) & 1000 \\
Schedule shape & polynomial, $p=2$, $s = 10^{-5}$ \\
Clip floor & $10^{-3}$ \\
Partial-noise range & $[100, 650]$ \\
Reverse steps & 1000 \\
Cutoff & \SI{5.0}{\angstrom} \\
\bottomrule
\end{tabular}

\subsubsection{Stage 4: Forward barrier gate}
\label{sec:fwd-barrier}

Contexts are first filtered by the forward energy barrier between the reactants and the TS, which is estimated by the MD-ET model. The forward barrier:
\begin{equation}
    \Delta E_\text{fwd} =
    \begin{cases}
        E_\text{TS} - (E_A + E_B) & \text{merged, components } A,B,\\
        E_\text{TS} - E_\text{start}  & \text{single-molecule,}
    \end{cases}
\end{equation}
is accepted if $\Delta E_\text{fwd} \in [\SI{0}{eV}, \SI{2.7211}{eV}]$.

\subsubsection{Stage 5: Hessian and IRC validation}
\label{sec:hessian-irc}

The bidirectional IRC algorithm (Hessian projection, adaptive displacement, bidirectional Newton relax, energy-based direction assignment) is shared with the PES loop and described in \ref{sec:shared-irc}. There are, however, several generative model specific choices:
\begin{enumerate}
    \item The candidate is accepted as a TS iff at least one significant eigenvalue ($|\lambda_i| > 10^{-4}\,\text{eV/\AA}^2$; \ref{sec:shared-hessian}) satisfies $\lambda_i < \SI{-0.01}{eV/\angstrom^2}$. This reduces validity in favor of candidate count. We chose to allow this since the exploration speedup is significant and the refinement evaluations show that a true saddle point is often close.
    \item The clip range is wider than the PES loop's because diffusion-proposed saddle points span a broader curvature distribution:
\begin{equation}
    \Delta x = \mathrm{clip}\bigl(\tfrac{0.2}{\sqrt{|\lambda_\text{TS}|}},\ \SI{0.05}{\angstrom},\ \SI{0.5}{\angstrom}\bigr).
\end{equation}
    \item Trust-region Newton (\ref{sec:shared-newton}) with max $350$ steps (longer than the PES loop's $120$, again reflecting the broader candidate distribution), $f_\text{max} = \SI{0.005}{eV/\angstrom}$. 
\end{enumerate}
\noindent\begin{tabular}{lr}
\toprule
Parameter & Value \\
\midrule
Eigenvalue threshold (TS) & \SI{-0.01}{eV/\angstrom^2} \\
Significance mask & $10^{-4}\,\text{eV/\AA}^2$ \\
Adaptive step formula & $0.2 \cdot |\lambda|^{-1/2}$~\AA, clamped $[0.05, 0.5]$ \\
IRC max steps / direction & 350 \\
IRC $f_\text{max}$ & \SI{0.005}{eV/\angstrom} \\
\bottomrule
\end{tabular}

\subsubsection{Stage 6: Endpoint identity}
\label{sec:endpoints}

We implement two modes for checking endpoint validity during generative search.
In \textbf{strict mode} exactly one endpoint must match the initially sampled reactant geometry, with the match defined by permutation-invariant RMSD (\ref{sec:shared-rmsd}) below $\SI{1.0}{\angstrom}$. For single-molecule contexts the geometry is compared directly while for merge contexts the endpoint geometry is fragmented (\ref{sec:shared-fragments}) and the validator tries to assign both fragments to the originally sampled pair of compounds. 
In \textbf{greedy} mode (enabled in the main experiment)  the validator drops the source-compound requirement and accepts the reaction if the two endpoints describe chemically different states, i.e., each endpoint geometry contains $\leq 2$ components with canonically distinct SMILES multisets, and no \emph{spectator species} (defined as a fragment that appears as both reactant and product but does not partake in the reaction). Without greedy mode, many valid TS proposals would be rejected because the IRC found a reaction whose reactant differs from the sampler's pick (different conformer, tautomer, or related-but-distinct species). A consequence of this choice is that the CRN does not necessarily grow as one connected graph. This is a deliberate trade-off rather than a defect: the objective is broad and valid reaction coverage, not a single spanning component, and separate components are bridged automatically whenever a later reaction is discovered that connects them.

\noindent\begin{tabular}{lr}
\toprule
Parameter & Value \\
\midrule
RMSD match threshold (strict) & \SI{1.0}{\angstrom} \\
Max fragments / side & 2 \\
\bottomrule
\end{tabular}

\subsubsection{Stage 7: Fragmentation detection}
\label{sec:fragmentation}

The product side is decomposed into connected components by the covalent-radius-scaled distance criterion (\ref{sec:shared-fragments}). A fragment is considered valid if it is either a hydrogen atom or consists of more than one atom. If any fragment is invalid the context is dropped entirely. For each valid fragment, the canonical SMILES is computed and an integer charge is inferred from the perceived bond pattern (\ref{sec:shared-smiles}) - the DFT refinement (\ref{sec:kinetics-dft}) later re-derives charges from a wavefunction calculation. When a product fragments into multiple components, each fragment is independently re-relaxed on its own PES (FIRE, \ref{sec:shared-fire}; $f_\text{max} = \SI{0.005}{eV/\angstrom}$, max $200$ steps) to remove the residual distortion left by the combined-system IRC.

\noindent\begin{tabular}{lr}
\toprule
Parameter & Value \\
\midrule
Covalent-radius scale & 1.3 \\
Per-fragment $f_\text{max}$ & \SI{0.005}{eV/\angstrom} \\
Per-fragment max steps & 200 \\
\bottomrule
\end{tabular}

\subsubsection{Stage 8: Backward barrier gate}
\label{sec:bwd-barrier}

Symmetric to Stage 4. The backward barrier is $\Delta E_\text{bwd} = E_\text{TS} - E_\text{product}$ (summed over fragments if applicable), accepted iff $\Delta E_\text{bwd} \in (\SI{0}{eV}, \SI{2.7211}{eV}]$. The barrier is computed by trajectory force integration (\ref{sec:shared-traj}). Surviving contexts pass to the graph-addition step, which registers reactant/product compounds, deduplicates minima against the per-compound PES graphs (\ref{sec:shared-rmsd}, \ref{sec:shared-neb}).

\subsubsection{Filtering Statistics}
\label{sec:funnel}

\begin{table}
\centering
\begin{tabular}{lrr}
\toprule
Stage & Total & Share of total contexts \\
\midrule
Submitted (Stage 1)                                  & $3\,814\,199$         & $100.00$\% \\
After Stages 2--3 (merge prep + diffusion + dedup)   & $3\,814\,199$         & $100.00$\% \\
After Stage 4 (fwd barrier)                          & $1\,798\,768$         & $47.16$\%  \\
After Stages 5--6 (Hessian + IRC + endpoint id)      & $172\,267$            & $4.52$\%   \\
After Stage 7 (fragmentation)                        & $172\,267$            & $4.52$\%   \\
After Stage 8 (bwd barrier)                          & $127\,670$            & $3.35$\%   \\
\midrule
Added to graph (post-dedup)                          & $30\,493$             & $0.80$\%   \\
\bottomrule
\end{tabular}
\caption{Generative-loop funnel for the main run, aggregated over all batches. For comparison, the PES loop (SI-\ref{si:pes}) added $16\,879$ of $69\,450$ escaped reactions, so the generative loop contributed $\sim 64\%$ of the reactions in the graph.}
\label{tab:funnel}
\end{table}
See table above. \textbf{Stage 4} filters $\sim 53\%$ of denoised candidates. Almost all rejections are above the upper bound ($> \SI{2}{eV}$ overflow: $1{,}093{,}685$ merge + $1{,}050{,}310$ single), corresponding to strained or partially dissociated denoiser outputs. Merged complexes are far more often rejected at the cheap energy gate because the fragment-sum baseline is much further from the TS. \textbf{Stages 5--6} remove $> 90\%$ of the remainder: of $1{,}798{,}768$ entering IRC, $418{,}475$ fail the Hessian check; $125{,}090$ and $89{,}898$ fail forward/backward relaxation; $706{,}221$ have both endpoints relax back to the source (\textsc{both\_match}); $412{,}454$ have neither match; only $46{,}630$ pass strict matching. Greedy fallback engages on the $1{,}118{,}675$ strict-mismatched cases and accepts $125{,}637$. \textbf{Stage 7} rejected zero contexts in the main run, as every IRC-passing geometry fragmented cleanly. \textbf{Stage 8} filtered a further $44{,}597$ ($\sim 26\%$) on near-zero backward barriers. \textbf{Graph deduplication} removes $97{,}177$ ($\sim 76\%$ of post-Stage-8 compounds).

\subsection{PES-Loop Reaction Discovery}
\label{si:pes}

Where the generative loop reaches outward to new intermolecular reactions, the PES loop looks inward: it takes a molecule we already know and asks what other shapes it can adopt, and what reactions it can undergo on its own, by physically shaking it with a short molecular-dynamics run and then climbing from the sampled geometries to the nearby transition states. The PES loop performs this local exploration around a known minimum. It works one compound at a time, MD-samples the configuration space near a chosen minimum, then optimises trajectory frames to first-order saddle points by Partitioned Rational Function Optimisation (P-RFO), and validates each saddle point by IRC-like descent. The endpoints of that descent fall into two cases: \emph{intramolecular} (both endpoints belong to the same compound's PES graph, a conformer-to-conformer reaction) or \emph{escape} (one endpoint is a structurally different compound, possibly fragmented into multiple species, i.e., a true chemical reaction). Intramolecular saddle points are stored as edges in the per-compound PES graph; escapes are passed to the same multi-stage critic used by the generative loop (SI-\ref{si:generative}, Stages 4--8) and admitted to the reaction graph if they pass all filters.

The pipeline has four stages: (1)~molecular dynamics, (2)~batched P-RFO and Hessian validation, (3)~bidirectional IRC, (4)~endpoint classification and commitment. Energies, forces, and Hessians come from the ML force field. The functionality shared with the generative path (Hessian projection, FIRE, trust-region Newton, IRC, NEB, RMSD, fragmentation, SMILES) are documented in SI-\ref{si:shared}. PES exploration is queued at the granularity of \emph{(compound, minimum)} pairs. Each newly discovered minimum of a compound with more than two atoms is enqueued as a separate work item. In the main experiment only compounds with kinetic concentration above \SI{e-9}{} were explored to focus resources on relevant chemistry.
\subsubsection{Stage 1: Molecular dynamics}
\label{sec:pes-md}
An MD trajectory using a Langevin thermostat is launched from the selected minimum to sample configuration space around it. The goal is to visit a diverse set of geometries that lie on or near reaction pathways departing from the minimum. The integrator follows
\begin{equation}
    m_i \ddot{\mathbf{r}}_i = \mathbf{F}_i - \gamma m_i \dot{\mathbf{r}}_i + \sqrt{2 \gamma m_i k_B T}\, \boldsymbol{\xi}(t),
\end{equation}
with friction coefficient $\gamma = 1/\tau$, coupling time $\tau = \SI{100}{fs}$, and Gaussian white noise $\boldsymbol{\xi}$. Initial velocities are drawn from the Maxwell--Boltzmann distribution at the target temperature. $2500$ steps $\times \SI{0.5}{fs} = \SI{1.25}{ps}$ at \SI{300}{K}, with a snapshot saved every $10$ steps ($250$ frames). From the 250 MD frames, $32$ candidates are selected by uniform random sampling. 

\noindent\begin{tabular}{lr}
\toprule
Parameter & Value \\
\midrule
MD steps & 2500 \\
Integration step & \SI{0.5}{fs} \\
Total simulation time & \SI{1.25}{ps} \\
Temperature & \SI{300}{K} \\
Thermostat & Langevin \\
Coupling $\tau$ & \SI{100}{fs} \\
\bottomrule
\end{tabular}

\subsubsection{Stage 2: Partitioned Rational Function Optimisation (P-RFO)}
\label{sec:pes-prfo}

P-RFO (Banerjee et al.\ 1985; Baker 1986) maximizes each candidate geometry along one mode and minimizes in the orthogonal subspace, converging to a first-order saddle point. The pipeline runs all 32 candidates in lockstep (one batched GPU Hessian evaluation + one batched energy evaluation per step via \ref{sec:shared-mdet}). At each step:

\begin{enumerate}
    \item Compute $E$, $\mathbf{f} = -\nabla E$, and the projected Hessian $\mathbf{H}_\text{proj}$ (eigenpairs $\{\lambda_i, \mathbf{u}_i\}$; \ref{sec:shared-hessian}). Project the gradient: $g_i = \mathbf{u}_i \cdot \mathbf{g}$.
    \item Compute the RFO step in each mode,
    \begin{equation}
        p_i = \frac{\mu_i}{g_i}, \qquad \mu_i = \frac{\lambda_i \pm \sqrt{\lambda_i^2 + 4 g_i^2}}{2},
    \end{equation}
    with $+$ on the TS mode (walk uphill) and $-$ on the others (walk downhill).
\end{enumerate}
At step $0$ the TS mode is the lowest-eigenvalue mode. At subsequent steps it is the eigenvector with the largest overlap to the previous TS eigenvector,
\begin{equation}
    i_\text{TS} = \arg\max_i \left| \mathbf{u}_i \cdot \mathbf{u}_\text{TS}^{(\text{prev})} \right|.
\end{equation}
This prevents the optimizer from jumping between modes when eigenvalues cross. Two adaptive trust radii are maintained: a TS-mode radius (initially $1.5\times$ the minimum-mode radius) and a minimisation-subspace radius. After each trial step, the trust-region ratio
\begin{equation}
    \rho = \frac{\Delta E_\text{actual}}{\Delta E_\text{predicted}}, \qquad \Delta E_\text{pred} = -\mathbf{f} \cdot \mathbf{p} + \tfrac{1}{2} \mathbf{p}^\top \mathbf{H} \mathbf{p}
\end{equation}
is computed. The step is accepted if $\rho > 0.1$ (otherwise the trust radius is halved and the step retried, up to $5$ times). After acceptance, the radius shrinks by $0.5$ if $\rho < 0.25$ or $\rho > 2$, and expands by $1.5$ if $0.5 < \rho < 1.5$ and the step hit the trust boundary. A candidate is declared converged when $\max_i |f_i| < \SI{0.01}{eV/\angstrom}$ \emph{and} $f_\text{RMS} < \SI{0.01}{eV/\angstrom}$ \emph{and} the projected Hessian has exactly one significant negative eigenvalue. After convergence the Hessian is re-examined. A single curvature threshold $\lambda_\text{zero} = \SI{0.0062}{eV/\angstrom^2}$ is used to mask numerically insignificant eigenvalues, and the candidate is admitted only if it retains its negative eigenvalue.

\noindent\begin{tabular}{lr}
\toprule
Parameter & Value \\
\midrule
Candidates per iteration & 32 \\
Max steps & 120 \\
$f_\text{max}$ tolerance & \SI{0.01}{eV/\angstrom} \\
$f_\text{RMS}$ tolerance & \SI{0.01}{eV/\angstrom} \\
Initial trust radius & \SI{0.5}{\angstrom} \\
Trust radius bounds & $[0.01, 0.3]$~\AA \\
TS-mode trust scale & 1.5 \\
Accept threshold ($\rho$) & 0.1 \\
Max retries / step & 5 \\
Minimum significant curvature  &$\SI{0.0062}{eV/\angstrom^2}$\\
\bottomrule
\end{tabular}

\subsubsection{Stage 3: Bidirectional IRC}
\label{sec:pes-irc}

The imaginary mode is followed in both directions and relaxed to a minimum on each side; the bidirectional IRC algorithm is shared with the generative loop and described in \ref{sec:shared-irc}. PES-loop-specific choices: 

\begin{enumerate}
    \item Both endpoints must lie energetically below the TS by at least $\SI{0.043}{eV}$:
\begin{equation}
    E_\text{left} - E_\text{TS} < \SI{-0.043}{eV} \quad \text{and} \quad E_\text{right} - E_\text{TS} < \SI{-0.043}{eV},
\end{equation}
\item Both must be structurally distinct, with permutation-invariant RMSD $> \SI{0.25}{\angstrom}$. The higher-energy endpoint is labelled \emph{forward} (reactant), the lower-energy \emph{backward} (product), so the reaction is consistently stored downhill in the PES-graph view.
\end{enumerate}
\noindent\begin{tabular}{lr}
\toprule
Parameter & Value \\
\midrule
Endpoint $f_\text{max}$ & \SI{0.005}{eV/\angstrom} \\
Endpoint max steps & 120 \\
Endpoint energy threshold & \SI{0.043}{eV} below TS (both sides) \\
Min endpoint RMSD & \SI{0.25}{\angstrom} \\
\bottomrule
\end{tabular}
\subsubsection{Stage 4: Endpoint classification and commitment}
\label{sec:pes-classify}

After IRC, a SMILES is inferred for each endpoint geometry from its bonded structure (\ref{sec:shared-smiles}). The reaction is then classified as one of three types. For \textbf{Intramolecular} reactions both endpoints have  SMILES descriptions matching the original compound. The TS connects two minima of the same compound's PES. The saddle point and its two endpoints are added to the per-compound graph (one PES graph per compound). New endpoints are deduplicated and merged into existing minima if any of the following criteria is met:
\begin{enumerate}
    \item RMSD (\ref{sec:shared-rmsd}) $<$ \SI{0.01}{\angstrom}
    \item RMSD $< \SI{0.20}{\angstrom}$ and $|\Delta E| < \SI{0.043}{eV}$
    \item otherwise NEB same-basin check (\ref{sec:shared-neb}) with \SI{0.043}{eV} barrier threshold
\end{enumerate}
Energy barriers are derived by force-integral (\ref{sec:shared-traj}) over the IRC trajectory.

An \textbf{Escaped} reaction includes one endpoint, which matches the originating compound's SMILES while the other does not. The IRC has discovered a reaction that leaves the originating compound's basin and lands in a chemically different species (possibly even multiple, if the relaxed product fragmented). The reaction is repackaged as an \texttt{ExplorationContext} with the originating compound as reactant and the escaped structure as product, and passed to the same multi-stage critic as the generative loop (SI-\ref{si:generative}, Stages 4--8). If neither endpoint matches the originating compound's SMILES or SMILES inference fails the reaction is classified as \textbf{anomalous} and discarded. In the main experiment, the PES loop produced ${\sim}24\,000$ intramolecular saddle points versus $69\,450$ escape candidates, of which $16\,879$ ($\sim 24\%$) survived the generative critic (SI-\ref{si:generative}, Stages 4--8). Surviving escape contexts are added to the reaction graph with \texttt{discovery\_method = "pes\_exploration"}. Compound registration and reaction deduplication are identical to the generative loop.
\subsubsection{Filtering statistics}
\label{sec:pes-funnel}

\begin{table}
\centering
\begin{tabular}{lr}
\toprule
Quantity & Count \\
\midrule
PES jobs run (minima explored)                        & $26\,282$ \\
Intramolecular saddle points found and committed            & $23\,720$ \\
Escape candidates produced                            & $69\,450$ \\
Escape candidates surviving generative critic (SI-\ref{si:generative}, Stages 4--8) & $16\,879$ \\
Escape critic survival rate                           & $\sim 24\%$ \\
\bottomrule
\end{tabular}
\caption{PES-loop totals for the main run. The escape critic survival rate ($24\%$) is much higher than the generative-loop submission-to-graph rate ($0.8\%$, SI-\ref{si:generative}).}
\label{tab:pes-funnel}
\end{table}

The PES loop contributed $\sim 36\%$ of all reactions in the graph ($16\,879$ of the $\sim 47\,000$ combined). It dominates two regions of chemical space where the generative loop is weak: (i)~conformational transitions within a compound (the entire intramolecular-saddle-point population, which the generative loop cannot produce because it does not sample within-compound saddle points), and (ii)~reactions whose TS geometry is far from the diffusion model's training distribution but is reachable by MD from a known minimum. 

\subsection{Reaction-Network Kinetics Solver}
\label{si:kinetics}

Exploration cannot afford to treat every discovered compound as equally worth pursuing. The kinetics solver supplies the priority signal: by simulating how the network discovered so far would actually evolve in time, it estimates which compounds accumulate in appreciable amounts, and the exploration then concentrates its proposals there. Concretely, the kinetics solver consumes the current reaction graph and integrates a mass-action ODE to a long-time steady state. The resulting steady-state composition is the distribution used by the generative-loop sampler (SI-\ref{si:generative}) and used to determine which compound geometries to explore (SI-\ref{si:pes}) using the PES exploration loop.

For $n_\text{species}$ species and $n_\text{reactions}$ reactions, the ODE is the standard mass-action system
\begin{equation}
    \frac{d\mathbf{y}}{dt} = \mathbf{S}_\text{net}\,\mathbf{r}(\mathbf{y}), \qquad r_j(\mathbf{y}) = k_j^+ \prod_i y_i^{\nu_{ij}^-} - k_j^- \prod_i y_i^{\nu_{ij}^+},
\end{equation}
where $\mathbf{S}_\text{net} = \mathbf{S}^+ - \mathbf{S}^-$ is the net stoichiometry matrix and $\boldsymbol{\nu}^\pm$ are the reactant/product stoichiometry exponents. The right-hand side and its analytical sparse Jacobian
\begin{equation}
    J_{ik} = \frac{\partial}{\partial y_k}\!\left[\sum_j S_{ij}\, r_j(\mathbf{y})\right]
\end{equation}
are implemented as Numba-JIT kernels (\texttt{numba\_kernels.py}, \texttt{@njit(cache=True)}) to keep the inner solver loop in compiled code. Jacobian sparsity follows from the stoichiometry pattern: $J_{ik} \neq 0$ only where species $k$ appears as a reactant or product of a reaction that touches species $i$.

For each discovered reaction the forward and backward rate constants are
\begin{equation}
    k = \min\!\Bigl(\tfrac{k_B T}{h}\,\exp(-E_a / k_B T),\ k_\text{diff}\Bigr), \qquad k_\text{diff} = \SI{e10}{M^{-1}s^{-1}},
\end{equation}
with $k_B = \SI{8.617e-5}{eV/K}$, $h = \SI{4.136e-15}{eV\,s}$, and the input barrier capped at $E_a \leq \SI{10}{eV}$ before the exponential.

The barrier that feeds Eyring is chosen as follows:
\begin{enumerate}
    \item If computed, PBE0/def2-TZVPP separated barriers supplied by the DFT refinement process (\ref{sec:kinetics-dft}).
    \item Else, force-integrated in-box ML barriers along the IRC trajectory.
\end{enumerate}

Four seed equilibria (water autoionisation, CO$_2$ hydration, H$_2$CO$_3$ dissociation, proton solvation; SI-\ref{si:seed}) bypass Eyring entirely. Their rate constants are stored as constants and applied directly at any temperature. 

Several filters are applied at build time. Reactions are dropped from the ODE if:
\begin{enumerate}
    \item either side contains dot-disconnected SMILES (van der Waals dimers stabilised by dispersion; non-chemical)
    \item either barrier is below $-\SI{0.05}{eV}$ (negative-$E_a$ artefacts from PES/IRC landing on dianion or zwitterion minima below the separated reactant pair), or 
    \item both barriers exceed the cutoff of (default \SI{10}{eV}, effectively unreachable and only bloating the system)
\end{enumerate}
Note that these threshold violations are very rare as they only occur if DFT-energy calculations disagree strongly with ML energy predictions used to filter barriers at exploration time.
Reactions are also direction-agnostic deduplicated: $A + B \to C$ and $C \to A + B$ are reduced to one bidirectional reaction using the lowest $\min(E_a^\text{fwd}, E_a^\text{bwd})$. 
The Eyring prefactor at $T = \SI{500}{K}$ gives $k_B T / h \approx \SI{1.04e13}{s^{-1}}$, so a $\SI{1.0}{eV}$ barrier corresponds to $\sim \SI{9e2}{s^{-1}}$ and a $\SI{2.0}{eV}$ barrier to $\sim \SI{8e-8}{s^{-1}}$. The \SI{500}{K} default is a sampling choice: at the higher temperature, more reactions are kinetically accessible within the integration window, producing a richer steady-state distribution for the generative sampler to draw from. 

\subsubsection{DFT energy refinement}
\label{sec:kinetics-dft}
We use DFT to asynchronously refine the energy barriers during exploration. PBE0/def2-TZVPP is used for single-point evaluations via PySCF (\texttt{dft.RKS} for closed shell, \texttt{dft.UKS} for open shell; spin guessed from electron count modulo 2). We set SCF convergence \texttt{conv\_tol} $= 10^{-9}$, \texttt{max\_cycle} $= 300$, with a Newton retry on non-convergence. All energies are returned in eV. For each non-manual reaction the pipeline computes three single-point energies on the in-box geometry (same atom set throughout, since the IRC is performed in the merged-system box):
\begin{itemize}
    \item $E_\text{TS}^\text{DFT}$ at the TS conformer geometry,
    \item $E_R^\text{DFT}$ at the relaxed reactant frame (the final position of the IRC reactant-side trajectory),
    \item $E_P^\text{DFT}$ at the relaxed product frame (final position of the IRC product-side trajectory).
\end{itemize}
The in-box DFT barriers are then $\Delta E_\text{fwd}^\text{in-box} = E_\text{TS} - E_R$ and $\Delta E_\text{bwd}^\text{in-box} = E_\text{TS} - E_P$. For each compound participating in any DFT-refined reaction, a single-point at the compound's \emph{lowest-energy minimum} (the PES-graph entry with the smallest ML energy) is computed. The \emph{separated} barriers (used by the kinetics solver) are derived as
\begin{equation}
    \Delta E_\text{fwd}^\text{sep} = E_\text{TS}^\text{DFT} - \!\!\sum_{c \in \text{reactants}}\!\! E_c^\text{DFT}, \qquad
    \Delta E_\text{bwd}^\text{sep} = E_\text{TS}^\text{DFT} - \!\!\sum_{c \in \text{products}}\!\! E_c^\text{DFT},
\end{equation}
where the sums respect reactant/product stoichiometry. The seed species with prescribed initial concentrations are included even if no reaction involves them ($\mathrm{H_2O}$, $\mathrm{CH_2O}$, $\mathrm{CO_2}$, $\mathrm{H_2}$, $\mathrm{H^+}$, $\mathrm{OH^-}$, $\mathrm{H_2CO_3}$, $\mathrm{HCO_3^-}$, $\mathrm{H_3O^+}$).

Every species in the system receives a uniform baseline of $c_0 = \SI{e-3}{M}$ at $t = 0$. A species-concentration noise floor of \SI{e-20}{M} is applied throughout post-processing: anything below this is treated as ``not present'' for entropy and distribution computations.

The ODE is integrated to $t_\text{max} = \SI{e8}{s}$ ($> 3$ years, well past any realistic reactor residence time) using PETSc's BDF time-stepper with Numba-JIT RHS and Jacobian kernels.

PETSc is configured as follows:
\begin{itemize}
    \item Time stepper: \texttt{TS} type \texttt{bdf}, exact-final-time interpolation, max $5\times 10^7$ steps.
    \item Initial step: $\Delta t_0 = \SI{e-10}{s}$. 
    \item Tolerances: $\texttt{atol} = \SI{e-16}{}$, $\texttt{rtol} = \SI{e-12}{}$.
    \item Newton inner solver: \texttt{SNES} with a \texttt{KSP} preconditioner-only LU factorisation. By default PETSc's built-in sequential LU is used, and a parallel direct solver such as MUMPS can be selected by environment variable when available.
\end{itemize}
The kinetics simulations outputs a concentration trajectory of $200$ geometrically spaced points, plus the canonical \emph{decade grid} of $19$ points at $10^{-10}, 10^{-9}, \ldots, 10^{8}$~s. 

\subsection{Shared Functionality: ML Force Field, Geometry, and Optimization}
\label{si:shared}

The discovery loops (SI-\ref{si:pes}, SI-\ref{si:generative}) and the kinetics pipeline (SI-\ref{si:kinetics}) share functionality which is not specific to any one loop. This functionality is described here.
\subsubsection{ML force field (MD-ET)}
\label{sec:shared-mdet}

All energies, forces, and Hessians used by the on-line pipeline come from a single transformer-based neural network potential, \textbf{MD-ET}. It is a \emph{direct force model}: forces are predicted as a per-atom 3-vector output head rather than obtained by differentiating an energy. Energy and force are independent heads, and the Hessian is obtained by automatic differentiation of the force head with respect to atomic positions. We use a dense formulation of MD-ET, i.e., there is no cutoff distance. See \cite{eissler2026simple} for details. We use the 12-layer model weights available in the \hyperlink{https://github.com/mx-e/md-et}{https://github.com/mx-e/md-et} github repo. We also use the inference code available there, specifically the ASE calculator interface. Other models could also be used.

A conventional MLFF obtains forces as $-\nabla_{\mathbf{r}} E$, which guarantees a conservative field and an exactly symmetric Hessian. However, obtaining a Hessian requires taking a second derivative which in practice requires backpropagation through the computational graph of the force prediction (which is already a differentiation through the neural network). This is not only expensive but also increases the magnitude of numerical inaccuracies. MD-ET instead predicts forces from a dedicated head which allows the prediction of Hessians using a single derivative step. While this is much faster and more stable, MD-ET's force field is not the exact gradient of a scalar energy, so the numerically differentiated Hessian $\mathbf{H}_{ij} = -\partial F_i/\partial x_j$ is not guaranteed to be symmetric. In practice, however, MD-ET predicts nearly conservative forces and the non-symmetric term in the Hessian is very small (i.e. below one percent of total magnitude). To still obtain a physically meaningful Hessian, and to guarantee that the Hessian matrix is positive semi-definite we project onto the symmetric part, $\mathbf{H} \leftarrow \tfrac{1}{2}(\mathbf{H} + \mathbf{H}^{\top})$, before any eigenanalysis. The projection is exact (information-free) for a perfectly conservative model and is the closest symmetric matrix to $\mathbf{H}$ in the Frobenius norm otherwise. We use the magnitude of the antisymmetric part of the Hessian prediction as a diagnostic of how far the model departs from energy conservation at a given geometry.

\subsubsection{Permutation-invariant RMSD}
\label{sec:shared-rmsd}

Comparing two molecular geometries requires aligning them in space and resolving the labeling/permutation ambiguity for chemically equivalent atoms. The system uses Kabsch alignment for rigid-body rotation and a two-strategy resolution for atom permutations:

\begin{enumerate}
    \item \textbf{Kabsch alignment.} Centre both structures at the origin and compute the optimal rotation via singular-value decomposition of the cross-covariance:
    \begin{equation}
        \mathbf{R}^* = \arg\min_{\mathbf{R} \in SO(3)} \sum_i \lVert \mathbf{r}_i^{(1)} - \mathbf{R}\,\mathbf{r}_i^{(2)} \rVert^2.
    \end{equation}
    The SVD-based form handles the right-handed-frame correction automatically (no reflections introduced).

    \item \textbf{Permutation resolution.} Define $\sigma$ over atoms of identical atomic number. RMSD is computed as
    \begin{equation}
        \text{RMSD} = \min_{\sigma}\ \sqrt{\frac{1}{N}\,\sum_i \lVert \mathbf{r}_i^{(1)} - \mathbf{R}^* \mathbf{r}_{\sigma(i)}^{(2)} \rVert^2}.
    \end{equation}
    The minimization uses a hierarchical heavy-atom-first strategy. Heavy atoms (C, N, O) typically form only small equivalent groups, so all permutations of equivalent heavy atoms are brute-force enumerated; for each heavy-atom permutation the abundant hydrogens are assigned by a pass of the Hungarian algorithm on the Kabsch-aligned configuration.
\end{enumerate}
Before any comparison the atoms are reordered into the RDKit canonical-rank order. This makes the RMSD deterministic across workers and across different input orderings.

\subsubsection{Connected-component fragment detection}
\label{sec:shared-fragments}

A molecular geometry is decomposed into chemically-meaningful fragments by building an adjacency matrix from a covalent-radius-scaled distance criterion,
\begin{equation}
    \text{adj}_{ij} = \mathbb{I}\bigl[\lVert r_i - r_j \rVert < 1.3\,(r^{\text{cov}}_i + r^{\text{cov}}_j)\bigr],
\end{equation}
with covalent radii from ASE. Connected components are extracted by depth-first search and returned as sorted index tensors. The $1.3\times$ scale is conservative; values around $1.15$--$1.20\times$ are more typical for cheminformatics but would mis-classify the stretched bonds at un-converged IRC endpoints as dissociated.

A fragment is accepted iff it is non-empty and is either larger than a single atom or the atom is hydrogen.

\subsubsection{Hessian computation and projection}
\label{sec:shared-hessian}

The Hessian is the central object for both saddle-point search and validation. The full $3N \times 3N$ matrix is computed by automatic differentiation of the ML forces with respect to atomic positions (\ref{sec:shared-mdet}), and is then projected to remove the six trivial modes (three translations, three rotations) that contaminate the spectrum at non-zero geometric centering or non-zero overall angular momentum:
\begin{equation}
    \mathbf{H}_\text{proj} = \mathbf{P}\,\mathbf{H}\,\mathbf{P}, \qquad \mathbf{P} = \mathbf{I} - \sum_{k=1}^{6} \mathbf{v}_k \mathbf{v}_k^\top,
\end{equation}
where $\{\mathbf{v}_k\}$ is a mass-weighted orthonormal basis of the translation/rotation subspace obtained by Gram--Schmidt on the six analytical T/R vectors. The projected Hessian is diagonalized into $\{\lambda_i, \mathbf{u}_i\}$ pairs (sorted ascending); eigenvalues with $|\lambda_i| < 10^{-4}\,\text{eV/\AA}^2$ are masked as numerical-noise floor before any chemical interpretation.

\subsubsection{FIRE optimiser}
\label{sec:shared-fire}

The Fast Inertial Relaxation Engine \cite{bitzek2006fire} (used force-only) is a velocity-Verlet-based first-order minimiser. Each iteration performs the following operations:
\begin{enumerate}
    \item Perform Velocity-Verlet update of positions and velocities under the current forces.
    \item Project velocity onto force direction: $\mathbf{v} \leftarrow (1-\alpha)\mathbf{v} + \alpha\,|\mathbf{v}|\,\hat{\mathbf{F}}$.
    \item If $\mathbf{F} \cdot \mathbf{v} > 0$ (descent direction) for the past $N_\text{min}$ steps, increase $\Delta t$ by a factor $f_\text{inc}$ (up to $\Delta t_\text{max}$) and decay $\alpha$ by $f_\alpha$; otherwise reset $\mathbf{v} = 0$, reset $\alpha$ to its initial value, and shrink $\Delta t$ by $f_\text{dec}$.
\end{enumerate}
Default constants follow the original paper: $\Delta t_\text{init} = 0.1$, $\Delta t_\text{max} = 1.0$, $\alpha_\text{start} = 0.1$, $N_\text{min} = 5$. FIRE is employed for every relaxation that does not require a Hessian: per-fragment relax (SI-\ref{si:generative} Stage 7), merge encounter prep (SI-\ref{si:generative} Stage 2), and the NEB deduplication (\ref{sec:shared-neb}).

\subsubsection{Trust-region Newton minimiser}
\label{sec:shared-newton}

When a Hessian is available, for example during IRC endpoint relaxation (SI-\ref{si:pes} Stage 6, SI-\ref{si:generative} Stage 5), a second-order trust-region Newton step is used instead of FIRE:
\begin{enumerate}
    \item Project out T/R modes (\ref{sec:shared-hessian}), diagonalize, and floor the absolute eigenvalues to $|\lambda_\text{min}| = 10^{-2}\,\text{eV/\AA}^2$ to prevent pathologically large steps in flat (floppy) directions.
    \item Compute the rational-function downhill step in each mode, $p_i = g_i / \lambda_i$, where $g_i$ is the gradient projection on the $i$-th eigenvector.
    \item Constrain the total step to the current trust radius (clamp on per-mode magnitudes if the global norm exceeds the radius).
    \item Evaluate the trial position, compute $\rho = \Delta E_\text{actual} / \Delta E_\text{predicted}$ where $\Delta E_\text{predicted} = -\mathbf{f}\cdot\mathbf{p} + \tfrac{1}{2}\mathbf{p}^\top\mathbf{H}\mathbf{p}$, and accept if $\rho > 0.1$. On rejection, shrink the trust radius by a factor of $2$ and retry (up to 5 retries before giving up).
    \item After acceptance: shrink the trust radius if $\rho < 0.25$ or $\rho > 2.0$ (poor quadratic model), expand if $0.5 < \rho < 1.5$ and the step hit the trust boundary (good model), by a factor of $2.0$ when the projected Hessian is positive definite and $1.5$ otherwise.
\end{enumerate}
The default trust radius for endpoint relaxation is $\Delta_0 = \SI{0.3}{\angstrom}$, bounded by $[\SI{0.005}{\angstrom}, \SI{2.0}{\angstrom}]$.

\subsubsection{Intrinsic Reaction Coordinate (IRC)}
\label{sec:shared-irc}

Both loops validate a candidate transition state by following the imaginary normal mode in both directions and relaxing each endpoint to a minimum. The shared algorithm:

\begin{enumerate}
    \item Project the Hessian (\ref{sec:shared-hessian}). The smallest eigenvector $\hat{\mathbf{u}}_\text{TS}$ is the imaginary mode.
    \item Compute a curvature-adaptive displacement magnitude $\Delta x$ from the harmonic-approximation form
    \begin{equation}
        \Delta x = \mathrm{clip}\!\left(\frac{c}{\sqrt{|\lambda_\text{TS}|}},\ \Delta x_\text{min},\ \Delta x_\text{max}\right),
    \end{equation}
    with $c$, $\Delta x_\text{min}$, $\Delta x_\text{max}$ chosen per loop (SI-\ref{si:pes}: $c = \sqrt{0.172}$, range $[0.02, 0.5]\,\si{\angstrom}$; SI-\ref{si:generative}: $c = 0.2$, range $[0.05, 0.5]\,\si{\angstrom}$).

    \item Build $\mathbf{R}^\pm = \mathbf{R}_\text{TS} \pm \Delta x\,\hat{\mathbf{u}}_\text{TS}$ and relax each independently to a minimum using trust-region Newton (\ref{sec:shared-newton}). Loop-specific convergence: SI-\ref{si:pes} uses $f_\text{max} = \SI{0.005}{eV/\angstrom}$, max $120$ steps; SI-\ref{si:generative} uses the same $f_\text{max}$ but $350$ steps, reflecting the wider distribution of saddle-point curvatures the diffuser produces.

    \item Assign reaction direction by energy: the higher-energy endpoint is labelled \emph{reactant} and the lower-energy \emph{product}. 
\end{enumerate}

\subsubsection{Nudged Elastic Band (NEB)}
\label{sec:shared-neb}

NEB \cite{henkelman2000climbing} (see Figure \ref{fig:concept}~c) is used \emph{only} for the same-basin same-or-different decision in conformer deduplication. Since PES exploration is expensive we want to merge PES minima within the same overall basin and no meaningful barrier between them. To facilitate this, the two endpoints are optimally aligned (Kabsch + permutation; \ref{sec:shared-rmsd}). The number of interior images is adaptive on the endpoint RMSD,
\begin{equation}
    N_\text{images} = \mathrm{clip}\!\left(\Bigl\lceil \tfrac{\text{RMSD}}{0.15\,\si{\angstrom}} \Bigr\rceil,\ 5,\ 12\right),
\end{equation}
and images are interpolated using the Image-Dependent Pair Potential (IDPP) method via ASE, which produces a physically reasonable initial band even for substantial rearrangements.
NEB is conducted with a total FIRE budget of $N_\text{FIRE} = 150$ steps and a spring constant $k_\text{spring} = \SI{0.1}{eV/\angstrom^2}$:
\begin{itemize}
    \item \textbf{Phase 1 (regular NEB):} $2/3$ of the budget. All images relaxed onto the MEP with the standard tangent projection of the spring forces.
    \item \textbf{Early stopping:} if the peak energy after Phase 1 exceeds $\SI{1.0}{eV}$ above the higher endpoint, the band is considered unphysical and the same-basin check returns ``different basins''.
    \item \textbf{Phase 2 (CI-NEB):} remaining $1/3$ of the budget with the climbing-image variant: the highest-energy image is pushed against the spring forces toward the true saddle point, refining the barrier estimate without re-relaxing the rest of the band.
\end{itemize}
We employ the following criterion to determine if minima are located in the same basin:
\begin{equation}
    \Delta E^\ddagger = \max_i(E_i) - \max(E_1, E_2),
\end{equation}
the barrier above the \emph{higher} endpoint. If $\Delta E^\ddagger < \SI{0.043}{eV}$ (the same-basin threshold set by the PES loop, about $1$~kcal/mol or $1.7\,k_B T$ at room temperature), the two minima are merged: any barrier smaller than this is freely crossed thermally and treating the two minima as distinct would over-resolve the conformational landscape. While a finer resolution would lead to more interesting PES graphs, we were mostly interested in finding reactions for the experiments conducted in this work and thus decided to aggressively merge conformational minima to allocate less overall resources to PES exploration.

  \subsubsection{SMILES canonicalisation and charge inference}
  \label{sec:shared-smiles}
  
A 3D geometry is converted to a canonical SMILES via RDKit's bond perception (\texttt{rdkit.Chem.rdDetermineBonds.DetermineBonds}) followed by \texttt{Chem.MolToSmiles(\dots, canonical=True, isomericSmiles=True)}. The isomeric form preserves both atom chirality (\texttt{@}, \texttt{@@}) and double-bond geometry (\texttt{/}, \texttt{\textbackslash}), so stereoisomers receive distinct canonical SMILES and are tracked as separate compounds.

Integer charge is inferred from the perceived bond pattern by summing formal charges atom-by-atom after RDKit's valence inference. The charge is used to check charge conservation across a reaction. The DFT refinement (SI-\ref{si:kinetics}) re-derives charges from a wavefunction calculation downstream.

\subsubsection{Trajectory force integration}
\label{sec:shared-traj}

The barrier from the trajectory endpoint to the TS is computed by integrating the force component along the path,
\begin{equation}
    \Delta E^\ddagger \;=\; \int_{\mathbf{R}_\text{min}}^{\mathbf{R}_\text{TS}} \mathbf{F}(\mathbf{r}) \cdot d\mathbf{r}
    \;\approx\; -\,\tfrac{1}{2}\sum_{n} \bigl(\mathbf{F}_n + \mathbf{F}_{n+1}\bigr) \cdot \bigl(\mathbf{R}_{n+1} - \mathbf{R}_{n}\bigr),
\end{equation}
discretised by the trapezoidal rule over the stored trajectory frames (positions and forces, optionally with Hessians for a second-order correction). This is preferred over the endpoint energy difference $E_\text{TS} - E_\text{min}$ because MD-ET's force prediction error is lower than its energy prediction error.

\subsection{Seed Compounds and Buffer Equilibria}
\label{si:seed}

For the main experiment the CRN is bootstrapped with eight compounds and four reactions. The compounds set the chemical scope and provide the starting conditions for the kinetics ODE; the four reactions inject acid-base equilibria that lie outside the scope of the electron-free MLFF and are therefore seeded explicitly (the reasoning is detailed below).
\begin{itemize}
    \item $\mathrm{CH_2O}$ (formaldehyde), the primary substrate and the only species at non-trace initial concentration in the upstream formose-style runs;
    \item $\mathrm{H_2O}$ (water), $\mathrm{H_2}$ (rendered as \texttt{[HH]} since the MLFF has no electron concept and so $\mathrm{H}$ / $\mathrm{H^+}$ share that SMILES), universal small fragments loaded from XYZ files;
    \item $\mathrm{CO_2}$, $\mathrm{H_2CO_3}$, $\mathrm{HCO_3^-}$, $\mathrm{OH^-}$, $\mathrm{H_3O^+}$, the buffer species needed to support the manual equilibria.
\end{itemize}
Every seed compound enters with a 3D geometry that is canonicalised through RDKit so that its SMILES matches what the discovery pipeline would produce, and is registered with a single PES-graph minimum. Larger buffer compounds ($\geq 3$ atoms: $\mathrm{H_2CO_3}$, $\mathrm{HCO_3^-}$) are enqueued for PES exploration, smaller ones ($\mathrm{OH^-}$, $\mathrm{H_3O^+}$) have no internal conformers. Each seed receives an initial concentration of \SI{1}{mM} in the kinetics ODE (SI-\ref{si:kinetics}). 

Four acid--base reactions are seeded with forward and backward rate constants. The acid-base channels cannot be found by the employed ML methods as they do not represent explicit electrons. Manual equilibria are introduced to model the important kinetic roles these pathways play (i.e., maintaining a self-consistent pH, ensuring $\mathrm{CO_2/H_2CO_3/HCO_3^-}$ buffering, providing a reservoir of solvated protons). 

\begin{center}
\begin{tabular}{lll@{\hspace{2em}}rr}
\toprule
Name & Forward direction & & $k_\text{fwd}$ & $k_\text{bwd}$ \\
\midrule
Water autoionisation
    & $\mathrm{H_2O \to H^+ + OH^-}$
    && $1.8 \times 10^{-3}$ & $10^{10}$ \\
$\mathrm{CO_2}$ hydration
    & $\mathrm{CO_2 + H_2O \to H_2CO_3}$
    && $5.5 \times 10^{-3}$ & $178$ \\
$\mathrm{H_2CO_3}$ dissociation
    & $\mathrm{H_2CO_3 \to H^+ + HCO_3^-}$
    && $2.5 \times 10^{6}$  & $10^{10}$ \\
Proton solvation
    & $\mathrm{H^+ + H_2O \to H_3O^+}$
    && $10^{10}$ & $1.0$ \\
\bottomrule
\end{tabular}
\end{center}

\noindent Units: $\si{s^{-1}}$ for unimolecular forward and \si{M^{-1}.s^{-1}} for bimolecular. The values are literature rate constants.

\putbib[bib]            %
\end{bibunit}

\end{document}